\documentclass{article} 
\usepackage[T1]{fontenc} 
\usepackage[final]{colm2026_conference}

\usepackage{hyperref}
\usepackage{microtype}
\usepackage{url}
\usepackage{booktabs}

\usepackage{lineno}
\usepackage{graphicx}
\usepackage{subcaption}
\usepackage{wrapfig}

\usepackage{amsmath}
\usepackage{amssymb}
\usepackage{mathtools}
\usepackage{amsthm}

\theoremstyle{plain}
\newtheorem{definition}{Definition}

\definecolor{darkblue}{rgb}{0, 0, 0.5}
\hypersetup{colorlinks=true, citecolor=darkblue, linkcolor=darkblue, urlcolor=darkblue}

\usepackage{multirow}
\usepackage{colortbl}
\usepackage{xspace}
\usepackage{tcolorbox}

\definecolor{Gray}{gray}{0.9}
\definecolor{lightred}{RGB}{255,179,179}
\newcommand{\epone}{$\theta_1$\xspace}
\newcommand{\eptwo}{$\theta_2$\xspace}
\newcommand{\eponeun}{$\theta_1'$\xspace}
\newcommand{\eptwoun}{$\theta_2'$\xspace}

\title{Understanding Machine Unlearning Through the Lens of Mode Connectivity}

\author{Jiali Cheng, Hadi Amiri\\
University of Massachusetts Lowell\\
\texttt{\{jiali\_cheng, hadi\_amiri\}@uml.edu}\\
}

\begin{document}

\ifcolmsubmission
\linenumbers
\fi

\maketitle

\begin{abstract}
Machine Unlearning aims to remove undesired information from trained models without full retraining from scratch. 
Despite recent progress, the loss landscape and optimization geometry of unlearning are poorly understood. 
In this paper, we study machine unlearning through the lens of mode connectivity--the phenomenon that independently trained models can often be connected by smooth low-loss paths in parameter space. 
We introduce {\em mode connectivity in unlearning} (MCU) and evaluate it across a range of settings, including curriculum learning, second-order optimization, and connectivity across different unlearning methods.
We find that many unlearned models lie in connected basins with smooth retain/forget behavior, while changes in training dynamics can move solutions into different basins.
MCU also reveals that models within the same basin can differ substantially on privacy metrics, and that unlearning progresses nonlinearly from the original model to the unlearned model. In addition, linear connectivity suggests that most approximate unlearning methods are mechanistically distinct from retraining. Finally, MCU-based ensembling can improve generalization and robustness to relearning attacks, and MCU smoothness correlates with unlearning difficulty.
To our knowledge, this is the first study of machine unlearning through the lens of mode connectivity\footnote{Code: \href{https://github.com/CLU-UML/Mode-Connectivity-in-Unlearning}{https://github.com/CLU-UML/Mode-Connectivity-in-Unlearning}}.\looseness-1
\end{abstract}

\section{Introduction}
The widespread deployment of machine learning models raises the need for {\em machine unlearning}--the process of removing specific knowledge from a trained model without affecting other knowledge~\citep{9519428,liu2024rethinking}. This need is driven by both legal and ethical imperatives, such as removing copyrighted data~\citep{eldan2023whos}, as well as practical necessity of purging outdated or incorrect information~\citep{dhingra-etal-2022-time}. As models scale in size and training cost, understanding unlearning methods is becoming an important research frontier in trustworthy machine learning.

\begin{figure}
\centering
    \includegraphics[width=0.98\linewidth]{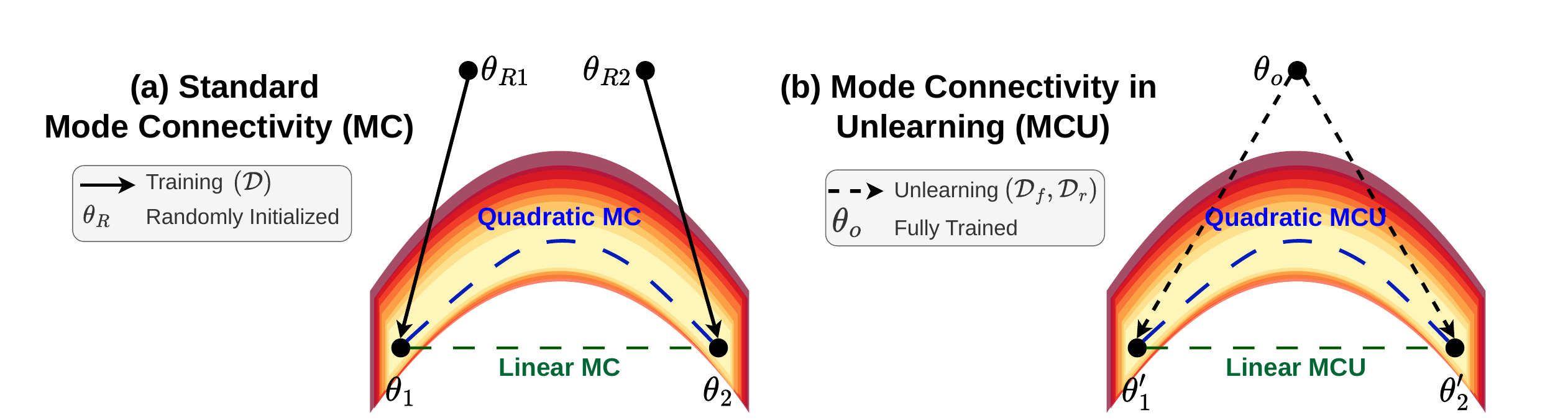}
    \caption{\textbf{(a)}: Illustration of standard mode connectivity (MC): MC finds a smooth curve connecting two endpoints that yields consistent low loss on $\mathcal{D}$.
    \textbf{(b)}: Illustration of mode connectivity in unlearning (MCU): unlearning removes knowledge of forget set $\mathcal{D}_f$ from the trained model $f_{\theta_o}$ while maintaining knowledge of retain set $\mathcal{D}_r = \mathcal{D} \setminus \mathcal{D}_f$. 
    MCU finds a smooth curve connecting the two unlearned models \eponeun and \eptwoun that yields consistent low loss on $\mathcal{D}_r$ and high loss on $\mathcal{D}_f$. See details in \S~\ref{sec:mcu}.\looseness-1}
\label{fig:fig1}
\end{figure}

Concurrently, the phenomenon of {\em mode connectivity} has shown that independently trained models can often be connected by low-loss paths in parameter space~\citep{garipov2018loss,qin-etal-2022-exploring} (Figure~\ref{fig:fig1}a).
These findings have important implications for understanding loss landscape, model ensembling, and generalization~\citep{garipov2018loss,Zhao2020Bridging}.\looseness-1

However, existing studies on mode connectivity have largely focused on image classification tasks~\citep{pmlr-v80-draxler18a,vrabel2025input} with simple optimization regimes. Whether mode connectivity extends to unlearning remains unexplored. 
More broadly, despite recent advances~\citep{liu2024protecting,hong-etal-2024-dissecting}, the loss landscape of unlearning remains poorly understood, along with its implications for the behavior and properties of unlearning.\looseness-1


To bridge these gaps, we introduce and formalize the concept of \textbf{Mode Connectivity in Unlearning (MCU)}--a framework to study unlearning from the loss landscape perspective. Specifically, we investigate the following research question:

\begin{tcolorbox}[colback=gray!8, colframe=black, boxrule=0.6pt, arc=2pt, left=3pt, right=3pt, top=0pt, bottom=0pt]
\textbf{RQs}: \emph{What can MCU reveal about unlearning methods, including loss landscape geometry, mechanistic similarity, and robustness to relearning attacks?}
\end{tcolorbox}


    


Unlike prior mode connectivity work, which studies standard training under a single objective, MCU must simultaneously preserve performance on the retain set and induce forgetting on the forget set.
Answering this question provides insight into the generalization, stability, and interpretability of unlearning methods. 
For instance, a smooth path between two unlearning solutions that maintains strong retain behavior and forgetting may indicate shared inductive biases or similar optimization dynamics, while a lack of connectivity may indicate divergent solution structures.

Through extensive experiments on diverse tasks and different training paradigms, we find that unlearning has a smooth loss landscape. Unlearned models with different optimizations tend to converge into the same smooth basin (\S\ref{sec:loss_landscape}--\S\ref{sec:difficulty}). 
Models within the same low-loss basin can function drastically different, especially on privacy-related metrics (\S\ref{sec:het_basin}).
We discover that along the change from original model to unlearned model in parameter space, unlearning behavior is not evenly distributed, and first suppresses memorization of textual content, while later commits to deeper forgetting (\S\ref{sec:orig_unlearn}).
MCU also demonstrates that unlearning is generally mechanistically different from 
retraining. 
Additionally, MCU-based ensemble can improve generalization and robustness of unlearning (\S\ref{sec:robustness}).
Finally, we show that MCU smoothness is a proxy for unlearning difficulty (\S\ref{sec:difficulty}).
These insights provide new directions for future unlearning research.
Our contributions are:\looseness-1
\begin{itemize}
    \item \textbf{Mode Connectivity in Unlearning (MCU)}: We introduce MCU--a framework for studying machine unlearning through loss landscape geometry. 



    \item \textbf{Analysis of Unlearning Loss Landscape}: We evaluate MCU across LLM and classification unlearning benchmarks, multiple unlearning algorithms, and diverse training settings including curriculum learning, second-order optimization, and cross-method connectivity.

    
    \item \textbf{Insights into Unlearning}: MCU reveals mechanistic differences between unlearning and retraining, functional heterogeneity within connected basins, and a link between connectivity smoothness and unlearning difficulty. We also show that MCU-based ensembling improves generalization and robustness to relearning attacks.\looseness-1
    
\end{itemize}

Overall, our results suggest that MCU is useful both as a diagnostic tool for understanding unlearning geometry and as a practical tool for improving robustness and model selection.

\section{Preliminaries}\label{sec:prem}

\paragraph{Notation}
Let $f_{\theta_o}$ be a model trained on dataset $\mathcal{D}$ with task loss $L$. In addition, assume that $\mathcal{D}$ can be divided into two disjoint sets: the forget set $\mathcal{D}_f$ and the retain set $\mathcal{D}_r = \mathcal{D} \setminus \mathcal{D}_f$.

\paragraph{Machine Unlearning} 
Machine unlearning aims to remove the influence of the forget set $\mathcal{D}_f$ from the trained model $f_{\theta_o}$ and preserve the knowledge of retain set $\mathcal{D}_r$. A good unlearning model $f'$ should achieve high loss on $\mathcal{D}_f$ and low loss on $\mathcal{D}_r$.
A commonly used solution is to fine-tune the original model $f_{\theta_o}$ to minimize the task loss on $\mathcal{D}_r$ while maximizing the task loss on $\mathcal{D}_f$~\citep{jia-etal-2024-soul,liu2024rethinking,Cheng2026}. 
%
For example, GradDiff~\citep{maini2024tofu} directly implements the above approach: 
\begin{equation}
    f' = \arg\min_{\theta'} L(\mathcal{D}_r) - L(\mathcal{D}_f).
\end{equation} Details of related work and unlearning methods are discussed in Appendix~\ref{sec:related} and \ref{app:methods}.



\paragraph{Mode Connectivity}
    Let \epone and \eptwo denote the weights of two independently trained models on some dataset $\mathcal{\mathcal{D}}$ using loss $L$. The objective of mode connectivity is to find a curve $\phi(t) \rightarrow \mathbb{R}^{|\theta|}, t \in [0, 1]$ in the parameter space that connects the two minimizers $\theta_1$ and $\theta_2$, where $\phi(0) = \theta_1$ and $\phi(1) = \theta_2$. Curve $\phi(t)$ connecting \epone and \eptwo satisfies mode connectivity if the path $\phi(t)$ does not yield ``barriers,'' defined as sudden increase in loss~\citep{garipov2018loss,pmlr-v202-lubana23a}. Formally, $\forall t \in [0, 1]$:
    \begin{equation}\label{eq:mc}
        L\big(\mathcal{\mathcal{D}}; \phi(t)\big) \leq (1-t) \cdot L(\mathcal{\mathcal{D}}; \theta_1) + t \cdot L(\mathcal{\mathcal{D}}; \theta_2).
    \end{equation}

In the loss landscape, mode connectivity tries to find a low loss path $\phi$ connecting \epone and \eptwo without hitting any barrier. 
In other words, every set of parameter induced by $\phi(t)$ yield comparable performance to the minimizers \epone and \eptwo. 
The parametrization of $\phi$ determines the shape of the curve connecting the two minimizers \epone, \eptwo. Below, we present two commonly used curve types: 
\begin{itemize}\label{sec:curve_shape}
    \item \textbf{Linear:} a linear interpolation of minimizers with no optimization involved, i.e. $\phi(t) = (1-t) \theta_1 + t \theta_2$. Stronger linear connectivity indicates stronger mechanistic similarity of minimizers, such as their inductive biases~\citep{pmlr-v202-lubana23a}.
    
    \item \textbf{Quadratic:} a smooth quadratic curve connecting two minimizers, i.e $\phi(t) = (1-t)^2 \theta_1 + 2t(1-t) \theta_{12} + t^2 \theta_2$, where $\theta_{12}$ needs to be trained explicitly. 
\end{itemize}
    
To find non-linear curve, we can minimize the total accumulated loss along the curve to find the midpoint $\theta_{12}$ which will later be used for interpolation. More details of the curve finding optimization are discussed in Appendix~\ref{sec:curve_finding}.

\section{Mode Connectivity in Unlearning (MCU)}\label{sec:mcu}

\begin{definition}[Mode Connectivity in Unlearning]\label{sec:mcu_definition}
    As illustrated in Figure~\ref{fig:fig1}, let \eponeun and \eptwoun denote the weights of two unlearned models of applying some unlearning procedure $\mathcal{U}$ on the original model $f_{\theta_o}$ with different configurations.
    %
    MCU holds if there exists a path $\phi_{\theta_1' \rightarrow \theta_2'}(t)$ in parameter space that connects \eponeun and \eptwoun without yielding barriers. Formally, $\forall t \in [0, 1]$
    \begin{equation}\label{eq:mcu_dr_eg}
        L\big(\mathcal{D}_r; \phi(t)\big) \leq (1-t) \cdot L(\mathcal{D}_r; \theta_1') + t \cdot L(\mathcal{D}_r; \theta_2'),
    \end{equation}
    \begin{equation}\label{eq:mcu_df_eg}
        L\big(\mathcal{D}_f; \phi(t)\big) \geq (1-t) \cdot L(\mathcal{D}_f; \theta_1') + t \cdot L(\mathcal{D}_f; \theta_2').
    \end{equation}
\end{definition}
In MCU, ``barriers'' on the retain set $\mathcal{D}_r$ refer to the sudden \emph{increase} of task loss $L$ (as in standard mode connectivity), while ``barriers'' on the unlearn set $\mathcal{D}_f$ refer to the sudden \emph{decrease} of task loss. Eq.~\ref{eq:mcu_dr_eg} ensures that the task loss on the retain set $\mathcal{D}_r$ remains both low and smooth along the mode connectivity curve, indicating consistent model behavior during the unlearning process. Similarly, Eq.~\ref{eq:mcu_df_eg} enforces a high and smooth loss on the forget set $\mathcal{D}_f$ along the mode connectivity path. In other words, MCU is realized when there exists a continuous path of model weights connecting the endpoints \eponeun and \eptwoun, such that loss remains low on $\mathcal{D}_r$ and high on $\mathcal{D}_f$ along the curve. 

\paragraph{Connection to Standard Mode Connectivity}
In contrast to standard mode connectivity, MCU must satisfy objectives on both $\mathcal{D}_f$ and $\mathcal{D}_r$. Essentially, MCU examines whether it is possible to find a continuous curve between two endpoints such that there are no significant loss barriers in \textbf{two distinct loss landscapes}--a key difference compared to standard mode connectivity,  which typically considers only a single task or dataset.
Another key difference is that in standard mode connectivity~\citep{garipov2018loss}, the endpoints \epone and \eptwo are obtained by training the model from two random initializations. In contrast, the endpoints \eponeun and \eptwoun in MCU are both derived from the exact same trained model $\theta_o$.\looseness-1

\subsection{Influence of Training Dynamics on Mode Connectivity in Unlearning}\label{sec:novel_condition}

Unlearned models can be trained by various paradigms, such as curriculum learning (CL)~\citep{bengio_cl,barbulescu2024textual,zhao2024what,cheng2024mubench,cheng25d_interspeech} 
and second-order (SO) optimization~\citep{jia-etal-2024-soul,zhang2024towards}. 
To systematically analyze the impact of these factors, we define several novel experimental conditions and evaluate MCU under these conditions. 
For CL, we define two settings:
(a): both endpoints are trained with CL (\textbf{-CL}); and 
(b): one endpoint is obtained through CL and the other does not (\textbf{CL-Non-CL}). 
Similar settings apply to SO.
These configurations allow us to assess whether changes in sample learning order affect the emergence of MCU. 


In addition, we examine whether endpoints derived from different unlearning methods can be smoothly connected. We hypothesize that methods with similar formulation and inner mechanisms are more likely to establish connectivity.
This experiment provides a lens into mechanistic similarity of different unlearning methods~\citep{pmlr-v202-lubana23a}.\looseness-1

To the best of our knowledge, with the exception of the randomness factor, all other factors discussed above are novel within the mode connectivity literature. Multiple configurations can be combined, as summarized in Table~\ref{tab:setting}. Together, they provide diverse and realistic perspectives on the training dynamics of unlearning, broaden the scope of mode connectivity research, and deepen our understanding of the factors that enable or prevent successful unlearning.\looseness-1

\begin{table*}[t]
\setlength{\tabcolsep}{3pt}
\small
\caption{MCU settings. 
CL and SO (FO) denote curriculum learning and second-order (first-order) optimization respectively. All settings except ``Rand'' are novel in mode connectivity.\looseness-1}
\begin{center}
\begin{tabular}{p{4.1cm}|c|c|c|c|c}
\toprule
    \textbf{Endpoints unlearned with} & \textbf{Standard} & \textbf{Both CL} & \textbf{Both SO} & \textbf{Mixed CL/Non-CL} & \textbf{Mixed FO/SO} \\
    \midrule
    Different Randomness & Rand & Rand-CL & Rand-SO & CL-Non-CL & FO-SO \\
    \midrule
    Different unlearning methods & Met & Met-CL & Met-SO & Met-CL-Non-CL & Met-FO-SO \\
\bottomrule
\end{tabular}
\end{center}
\label{tab:setting}
\end{table*}

\section{Experimental Setup} \label{sec:experiment}

%

\paragraph{Datasets and Forget Sets}
We analyze MCU on widely adopted LLM unlearning and classification unlearning benchmarks. For LLM unlearning, we use TOFU~\citep{maini2024tofu}, MUSE~\citep{shi2025muse}, and WMDP~\citep{li2024the} dataset. For classification, we use three datasets from MU-Bench~\citep{cheng2024mubench}: image classification on CIFAR-10~\citep{cifar100}, biomedical text relation classification on DDI2013~\citep{segura-bedmar-etal-2013-semeval}, and image-text visual entailment on NLVR2~\citep{suhr-etal-2019-corpus}. The original models and standard data splits are provided by MU-Bench~\citep{cheng2024mubench,cheng2023multimodal} and open-unlearning~\citep{dorna2025openunlearning}. Specifically, TOFU has 1\%, 5\%, 10\% of forget set. MUSE has \texttt{Books} and \texttt{News} as forget sets, while WMDP has \texttt{Cyber-security} (Cyber) as forget sets. MU-Bench provides 2\%, 4\%, 6\%, 8\%, 10\% forget set.

\paragraph{Unlearning Methods}
We use MCU to analyze the following LLM unlearning methods: 
1) Gradient Ascent (GA)~\citep{Golatkar2020EternalSO}, 2) GradDiff (GD)~\citep{maini2024tofu}, 3) Negative Preference Optimization (NPO)~\citep{zhang2024negative}, 4) SimNPO~\citep{fan2024simplicity}, and 5) RMU~\citep{li2024the}.
For classification tasks we use: 
1) Gradient Ascent (GA)~\citep{Golatkar2020EternalSO}, 2) Random Labeling (RL)~\citep{amnesiac_2021}, 3) Bad Teaching (BT)~\citep{Chundawat_Tarun_Mandal_Kankanhalli_2023}, and 4) Saliency Unlearning (SU)~\citep{fan2024salun}. 
These methods cover a diverse set of unlearning paradigms and are commonly used in existing works. Appendix~\ref{app:methods} provides additional details.

\paragraph{Evaluation}
To evaluate MCU, we sample multiple points by varying the interpolation weight $t \in [0, 1]$ with small step size. Each value of $t$ induces a set of model weights according to the parametrization of the curve $\phi_{\theta}(t)$ (\S~\ref{sec:curve_shape}). Following previous mode connectivity work on language models~\citep{qin-etal-2022-exploring}, we sample 16 points with equal step size in $[0, 1]$.
For each induced model $\theta_{\phi}(t)$, we evaluate its performance using the standard benchmark-specific metrics. 
On TOFU, we use Forget Quality ($\uparrow$) and Model Utility ($\uparrow$) which is a $p$-value of Kolmogorov-Smirnov test (KS-Test). 
On MUSE, we use Forget VerbMem ($\downarrow$), Forget KnowMem ($\downarrow$), Retain KnowMem ($\uparrow$), and privacy leakage (\textbar PrivLeak\textbar, $\downarrow$).
On WMDP, we use accuracy on forget set ($\downarrow$) and accuracy on MMLU evaluation set ($\uparrow$).
Appendix~\ref{sec:metrics} provides additional details and metrics.
On classification unlearning tasks, we use
accuracy on test set $D_t (\uparrow)$, 
accuracy on forget set $D_f (\downarrow)$, 
accuracy on retain set $D_r (\uparrow)$, and 
Zero-Retrain Forgetting score ZFR ($\uparrow$) which measures the prediction similarity of $D_f$ between unlearned and original models.

\begin{figure}[t]
\begin{center}
\includegraphics[width=0.99\linewidth]{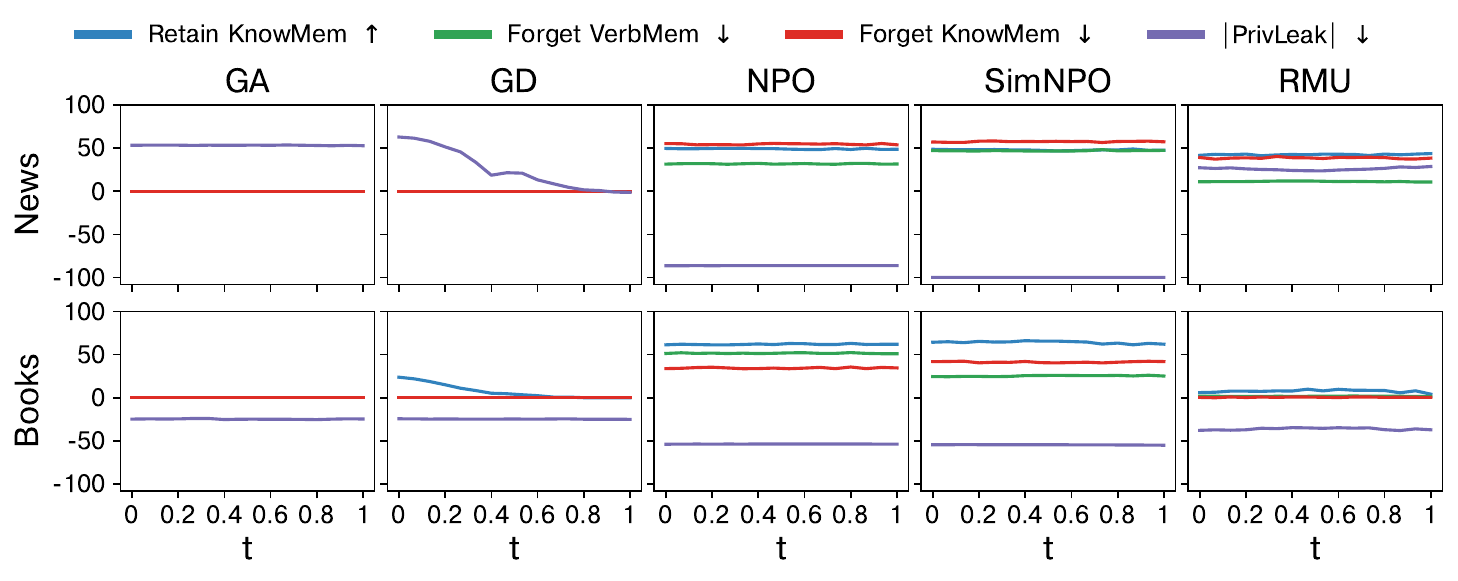}
\end{center}
\caption{MCU under Rand setting on MUSE dataset. Additional results are shown in Appendix~\ref{sec:additional_result} Figure~\ref{fig:tofu-rand}--\ref{fig:cls-fo-so}.}
\label{fig:fig2}
\end{figure}

\section{Results}\label{sec:res1}
With MCU, we study what geometric regimes arise in unlearning, and what they reveal about the solution space (\S\ref{sec:loss_landscape}--\ref{sec:difficulty}). Then we leverage MCU to analyze unlearning from different perspectives, including 
1) parameter displacement from original model (\S\ref{sec:orig_unlearn}), 
2) mechanistic similarity of unlearning methods (\S\ref{sec:similarity}), 
3) robustness to attacks (\S\ref{sec:robustness}), and 
4) unlearning difficulty (\S\ref{sec:difficulty}).

\subsection{Unlearning Has Smooth Loss Landscape}\label{sec:loss_landscape}
We investigate the conditions under which mode connectivity in unlearning (MCU) emerges across different models, datasets, and optimization strategies. Our results show that MCU is often prevalent.

We observe smooth curves with no fluctuation of unlearning quality (Forget VerbMem \& KnowMem) and utility retention (Retain KnowMem) on MUSE when both endpoints are unlearned with different random seeds, shown in Figure~\ref{fig:fig2}. MCU curves on NPO, SimNPO show consistently high forget quality (low forget ROUGE) and utility retention (high retain KnowMem), which suggests that unlearning solutions reside on a connected low-loss manifold. This means we find a basin with consistent high unlearning efficacy.
In other cases on GA, GD, RMU, smooth curves can appear between low quality unlearned models, i.e. a basin of low unlearning performance. This observation aligns with findings in standard mode connectivity~\citep{pmlr-v80-draxler18a}, where minima are not isolated but from a single connected manifold of low loss in parameter space, see Figure~\ref{fig:fig2}.
The existence of mode connectivity paths suggests that modern neural networks have enough parameters such that they can achieve good predictions while a big part of the network undergoes structural changes. 


\begin{wraptable}{r}{0.45\textwidth}
\small
\caption{Scaling of MCU smoothness.}
\begin{tabular}{r|ccc|ccc}
\toprule
\multirow{2}{*}{$|\mathcal{D}_f|$} & \multicolumn{3}{c|}{Max Barrier} & \multicolumn{3}{c}{Mean Barrier} \\
 \cmidrule{2-7}
 & 1B & 3B & 8B & 1B & 3B & 8B \\
\midrule
 1\% & 18.1 & 9.6 & 7.5 & 4.1 & 3.0 & 2.8 \\
 5\% &  8.2 & 5.9 & 3.0 & 0.9 & 0.8 & 0.8 \\
10\% &  2.9 & 1.7 & 0.0 & 0.3 & 0.1 & 0.0 \\
\bottomrule
\end{tabular}
\label{tab:scaling}
\end{wraptable}

\paragraph{Scaling}
We study the change of loss landscape geometry along two scaling axes: 1) increasing forget set, and 2) increasing model size. 
We define two metrics to quantify the barrier of MCU curve. 1) \emph{Max barrier} measures the largest interior rise of the curve above the worse endpoint. 

Given a scalar curve $f_{\theta(t)}, t \in [0,1]$, the max barrier is $B_{\max} = \max_{t\in[0,1]} f_{\theta(t)} - \max\{f_{\theta_0}, f_{\theta_1}\}.$
2) \emph{Mean barrier} measures the average positive excess above the worse endpoint, i.e. $B_{\mathrm{mean}} = \int_0^1 \max \Big( f_{\theta(t)}-\max\{f_{\theta_0}, f_{\theta_1}\}, 0 \Big) dt $. We only compute the average barrier on all $\mathcal{D}_f$ metrics, since the curve on $\mathcal{D}_r$ has minimal fluctuations in most cases.

Table~\ref{tab:scaling} shows a consistent pattern: MCU becomes more smooth as the forget set or the model scales up. However, the smoothness comes at the cost of bad unlearning quality. This suggests that scaling unlearning more knowledge on larger models is more challenging.

\begin{figure}[t]
\centering
    \begin{subfigure}{0.49\textwidth}
        \centering
        \includegraphics[width=\linewidth]{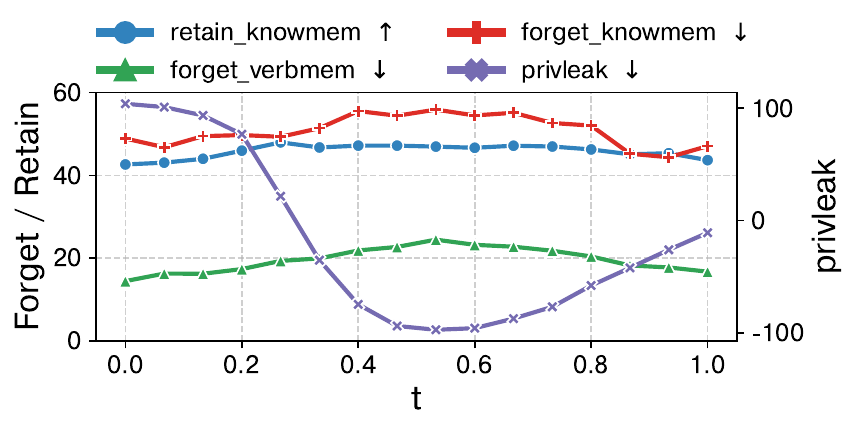}
        \caption{MCU on MUSE-News with RMU Unlearning}
    \end{subfigure}
    \begin{subfigure}{0.49\textwidth}
        \centering
        \includegraphics[width=\linewidth]{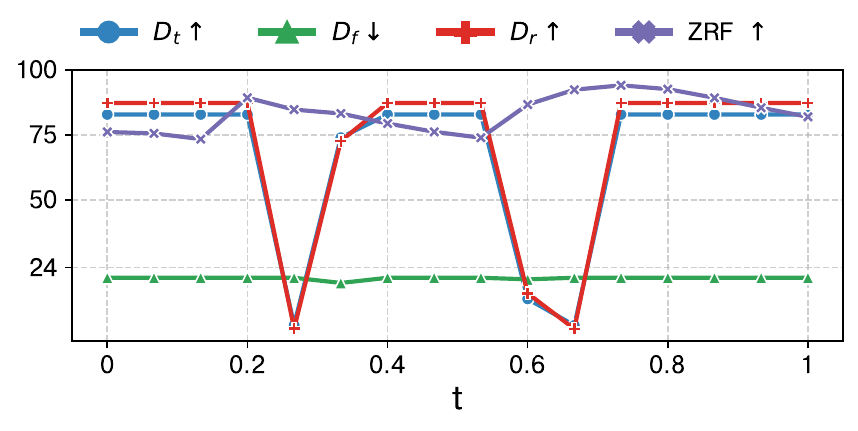}
        \caption{MCU on DDI with BT Unlearning}
    \end{subfigure}
\caption{Functionally heterogeneous basin. Models within the same basin in loss landscape can have different functional performance in different dimensions. (a) On MUSE with RMU, performance on $\mathcal{D}_f$ and $\mathcal{D}_r$ remains similar, but PrivLeak has large fluctuations within the range $[-100, 100]$.
(b) On DDI with BT, performance on $\mathcal{D}_f$ and ZRF remains stable, whereas $\mathcal{D}_t$ and $\mathcal{D}_r$ vary substantially.}
\label{fig:fragment}
\end{figure}

\begin{table}
    \centering
    \caption{Average barrier of $\mathcal{D}_f$ metrics of different MCU configurations.}
    \begin{tabular}{l|rrrrr}
    \toprule
    Barrier & Rand &  Rand-CL &  Rand-SO &  CL-Non-CL &    FO-SO \\
    \midrule
    Mean & 1.2 & 0.7 & 0.2 &   2.0 & 1.8 \\
    Max  & 3.0 & 2.2 & 1.2 &   8.8 & 8.1 \\
    \bottomrule
    \end{tabular}
    \label{tab:config}
\end{table}

\paragraph{CL and SO Make Unlearning Converge to Different Basins}\label{sec:training_paradigm}
We investigate if varying training paradigms can make unlearning methods converge to different basins. Note training paradigms do not change the unlearning algorithm (loss function). Therefore, the loss landscape remains unchanged for each algorithm.

These results in Table~\ref{tab:config} suggest that under CL-Non-CL and FO-SO, the MCU curves show large max barrier, indicating that two endpoints are likely to reside in two different and disconnected basins. When both endpoints are optimized with CL or SO, this behavior does not appear. This indicates that CL and SO can help unlearning converge to different solutions, highlighting potential future research directions.

\subsection{Functionally Heterogeneous Basin}\label{sec:het_basin}
We notice functional smoothness can differ significantly across different dimensions. 
Even minimizers reside in the same basin with similar low loss, their functional performances can vary significantly across different dimensions. 

For example, with RMU on MUSE-News, we observe a relatively smooth MCU curve, with trivial performance variation on Forget VerbMem, Forget KnowMem, and Retain KnowMem. However, performance on PrivLeak can fluctuate significantly along the same MCU curve, ranging from +100 to -100, see Figure~\ref{fig:fragment}(a). This indicates that although close to each other in parameter space, some minimizers are extremely prone to attackers and may leak information, while others may be much more robust. Another example is with BT on DDI. We observe stable accuracy on $\mathcal{D}_f$ along the MCU curve, but significant fluctuations on $\mathcal{D}_r$, $\mathcal{D}_t$ and moderate fluctuation on ZRF, see Figure~\ref{fig:fragment}(b). This indicates that although close to each other in parameter space, some minimizers maintain good amount of knowledge from the original model ($\mathcal{D}r$ and $\mathcal{D}_t$), while others may have completely forgotten.

On retain set $\mathcal{D}_r$, smooth connectivity are more likely to occur and easier to find. 
This could be because $\mathcal{D}_r$ is typically much larger than $\mathcal{D}_f$, which leads to a more stable optimization signal and a smoother curve in the retain region of the loss landscape.  

This behavior demonstrates that although residing in the same basin with similar loss, the unlearning behaviors can differ significantly in different functional dimensions, particularly for privacy related metrics. This also shows a limitation of current evaluation protocols for unlearning and motivates the need for richer and intrinsic evaluation on model parameters~\citep{hong2024intrinsic}.\looseness-1

\subsection{Probing the Unlearning Direction}\label{sec:orig_unlearn}
\begin{figure}[t]
\begin{center}
\includegraphics[width=0.99\linewidth]{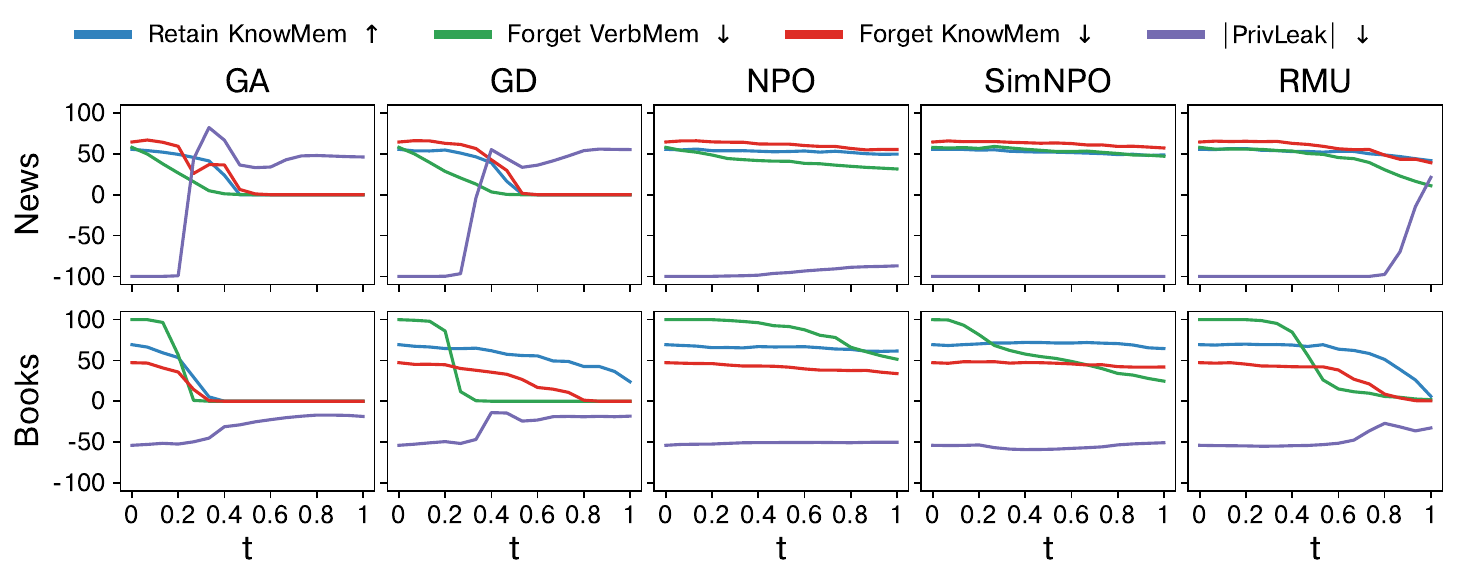}
\end{center}
\caption{Linear interpolation between the original model $\theta_o$ and the unlearned model $\theta_u$ on MUSE dataset. Additional results are shown in Appendix~\ref{sec:additional_result} Figure~\ref{fig:tofu-rand}--\ref{fig:cls-fo-so}.}
\label{fig:orig_unlearn}
\end{figure}

To probe whether parameter displacement along the unlearning direction is meaningful, we apply linear interpolation between the original model $\theta_o$ and the unlearned model $\theta_u$, i.e. $\theta(t) = (1 - t) \theta_o + t \theta_u, \quad t \in [0,1]$. 
This probe results in Figure~\ref{fig:orig_unlearn} reveals several important properties of unlearning. First, the clear asymmetry between the retain set $\mathcal{D}_r$ and the forget set $\mathcal{D}_f$ indicates the asymmetry of loss landscape. 



\paragraph{Non-linearly Distributed Unlearning}
We observe that the unlearning behavior does not accumulate linearly as $t$ increases. Unlearning does not emerge until deviation is accumulated to certain threshold.
In other words, model may remain functionally close to the original model for much of the path, and only near specific regions does strong forgetting activate.

\paragraph{Suppress First, Unlearn Later}
A recurring pattern is that lexical overlap metrics (e.g. forget ROUGE) improves earlier and more smoothly than overall unlearning efficacy metrics (e.g. forget quality). Similar pattern appear on MUSE in Figure~\ref{fig:orig_unlearn}, where verbatim memorization (Forget VerbMem) decreases first and knowledge memorization (Forget KnowMem) decreases later. This suggests that unlearning follows a \emph{first suppress, then unlearn} manner, where models first stop reproducing target content in surface form, while remaining relatively competent on the underlying knowledge to forget. As unlearning proceeds to a certain extent, models exhibit deeper level forgetting, such as change of the generated distributions. From this perspective, our analysis provides a complementary geometric lens for understanding how superficial unlearning emerges~\citep{hong2024intrinsic}.

\subsection{MCU Reveals Mechanistic (Dis)similarity between Unlearning Methods}\label{sec:similarity}
\citet{pmlr-v202-lubana23a} have found that if linear MC exists between two endpoints, they share internal mechanisms or rely on similar features when making predictions. We use this to analyze the mechanistic similarity between retraining and unlearning, and between two different unlearning methods. We find that utility retention is a poor proxy for mechanistic similarity. Most methods (except for GA) maintain relatively stable utility between retraining and unlearning.

\begin{figure}[t]
\centering
    \includegraphics[width=\linewidth]{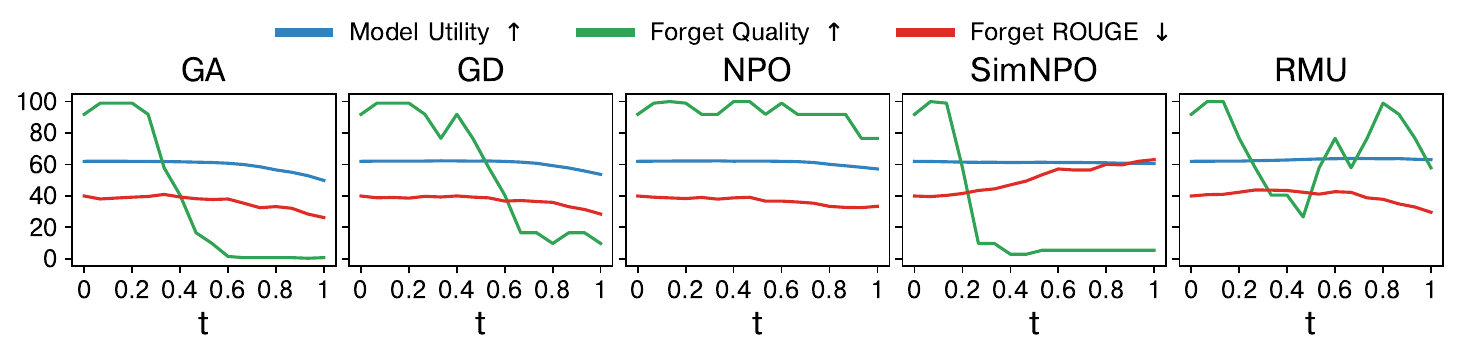}
\caption{Mechanistic dissimilarity between retraining and unlearning on TOFU. Additional results are shown in Appendix~\ref{sec:additional_result}, Figures~\ref{fig:tofu-met}--\ref{fig:tofu-met-fo-so}, Figures~\ref{fig:cls-met}--\ref{fig:cls-met-fo-so}.\looseness-1}
\label{fig:retrain_unlearn}
\end{figure}

\paragraph{Retraining vs. Unlearning}
In this setting, we probe the linear MCU between a model retrained from scratch ($\theta'_1 = \theta_{\text{Retrain}}$) and an unlearned model ($\theta'_2 = \theta_{\text{Unlearn}}$). We find that most unlearning methods are not smoothly connected to retraining under forgetting metrics, even when endpoint utility is comparable. This indicates that approximate unlearning often reaches functionally valid endpoints through mechanisms that are distinct from retraining, rather than by approximating the retraining solution along a shared linear subspace. Among the evaluated methods, only NPO demonstrates mechanistic similarity to retraining, shown in Figure~\ref{fig:retrain_unlearn}.

\paragraph{Different Unlearning Methods}
In this setting, we probe the linear MCU between two different unlearned models.
Figure~\ref{fig:muse-book-met} in Appendix~\ref{sec:additional_result} shows the pair-wise linear MCU between all methods. We can see that GA shows smooth MCU on forget set (knowmem and verbmem) with all other unlearning methods, indicating that all methods share similar internal mechanisms of handling forget set samples. They do differ in retain set significantly (Row 1 $t\rightarrow1$), since GA does not have retain mechanism. 

We observe high similarity on retain knowledge between NPO and SimNPO, an improved version of NPO. However, they differ from each other on forget knowledge and extraction strengths. Both of them are significantly different from GD, since GD simply minimizes loss on $D_r$ and maximizes loss on $D_f$, while NPO and SimNPO involves implicit reward modeling.




\subsection{MCU-based Ensemble Improves Generalization and Robustness}\label{sec:robustness}

\paragraph{Generalization}
Prior work has found that mode connectivity indicates generalization, where minimizers converge to a minima with smooth loss landscape and constant low error, potentially leading to improved performance~\citep{garipov2018loss,Wang_2023_CVPR}. 
Previous experiments demonstrate that intermediate minimizers along the MCU curve may outperform both endpoints. We then propose a generalization method, sampling $N=3$ minimizers along the curve with equal distance. We average the parameters of the sampled models and two endpoints, resulting in a new merged unlearner. Results on WMDP show that our ensemble strategy can outperform the endpoints in both unlearning effectiveness (forget accuracy), see Table~\ref{tab:ensemble}. This suggests that interpolated models may achieve a better trade-off between forgetting and retaining, and presents a promising directions for ensembling or model selection using mode connectivity. Other examples include RL on NLVR2.

\begin{wrapfigure}{r}{0.45\textwidth}
\begin{center}
\includegraphics[width=0.98\linewidth]{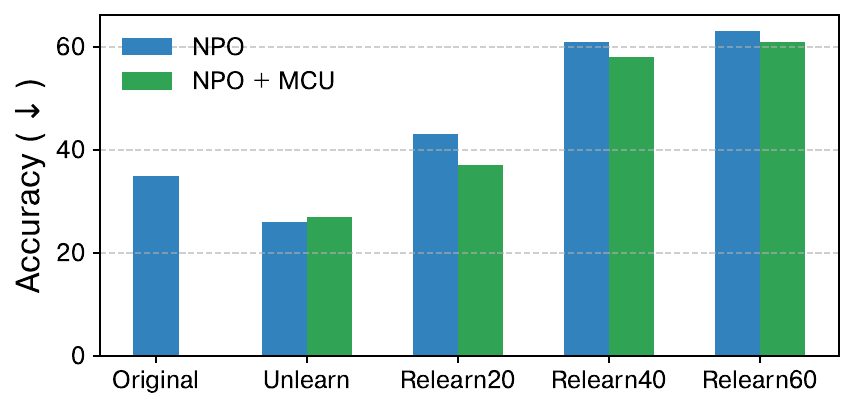}
\caption{MCU finds minimizers with smooth loss landscape and robust to relearning attack.}
\label{fig:attack}
\end{center}
\end{wrapfigure}

\paragraph{Robustness to Relearning}
It is well established that a smoother loss landscape is associated with greater robustness in deep learning models~\citep{zhang2017understanding,foret2021sharpnessaware,zhang2024duality,li2025tilted}. Specifically, minimizers along smooth mode connectivity curves are typically more robust than those obtained by standard training. Building on this insight, our findings suggest that MCU can identify unlearned models that are more resilient to adversarial threats, such as 
relearning attacks~\citep{hu2024unlearning,chen2026cure}.

Following prior work on unlearning robustness~\citep{fan2025towards,chen2026tracer}, we attack an unlearned model for WMDP Cyber by fine-tuning it on generic Wikipedia text and subsequently evaluate its unlearning effectiveness post-attack. As shown in Figure~\ref{fig:attack}, models unlearned via standard NPO methods are susceptible to relearning, with their forgetting effects quickly deteriorating. In contrast, models obtained through MCU - by averaging across multiple minimizers - demonstrate markedly greater robustness to such attacks. We attribute this improvement to the smoother loss landscape induced by the averaging process.
This observation is consistent with recent work showing that minimizing loss landscape sharpness (encouraging smoothness) during optimization enhances the robustness to adversarial attacks for LLM unlearning~\citep{fan2025towards}.

\begin{table}[t]
\centering
    \begin{minipage}[t]{0.45\textwidth}
    \centering
    \caption{MCU-based ensemble improves generalization.}
    \begin{tabular}{c|c|c}
    \toprule
    $\mathcal{D}_f$ Acc ($\downarrow$) & NPO & + MCU Ensemble \\
    \midrule
    63.7 & 24.1 & 21.5 \\
    \bottomrule
    \end{tabular}
    \label{tab:ensemble}
    \end{minipage}
\hfill
    \begin{minipage}[t]{0.45\textwidth}
    \centering
    \caption{MCU smoothness is a proxy for unlearning difficulty.}
    \begin{tabular}{l|c|c|c}
    \toprule
    Difficulty & Default & Hard & Easy \\
    \midrule
    Max Bar.  & 18.1 & 18.5 & 10.3 \\
    Mean Bar. &  4.1 &  6.3 &  3.2 \\ 
    \bottomrule
    \end{tabular}
    \label{tab:difficulty}
    \end{minipage}
\end{table}

\subsection{MCU as A Proxy for Unlearning Difficulty}\label{sec:difficulty}

Many prior work has discovered different forget sets can induce substantially different levels of unlearning difficulty. In particular, challenging forget sets may be adversarial~\citep{fan2024challenging}, closely adjacent to the test set~\citep{cheng2023gnndelete,chen2025frog}, highly memorized~\citep{zhao2024what}, paraphrased by LLMs~\citep{lucki2024adversarial,lynch2024eight,cheng2025tool}, mechanistically difficult as characterized by circuits~\citep{cheng-etal-2026-mechanistic}, or samples with high forget-retain overlaps~\citep{chen2025future,reisizadeh2026blur,chen2026cure,chen2026tracer}.


We hypothesize that the smoothness of the MCU basin can serve as a proxy for unlearning difficulty from the perspective of loss landscape geometry. In particular, smoother MCU may reflect a flatter and more navigable loss landscape, whereas irregular or disconnected MCU may indicate sharper geometry, more difficult optimization, and a more challenging forget-retain tradeoff.

To test this, we leverage the easy and hard forget set splits from \citet{cheng-etal-2026-mechanistic} as ground truth difficulty and computed the MCU smoothness of both splits. 
We find that both the mean and maximum barriers are larger on the hard split than on the easy split.
As shown in Table~\ref{tab:difficulty}, the MCU curve has more barriers with larger maximum barrier on the hard split, while on the easy split, MCU curve is smoother.

\section{Conclusion}

We introduce mode connectivity in unlearning (MCU) as a framework for understanding the loss landscape and optimization dynamics of machine unlearning. To the best of our knowledge, this is the first work that studies the loss landscape of unlearning with mode connectivity across various settings. We find that the emergence of mode connectivity can be influenced by task complexity, forget set size, and optimization strategies like curriculum learning and second-order methods.

Our experiments across diverse tasks, unlearning methods, and training configurations show that MCU provides a useful lens for analyzing the loss landscape of machine unlearning, and provides insights into mechanistic similarity, robustness, and unlearning difficulty.
We further show that MCU can be used as a diagnostic tool and open new directions for improving unlearning methods.




\section*{Ethical Considerations}
This work focuses on improving the transparency and reliability of machine unlearning, which is motivated by ethical considerations such as user data privacy, regulatory compliance, and the right to be forgotten. All experiments are conducted on publicly available datasets, and no personally identifiable or sensitive data is used.

\bibliography{reference,reference_mc,anthology}
\bibliographystyle{colm2026_conference}

\newpage
\appendix
\section{Related work}\label{sec:related}
\paragraph{Machine Unlearning}
Early unlearning methods span across efficiently retraining~\citep{bourtoule2021machine,pmlr-v119-wu20b}, model pruning~\citep{jia2023model}, manipulating gradients~\citep{pmlr-v134-ullah21a,Hoang_2024_WACV}, 
adversarial unlearning~\citep{setlur2022adversarial,wei2023shared}, and data augmentation~\citep{Choi_2024_CVPR}. 
Unlearning on LLMs recently draws more attention~\citep{eldan2023whos,ji2024reversing,kassem-etal-2023-preserving,cheng2025tool}. 
However, there is less attention on mechanistically understanding the loss landscape of machine unlearning methods.

\paragraph{Mode Connectivity}
Furthermore, several studies have shown that independently trained minimizers can be connected by low loss paths, a phenomenon known as mode connectivity~\citep{pmlr-v80-draxler18a,garipov2018loss,pmlr-v119-frankle20a}, across both vision and language models~\citep{qin-etal-2022-exploring}. During pruning, linear mode connectivity emerges only at early stage of training. This connectivity has been extended to multi-dimensional manifolds\citep{pmlr-v139-benton21a}, and alternative topologies such as star-shaped and geodesic connectivity\citep{lin2024exploring}.
Mode connectivity can also lead to more effective~\citep{garipov2018loss} or adversarially robust~\citep{Zhao2020Bridging,Wang_2023_CVPR} models if ensembling along the curve. \citet{vrabel2025input} discover that mode connectivity can happen in input space.
Existing works on mode connectivity focus on the learning process. There is no prior work that investigates mode connectivity in machine unlearning.

\section{Details of Unlearning Methods}\label{app:methods}
Below, we present the details of the unlearning methods used in our study.

\paragraph{Gradient Ascent}
Gradient Ascent (GA)~\citep{Golatkar2020EternalSO} performs gradient ascent on $D_f$ without any mechanism to maintain utility on the retain set $D_r$.
    
\paragraph{Random Labeling}
Random Labeling (RL)~\citep{Golatkar2020EternalSO} fine-tunes $f_{\theta_o}$ on $D_f$ with corrupted labels and the original $D_r$ (or a fraction of it if the entire $D_r$ is too large). This method aims to inject errors to the forget set.

\paragraph{Saliency Unlearning}
SalUn (SU)~\citep{fan2024salun} first finds parameters that are salient to unlearning $D_f$. Next, it performs Random Labeling but only updates the salient parameters.
    
\paragraph{Bad Teaching}
Bad Teaching (BT)~\citep{Chundawat_Tarun_Mandal_Kankanhalli_2023} forces the unlearned model to predict $D_r$ similarly to the original model and to predict $D_f$ similarly to an incompetent model (e.g. a randomly initialized model). It minimizes the KL-Divergence between prediction logits $\mathbb{KL}(f'(D_r) || f(D_r))$ on $D_r$ and maximizes KL-Divergence between prediction logits $\mathbb{KL}(f'(D_f) || f_d(D_f))$ on $D_f$, where $f_d$ is the incompetent model, e.g. a randomly initialized model.
    
\paragraph{Gradient Difference}
GradDiff (GD)~\citep{maini2024tofu} minimizes task loss on $D_r$ and maximizes task loss on $D_f$. 
    
\paragraph{Negative Preference Optimization}
NPO~\citep{zhang2024negative} is built upon the DPO~\citep{rafailov2023direct} algorithm to post-train LLMs. In the original DPO, each query $q$ corresponds to a winning response $y_w$ to prioritize and a losing response $y_l$ to suppress. NPO uses only the losing response with no winning response.

\section{Details of Evaluation Metrics}\label{sec:metrics}
We provide detailed descriptions of the evaluation metrics used in our analysis. On MU-Bench~\citep{cheng2024mubench} tasks, we follow the original paper to adopt accuracy as the evaluation metric. In addition, we employ Zero-Retrain Forgetting score ($\uparrow$)~\citep{Chundawat_Tarun_Mandal_Kankanhalli_2023}, which measures the similarity of prediction logits on $D_f$ between the unlearned model and a random model.

TOFU~\citep{maini2024tofu} evaluates the unlearned model using $p$-value of Kolmogorov-Smirnov test for Model Utility ($\uparrow$) and Forget Utility ($\uparrow$), which measure the similarity of probability distributions between the unlearned and retrained model. Additionally, we also include verbatim evaluation using ROUGE-L recall score on Retain Authors ($\uparrow$), Forget Authors ($\downarrow$), Real Authors ($\uparrow$), and World Knowledge ($\uparrow$).

MUSE~\citep{shi2025muse} evaluates the unlearned model using verbatim memorization on forget set (forget\_verbmem $\downarrow$) and knowledge memorization on forget and retain set (forget\_knowmem $\downarrow$, retain\_knowmem $\uparrow$), by probing the unlearned model with a series of question related to forget set. All of these scores are measured by ROUGE-L.

On WMDP~\citep{li2024the} evaluates the unlearned model using accuracy on WMDP forget set. It also evaluates the general utility of the unlearned model using the MMLU benchmark~\citep{hendrycks2021measuring}.


\section{Details of Curve Finding Process}\label{sec:curve_finding}

To find the curve that connects \epone and \eptwo, we can first compute the average loss along the curve: 

\begin{equation}
    \hat{\ell}(\theta) = \frac{\int L(\phi_\theta) \, d\phi_\theta}{\int d\phi_\theta}.
\end{equation}

The numerator $\int L(\phi_\theta) \, d\phi_\theta$ is the line integral of the loss $L$ along the curve $\phi_{\theta}$. It sums up the loss values at all points along the curve, weighted by the length of the curve in the parameter space.
Intuitively, it measures the total accumulated loss along the curve, accounting for how long the curve is in regions with high or low loss.

The denominator $\int d\phi_\theta$ is the total length of the curve in the parameter space. It normalizes the numerator by the total length, ensuring that the result does not depend on the specific parameterization of the curve (e.g., stretching or shrinking segments artificially).

Minimizing the above loss ensures that the path between the two sets of weights corresponds to models with consistently high accuracy. 

The integrals can be rewritten in terms of the parameter $t \in [0, 1]$ as 

\begin{equation}
    \hat{\ell}(\theta) = \int_0^1 L(\phi_\theta(t)) q_\theta(t) \, dt,
\end{equation}

\begin{equation}
    q_\theta(t) = \frac{\|\phi_\theta'(t)\|}{\int_0^1 \|\phi_\theta'(t)\| \, dt}.
\end{equation}

\begin{equation}
    \mathbb{E}_{t \sim [0,1]} \hat{\ell}(\theta) = \int_0^1 L(\phi_\theta(t)) q_\theta(t) \, dt.
\end{equation}

\section{Additional Results}\label{sec:additional_result}
We present detailed results on TOFU in Figure~\ref{fig:tofu-rand}--\ref{fig:tofu-met-fo-so} and on classification datasets in Figure~\ref{fig:cls-rand}--\ref{fig:cls-met-fo-so}.

\subsection{MCU under independently unlearned minimizers}\label{sec:additional_result_rand}
On TOFU, we find almost perfectly smooth curve with no degradation of unlearning quality on 3 out of 4 unlearning methods (GA, GD, and NPO). Along the curves, all model weights yield consistent unlearning quality, measured by a series of evaluation metrics, including forget quality, model utility, and ROUGE score. 
On the other hand when using method RL, the model weights along the curve is of consistently high quality in model utility but have slightly different forget quality. Specifically, in the middle part of the curve, we observe a drop of 0.1 point in forget quality and an increase of 0.05 point in forget ROUGE ($\downarrow$). However, since forget quality is the $p$-value of KS test, any value greater than 0.05 is considered as good unleared model, see Figure~\ref{fig:tofu-rand} for details.
As the size of forget set increases, indicated by different rows in Figure~\ref{fig:tofu-rand}, there is trivial variation of forget quality and model utility along the linear and quadratic curve on GA, GD, and NPO. On RL, we notice interesting behaviors. When $|D_f|=1\%$, forget quality degrades in the middle of the curve. When $|D_f|=5\%$, forget quality does not change significantly. When $|D_f|=10\%$, forget quality significantly increases in the middle of the curve. These behaviors are consistent on both linear and quadratic curves. We attribute these to the fact that RL is not an appropriate unlearning method for TOFU, which stuck in local optima and cannot ultimately converge to the low loss valley.

Therefore, we can find that the loss landscape of most unlearning methods on TOFU has essentially a flat low-loss valley where barriers, i.e. sudden performance degradation, rarely appear. This implies that, similar to learning~\citep{pmlr-v80-draxler18a}, minima of unlearning are perhaps best seen as points on a single connected manifold of low loss, rather than as the bottoms of distinct valleys for each individual unlearning method. The existence of mode connectivity paths suggests that modern neural networks have enough parameters such that they can achieve good predictions while a big part of the network undergoes structural changes. However, some unlearning methods may not converge to the low loss manifold, such as RL on TOFU dataset.

On classification dataset, we observe different patterns across different unlearning methods. On GA, it is generally easier to observe smooth MCU curve, both linear and quadratic, with small variation in forget set performance when $|D_f|=1\%$. 
Due to the similarity in design, RL and SU have very similar MCU patterns. Both types of curve yield models with degraded forget set performance ($\downarrow$) in the middle part of the curve (green line in Figure~\ref{fig:cls-rand}), with more prominent degradation on linear than quadratic curves. 
On BT, there is a strong linear MCU but the curve finding process fails to converge to meaningful quadratic MCU. This demonstrates that simpler connectivity may appear but hard to detect. We hypothesize that BT has a more rugged loss landscape than other methods, likely because it computes loss based on representations not directly on tasks loss.
These results highlight the difference in loss landscape of unlearning methods.

\paragraph{CL and Non-CL}
On classification datasets, GA shows strong linear and quadratic MCU. RL and SU show quadratic (but not linear) connectivity, with slight degradation of forget set performance. This indicates that CL-based and Non-CL-based methods can converge to the same low loss manifold. On BT when $|D_f|=4\%$, although there is almost no variation in forget set performance on linear curve, there is a major drop on retain set performance at the middle of the curve. Since MCU considers both forget set and retain set performance, this is not an emergence of MCU.

\begin{figure}[t]
\begin{center}
\includegraphics[width=0.8\linewidth]{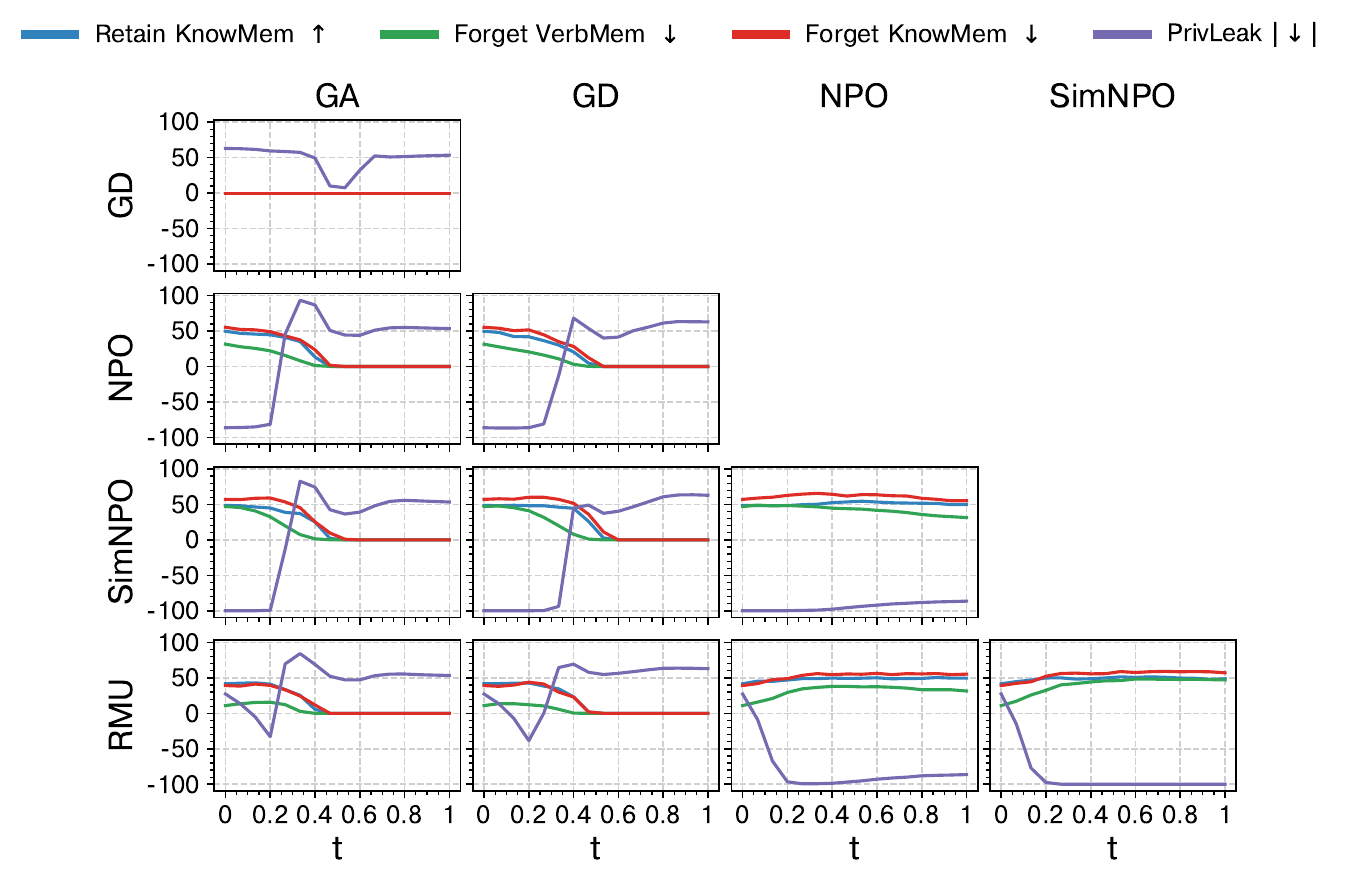}
\end{center}
\caption{MCU under \textbf{Met} setting on \textbf{MUSE News dataset}. Methods on rows and columns correspond to \eponeun and \eptwoun respectively. MCU is symmetric. Additional results are shown in Appendix~\ref{sec:additional_result}, Figures~\ref{fig:tofu-met}--\ref{fig:tofu-met-fo-so}, Figures~\ref{fig:cls-met}--\ref{fig:cls-met-fo-so}.\looseness-1}
\label{fig:muse-news-met}
\end{figure}

\begin{figure}[t]
\begin{center}
\includegraphics[width=0.8\linewidth]{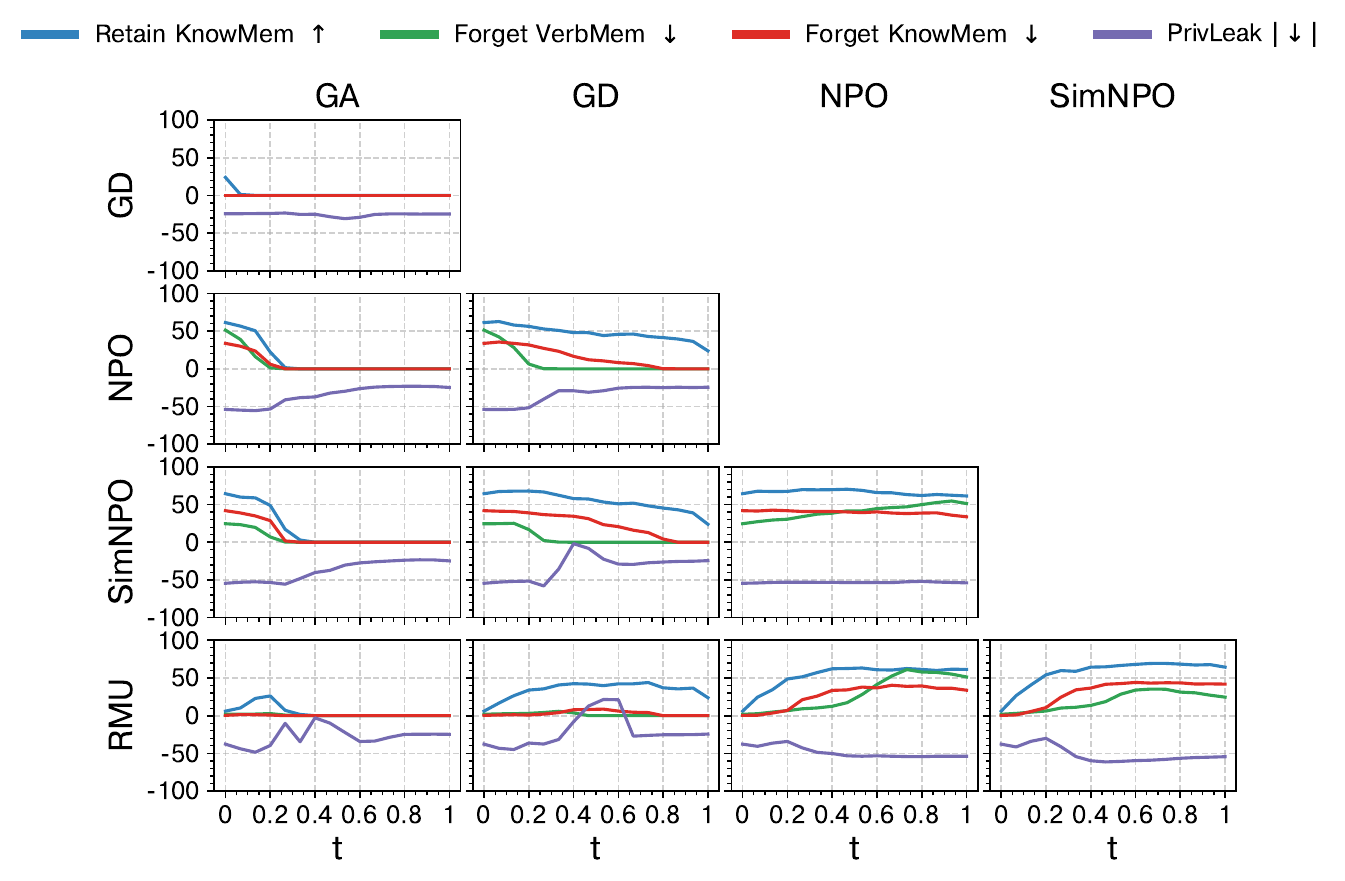}
\end{center}
\caption{MCU under \textbf{Met} setting on \textbf{MUSE Books dataset}. Methods on rows and columns correspond to \eponeun and \eptwoun respectively. MCU is symmetric. Additional results are shown in Appendix~\ref{sec:additional_result}, Figures~\ref{fig:tofu-met}--\ref{fig:tofu-met-fo-so}, Figures~\ref{fig:cls-met}--\ref{fig:cls-met-fo-so}.\looseness-1}
\label{fig:muse-book-met}
\end{figure}

\begin{figure}
    \centering
    \includegraphics[width=0.99\linewidth]{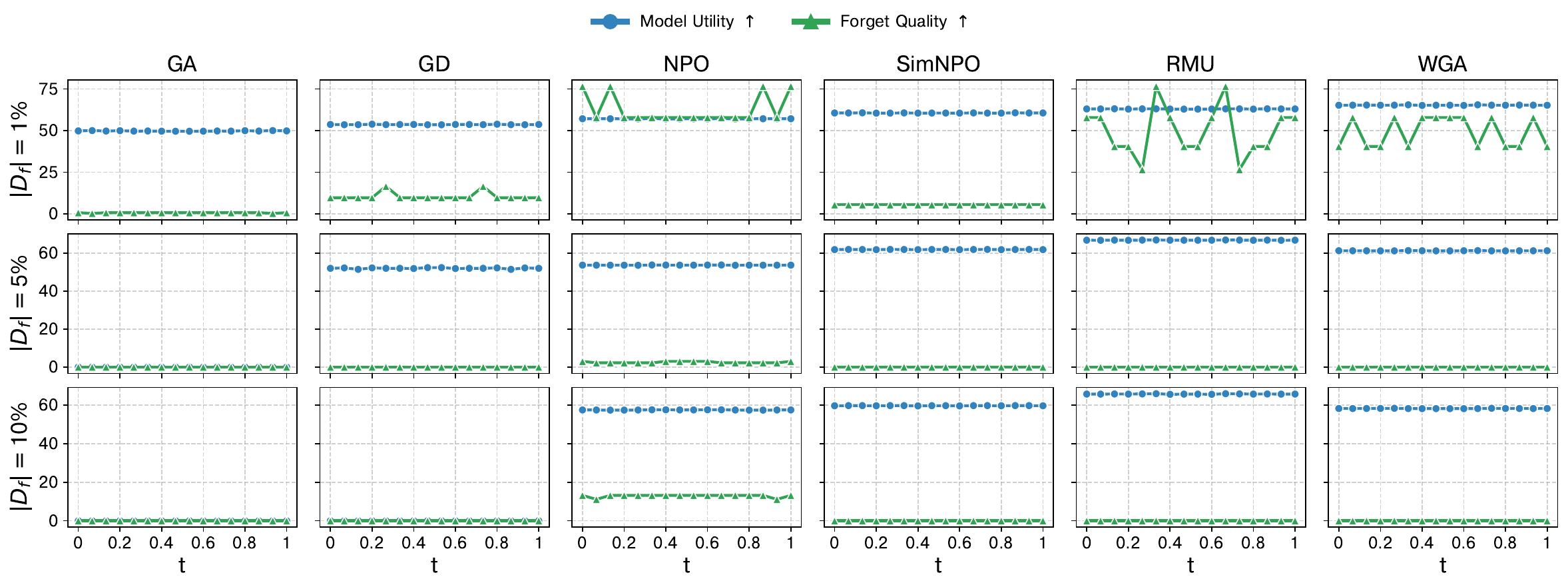}
    \caption{MCU under \textbf{Rand} setting on \textbf{TOFU dataset}.}
    \label{fig:tofu-rand}
\end{figure}

\begin{figure}
    \centering
    \includegraphics[width=0.99\linewidth]{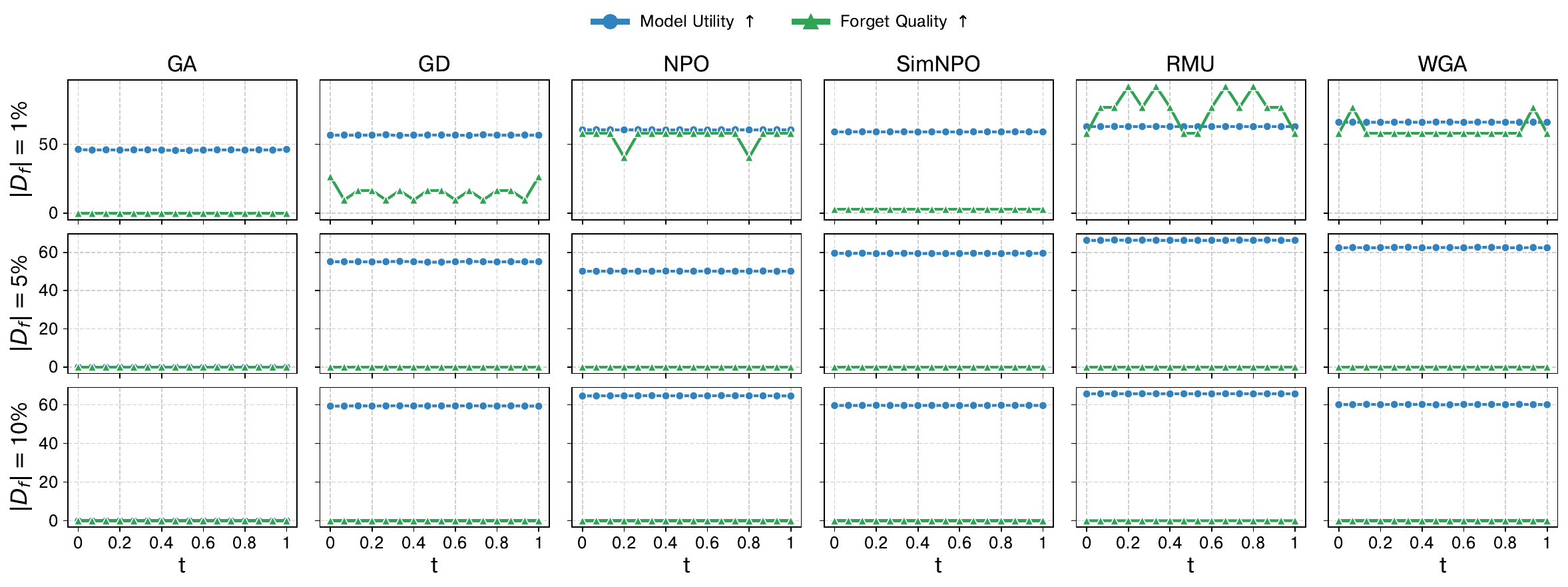}
    \caption{MCU under \textbf{Rand-CL} setting on \textbf{TOFU dataset}.}
    \label{fig:tofu-rand-cl}
\end{figure}

\begin{figure}
    \centering
    \includegraphics[width=0.99\linewidth]{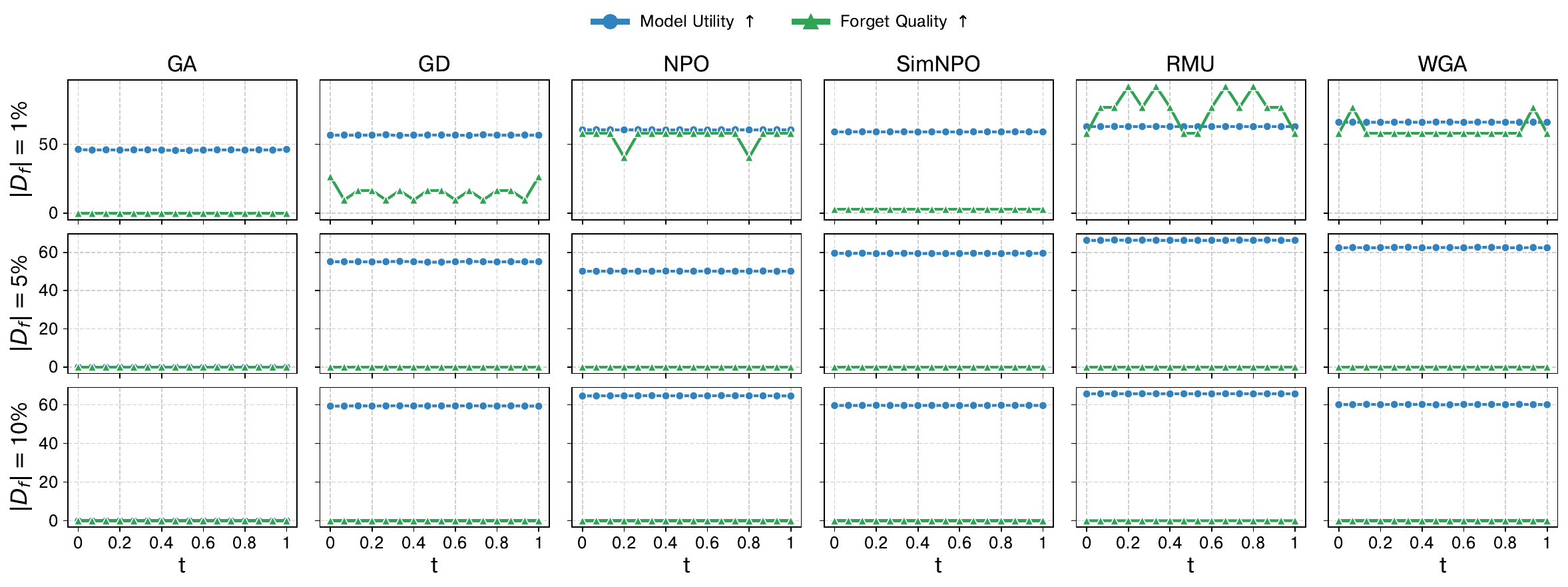}
    \caption{MCU under \textbf{Rand-SO} setting on \textbf{TOFU dataset}.}
    \label{fig:tofu-rand-so}
\end{figure}

\begin{figure}
    \centering
    \includegraphics[width=0.99\linewidth]{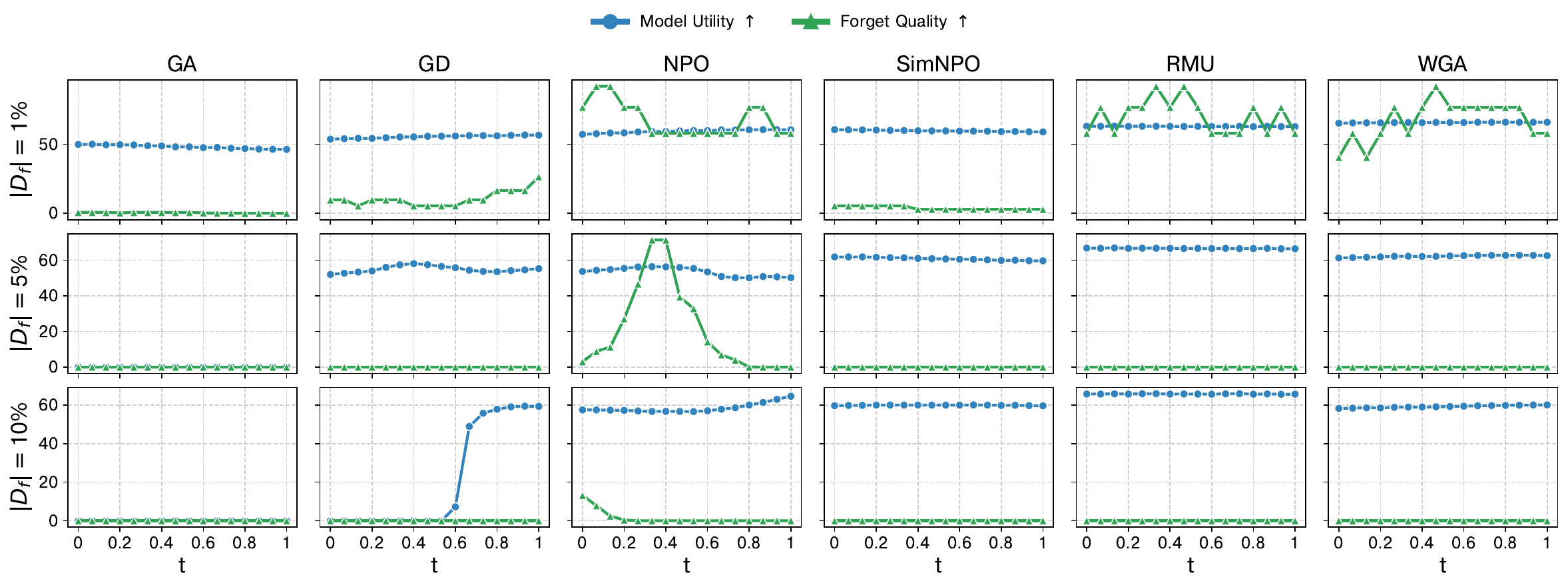}
    \caption{MCU under \textbf{CL-Non-CL} setting on \textbf{TOFU dataset}.}
    \label{fig:tofu-cl-non-cl}
\end{figure}

\begin{figure}
    \centering
    \includegraphics[width=0.99\linewidth]{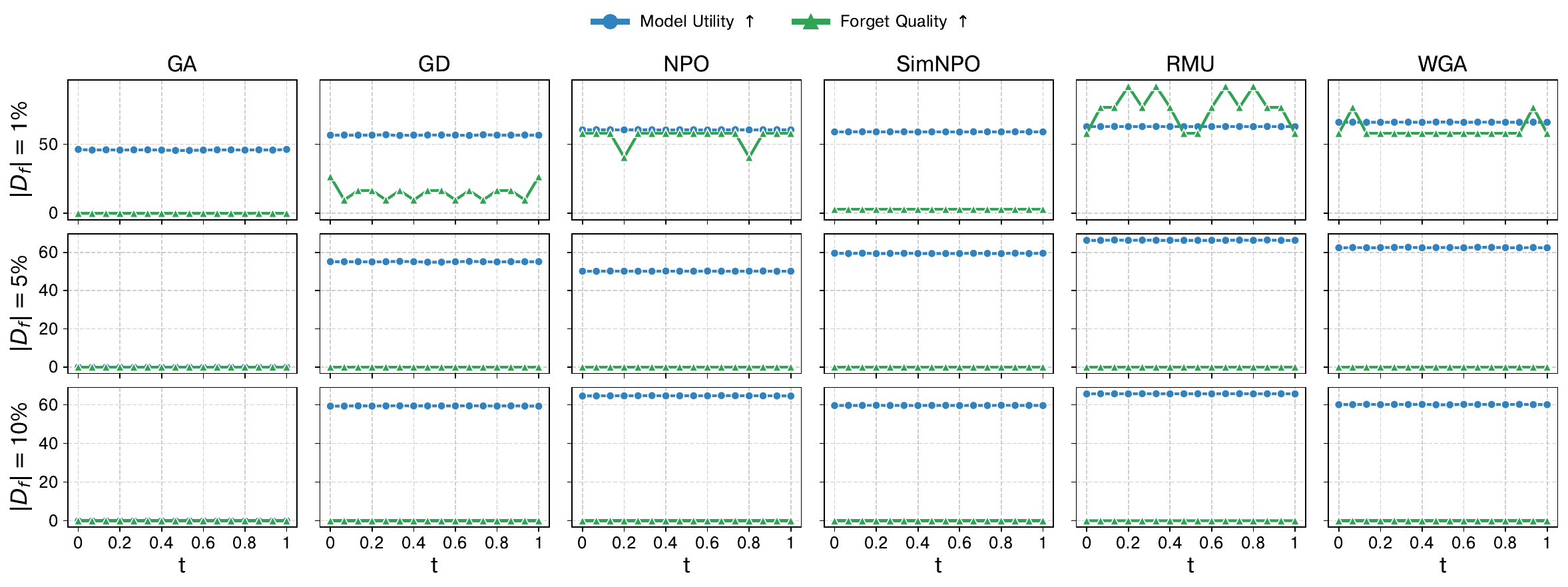}
    \caption{MCU under \textbf{FO-SO} setting on \textbf{TOFU dataset}.}
    \label{fig:tofu-fo-so}
\end{figure}

\begin{figure}
\begin{center}
    \begin{subfigure}{0.49\textwidth}
        \includegraphics[width=\linewidth]{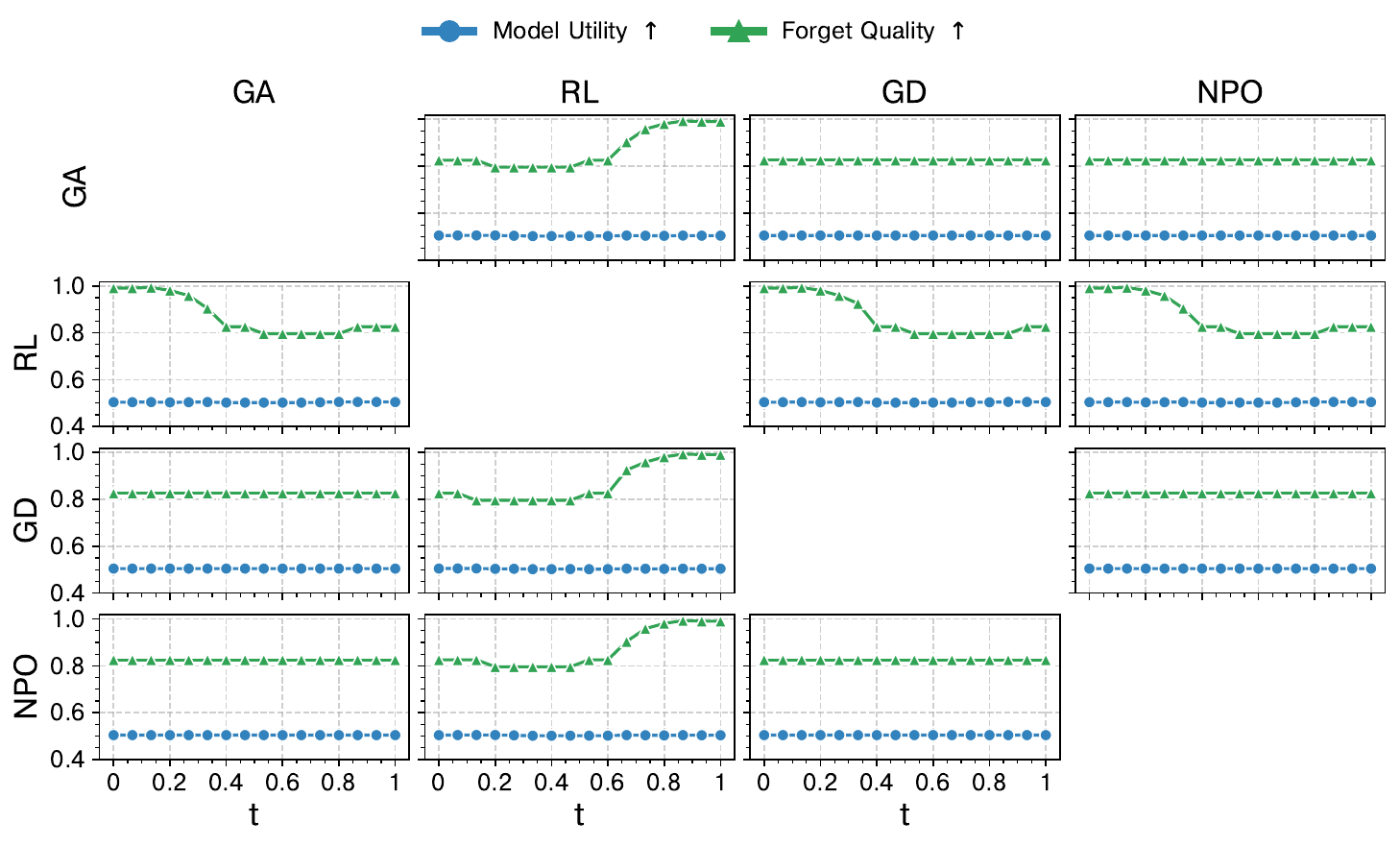}
        \caption{Linear MCU when $|D_f|=1\%$}
    \end{subfigure}
    \begin{subfigure}{0.49\textwidth}
        \includegraphics[width=\linewidth]{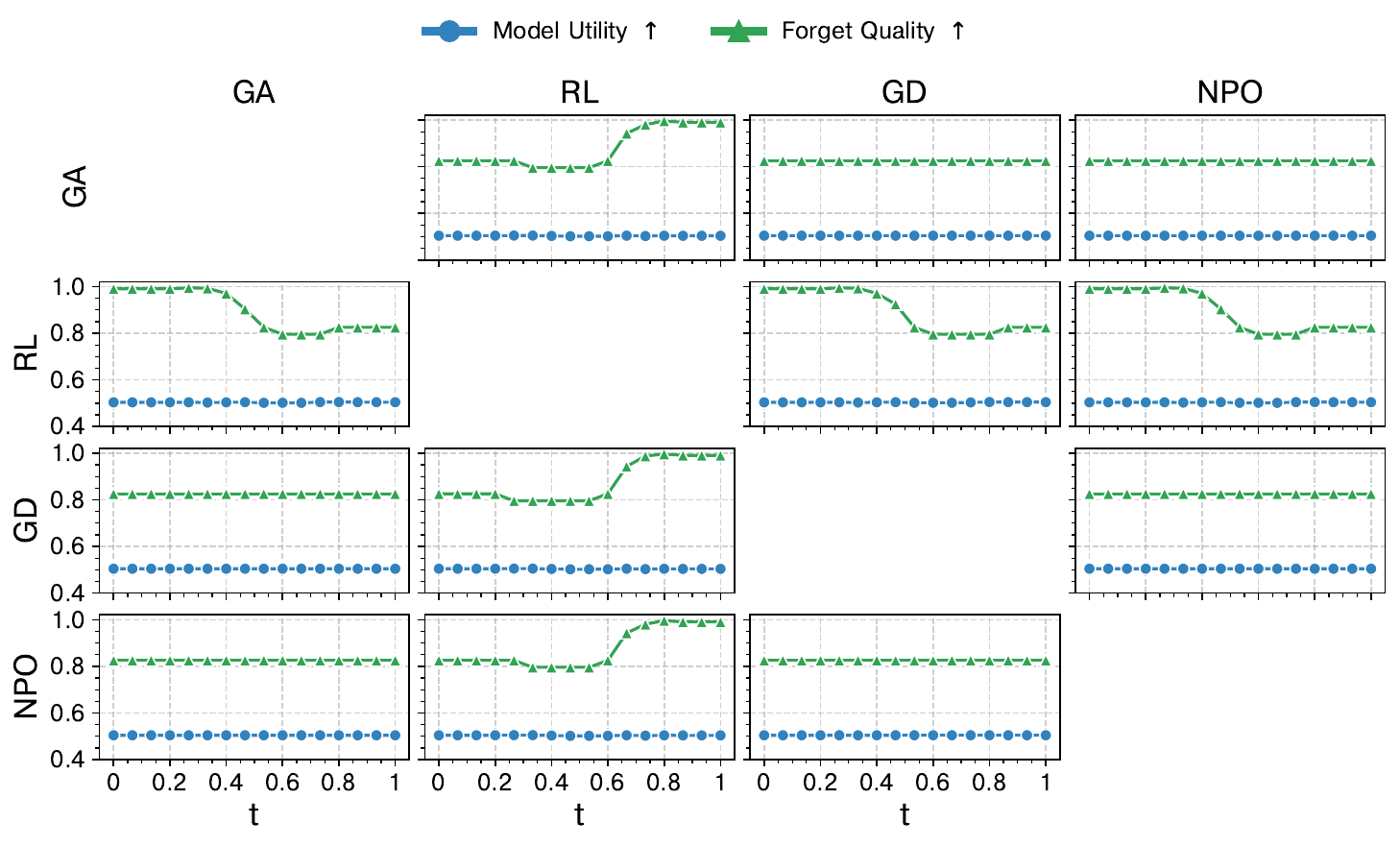}
        \caption{Quadratic MCU when $|D_f|=1\%$}
    \end{subfigure}
    \begin{subfigure}{0.49\textwidth}
        \centering
        \includegraphics[width=\linewidth]{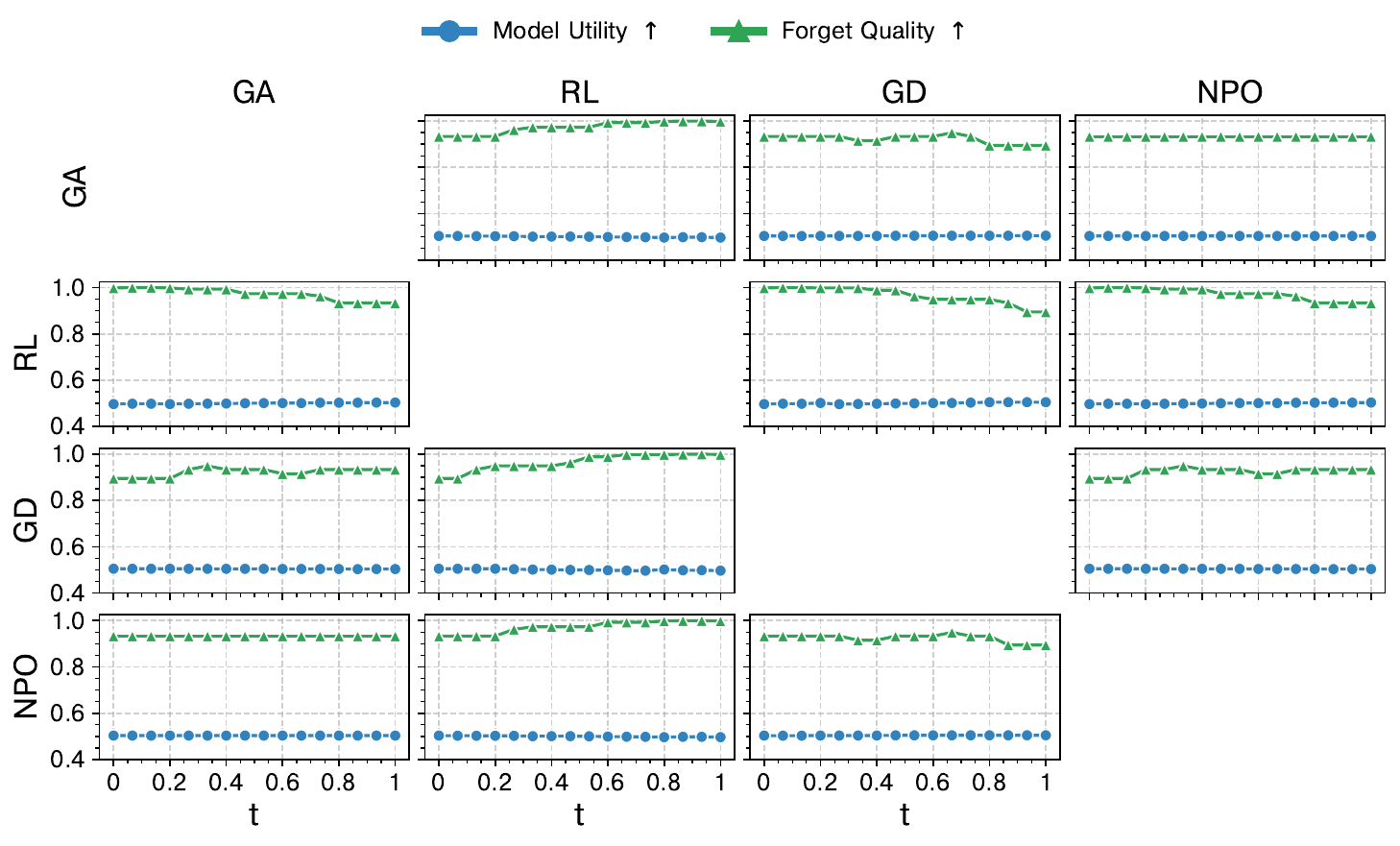}
        \caption{Linear MCU when $|D_f|=5\%$}
    \end{subfigure}
    \begin{subfigure}{0.49\textwidth}
        \centering
        \includegraphics[width=\linewidth]{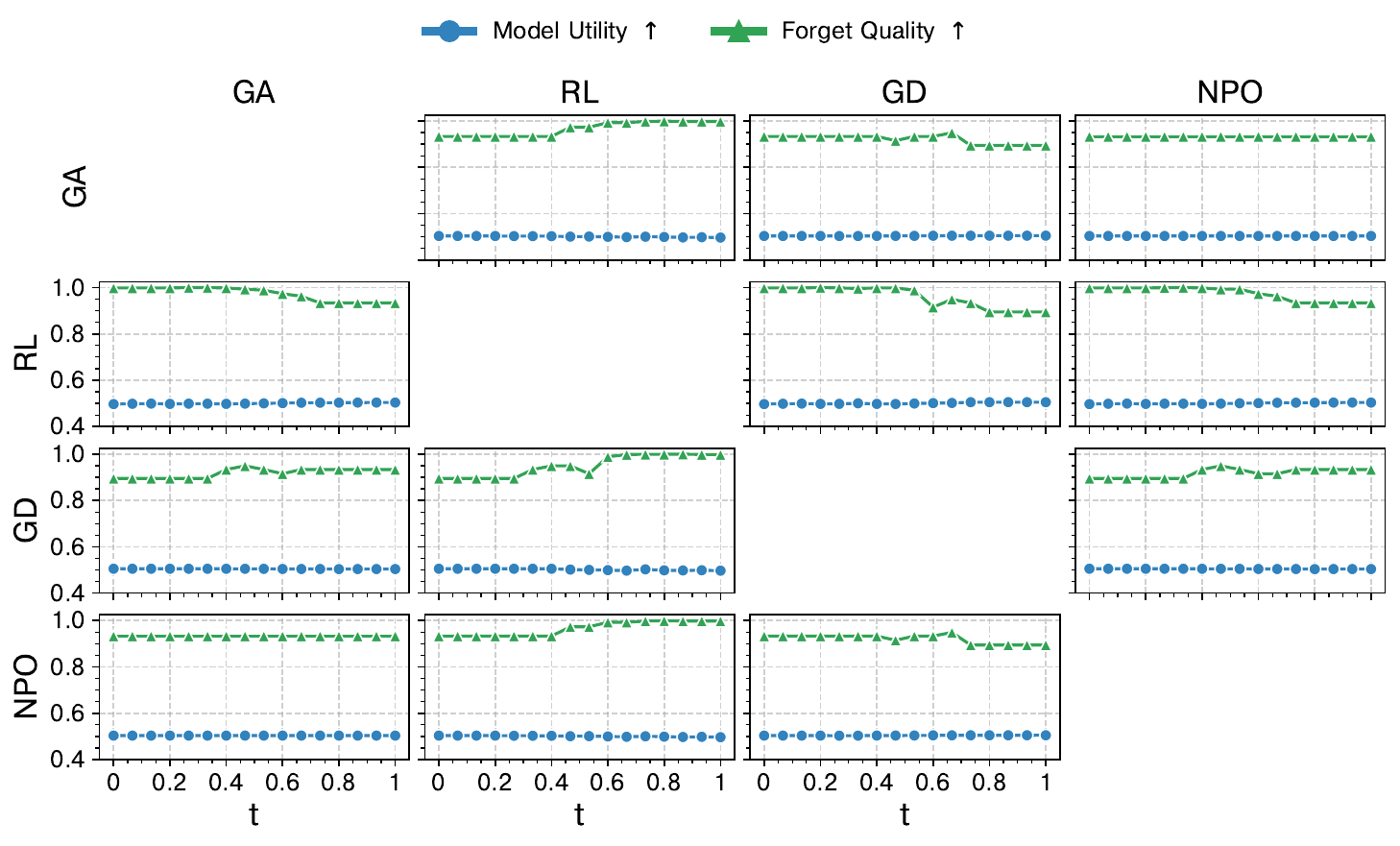}
        \caption{Quadratic MCU when $|D_f|=5\%$}
    \end{subfigure}
    \begin{subfigure}{0.49\textwidth}
        \centering
        \includegraphics[width=\linewidth]{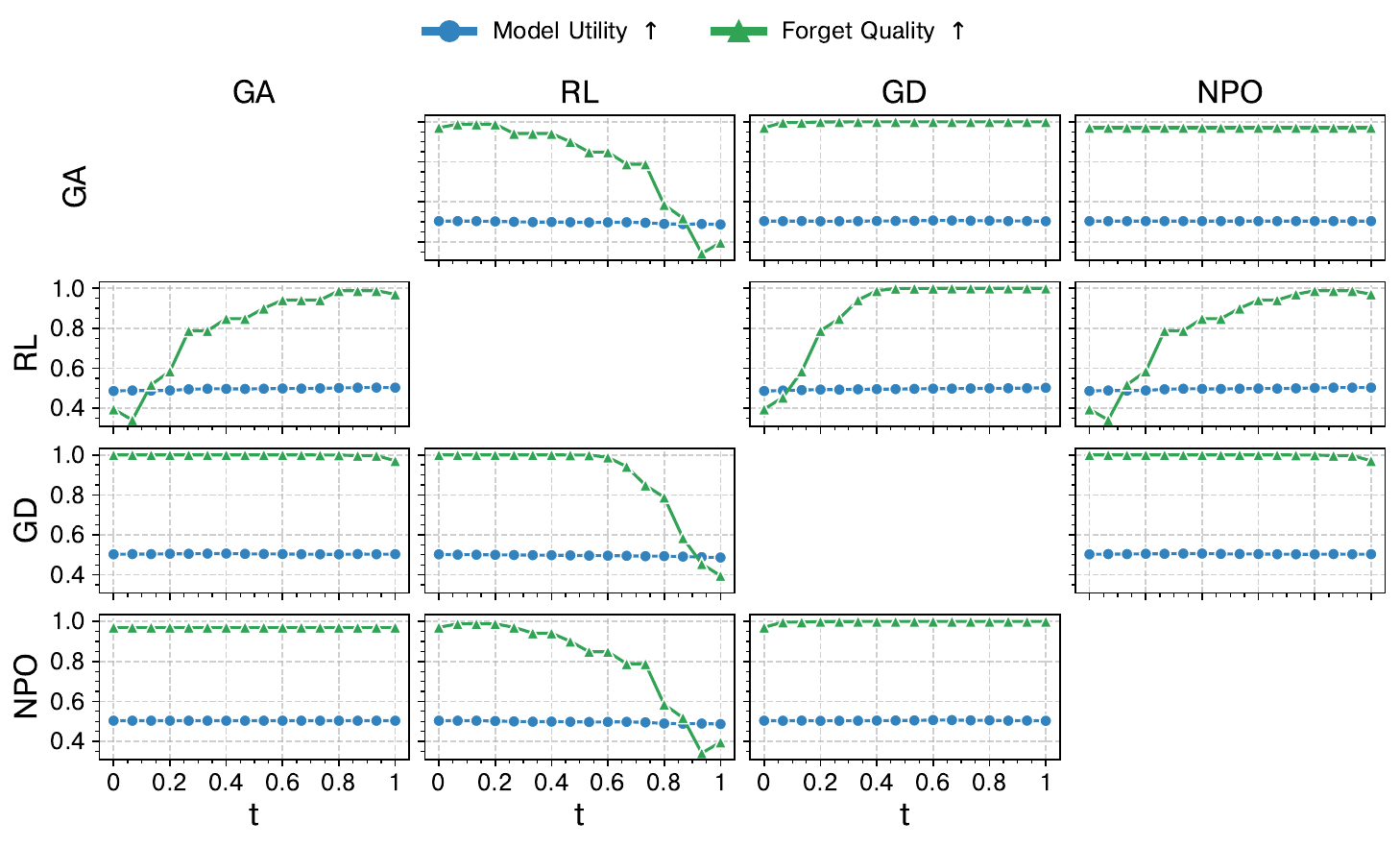}
        \caption{Linear MCU when $|D_f|=10\%$}
    \end{subfigure}
    \begin{subfigure}{0.49\textwidth}
        \centering
        \includegraphics[width=\linewidth]{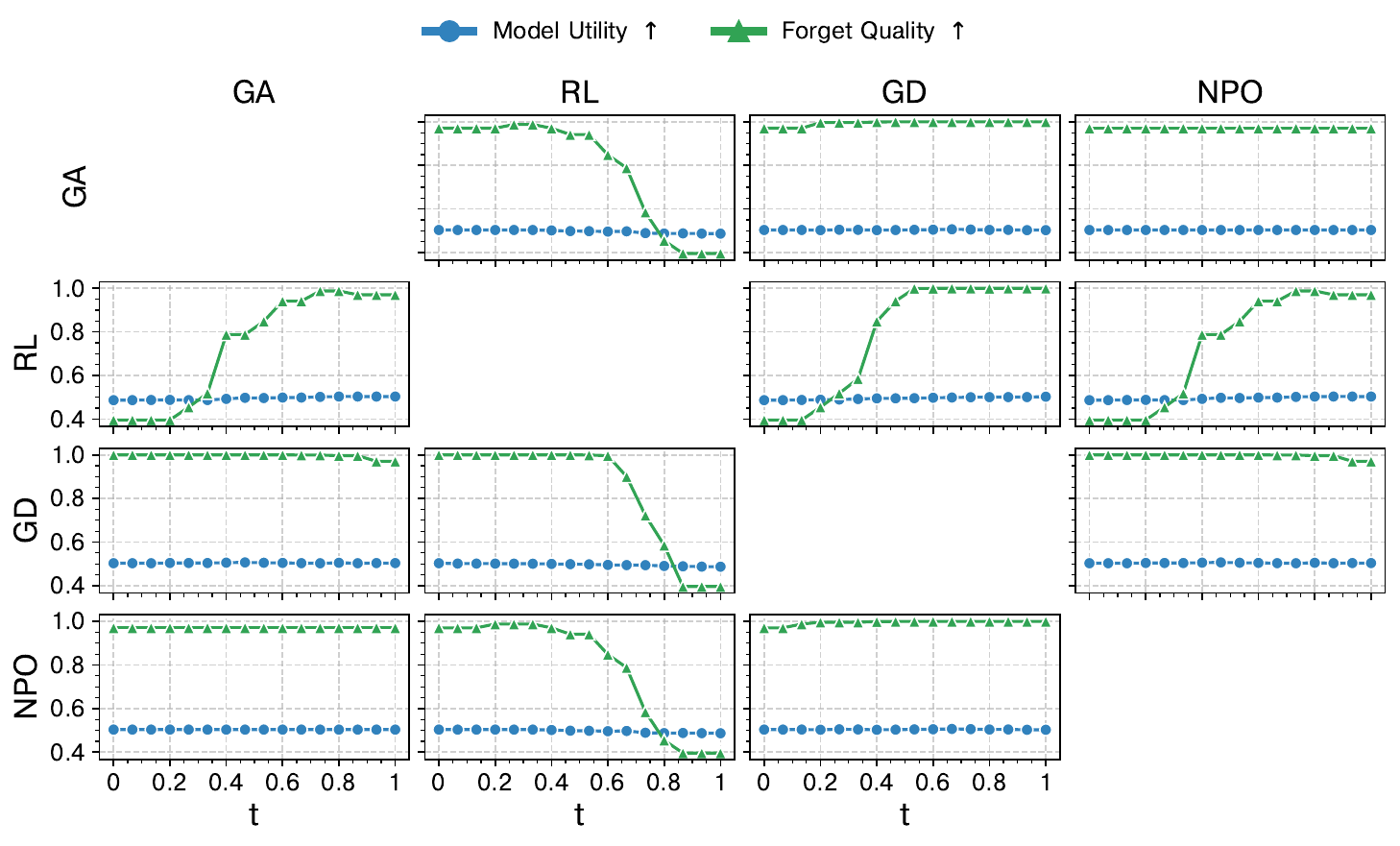}
        \caption{Quadratic MCU when $|D_f|=10\%$}
    \end{subfigure}
\end{center}
\caption{MCU under \textbf{Met} setting on \textbf{TOFU dataset}.}
\label{fig:tofu-met}
\end{figure}

\begin{figure}
\begin{center}
    \begin{subfigure}{0.49\textwidth}
        \includegraphics[width=\linewidth]{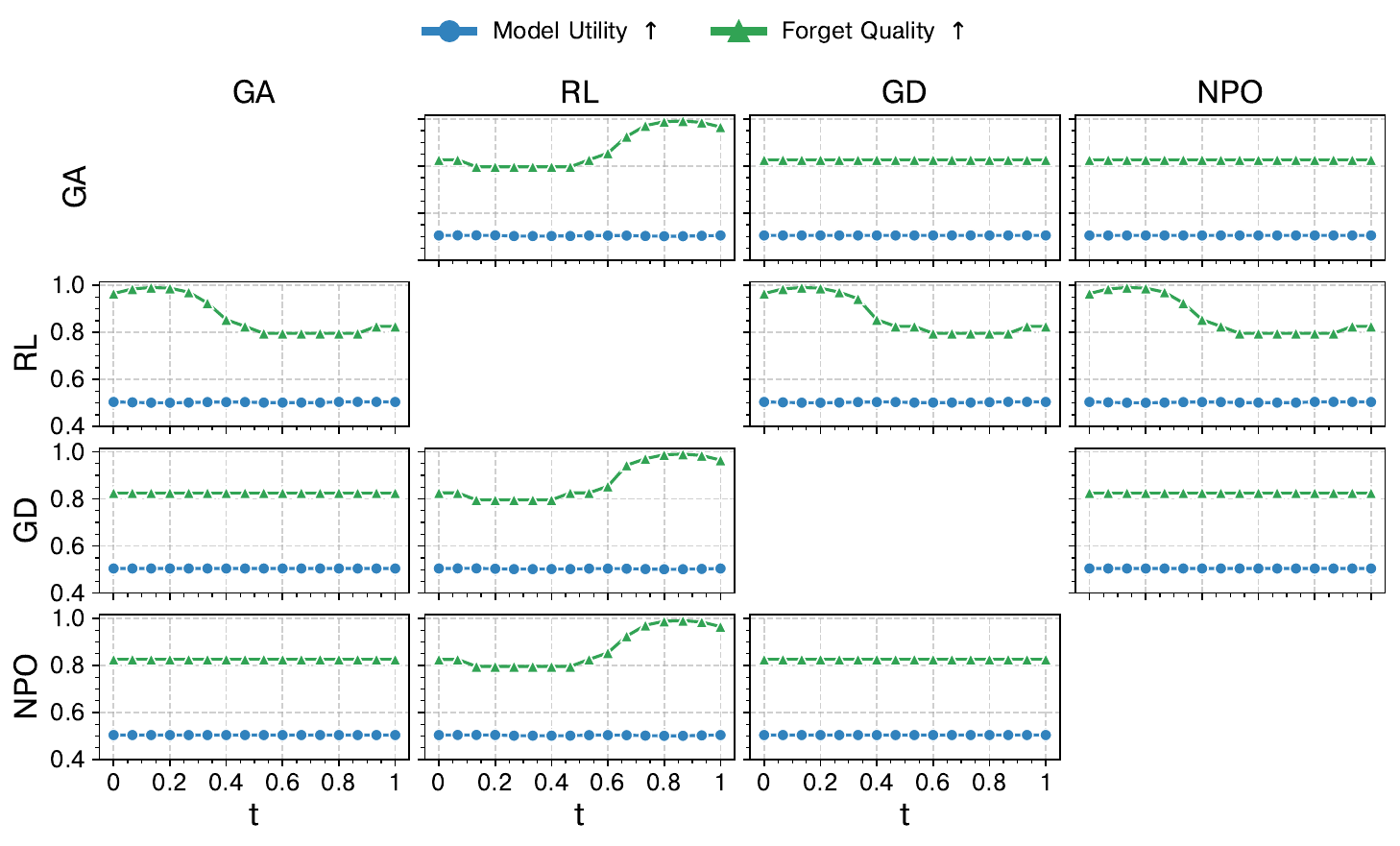}
        \caption{Linear MCU when $|D_f|=1\%$}
    \end{subfigure}
    \begin{subfigure}{0.49\textwidth}
        \includegraphics[width=\linewidth]{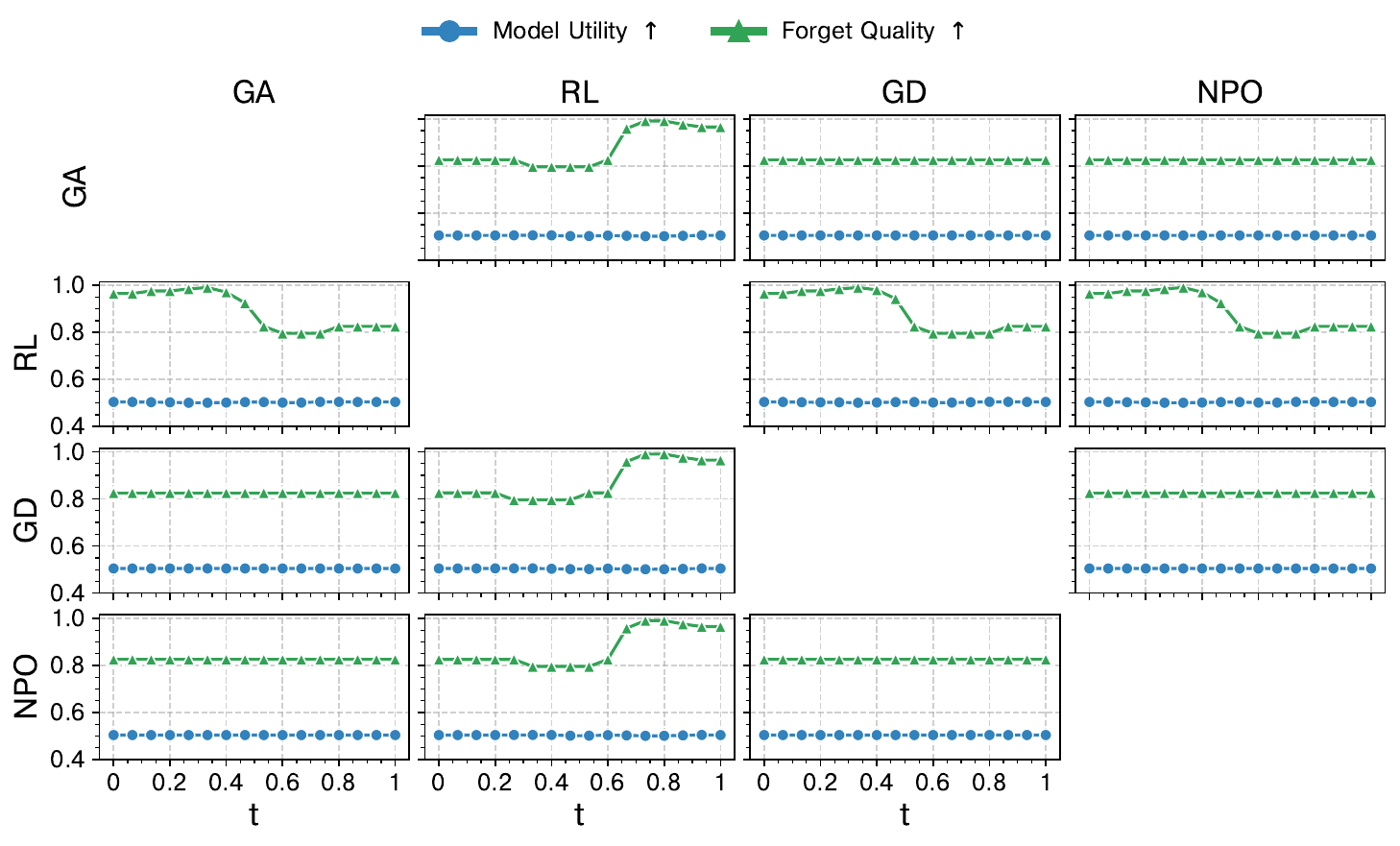}
        \caption{Bezier MCU when $|D_f|=1\%$}
    \end{subfigure}
    \begin{subfigure}{0.49\textwidth}
        \centering
        \includegraphics[width=\linewidth]{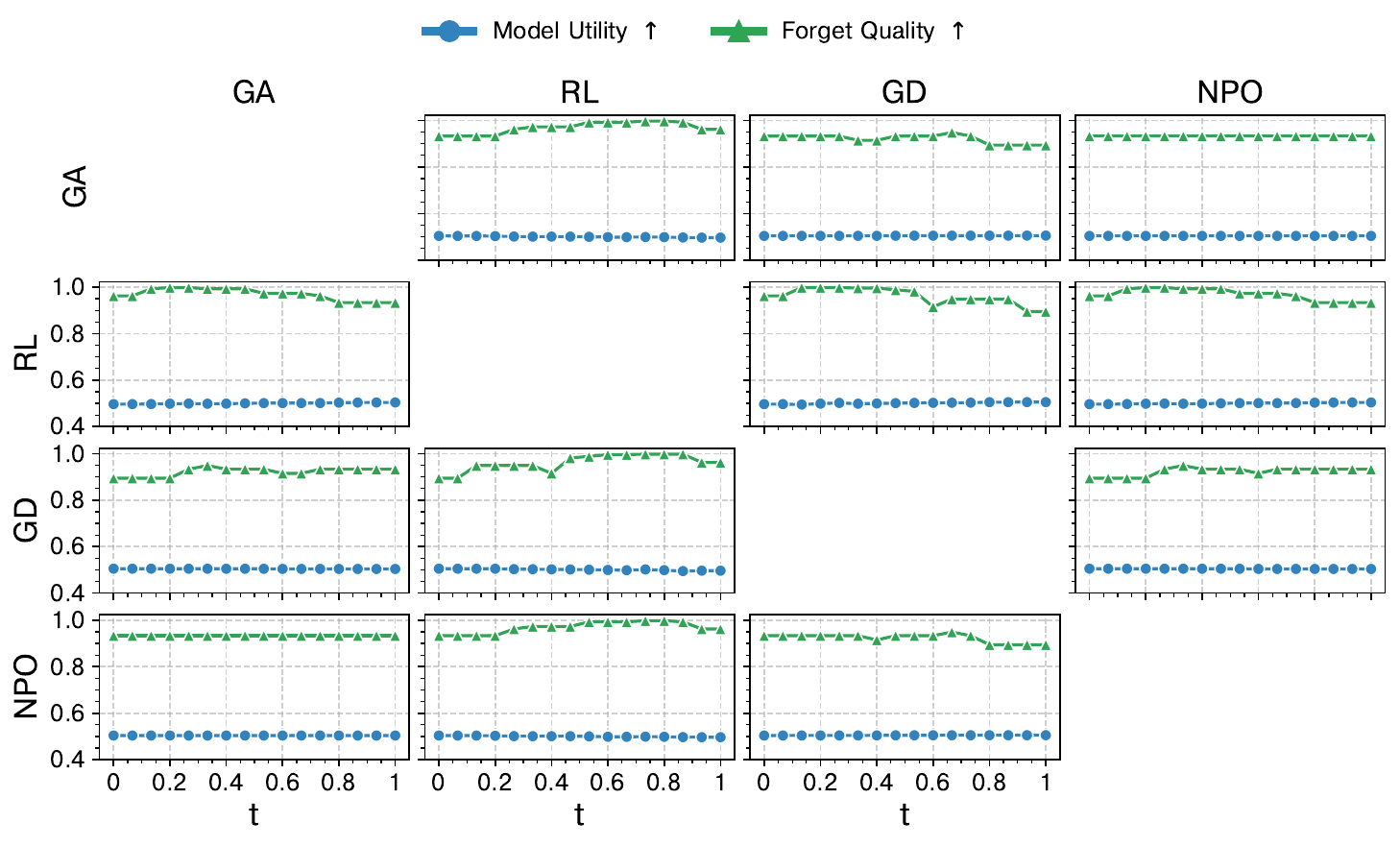}
        \caption{Linear MCU when $|D_f|=5\%$}
    \end{subfigure}
    \begin{subfigure}{0.49\textwidth}
        \centering
        \includegraphics[width=\linewidth]{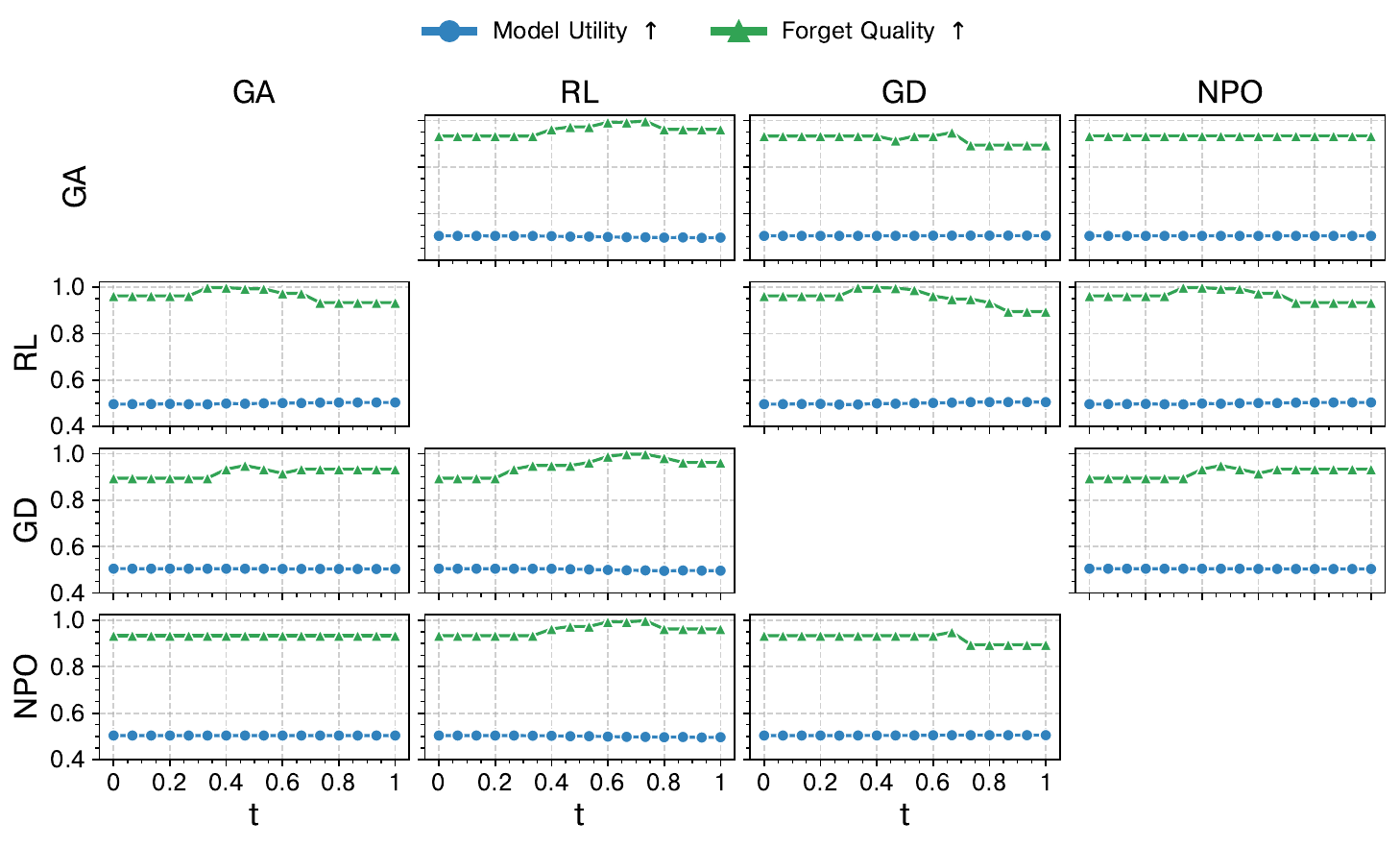}
        \caption{Bezier MCU when $|D_f|=5\%$}
    \end{subfigure}
    \begin{subfigure}{0.49\textwidth}
        \centering
        \includegraphics[width=\linewidth]{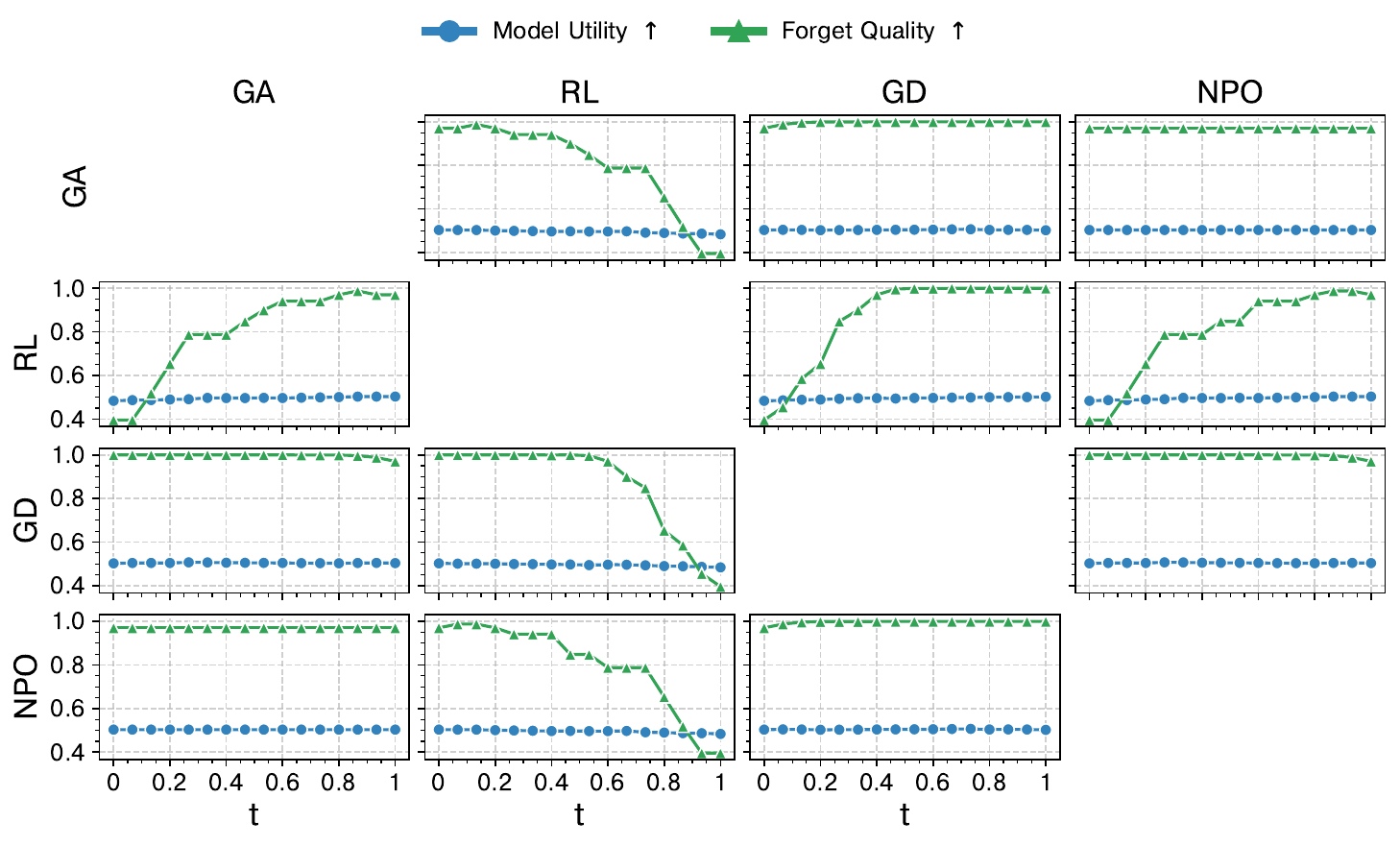}
        \caption{Linear MCU when $|D_f|=10\%$}
    \end{subfigure}
    \begin{subfigure}{0.49\textwidth}
        \centering
        \includegraphics[width=\linewidth]{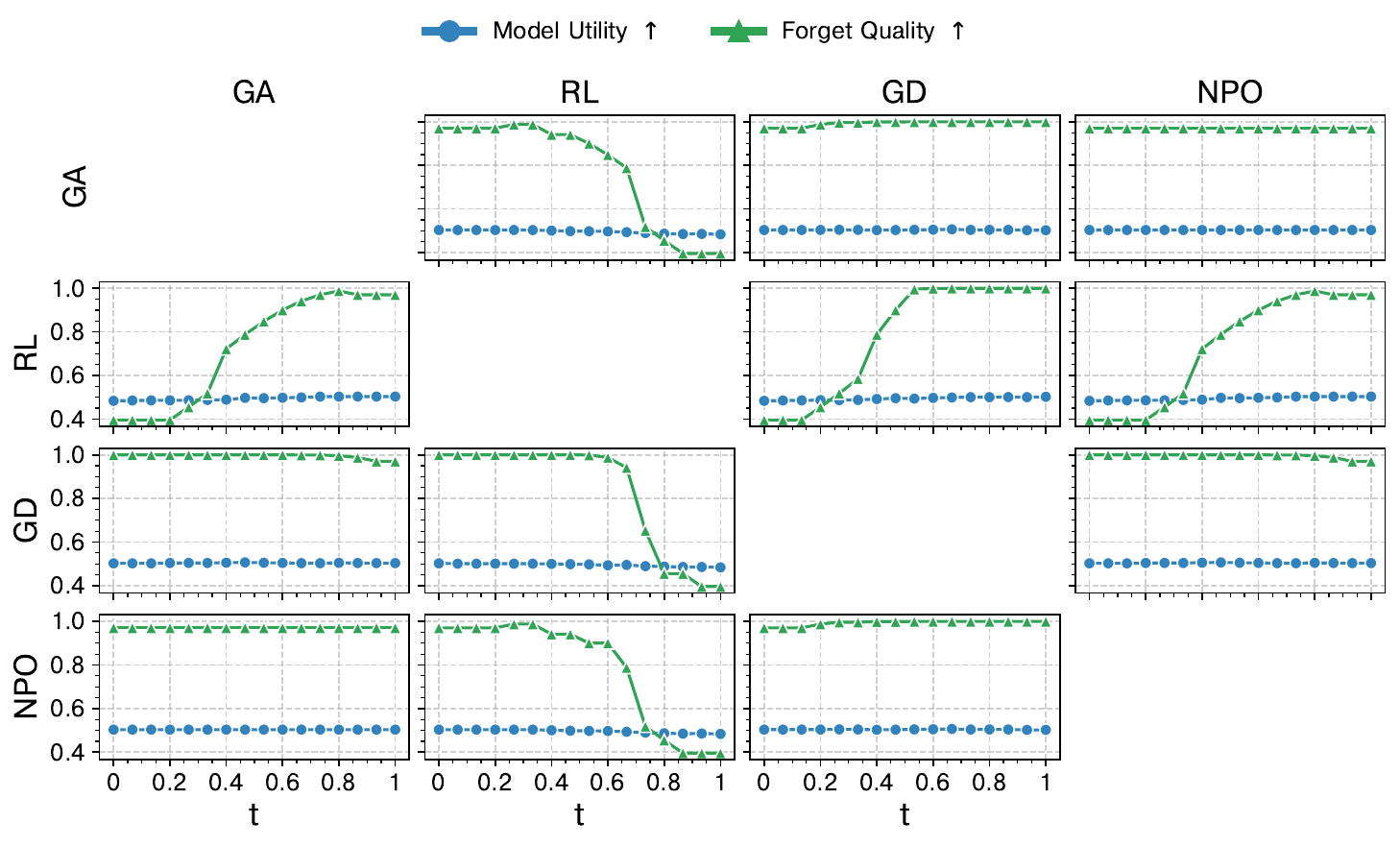}
        \caption{Bezier MCU when $|D_f|=10\%$}
    \end{subfigure}
\end{center}
\caption{MCU under \textbf{Met-CL} setting on \textbf{TOFU dataset}.}
\label{fig:tofu-met-cl}
\end{figure}

\begin{figure}
\begin{center}
    \begin{subfigure}{0.49\textwidth}
        \includegraphics[width=\linewidth]{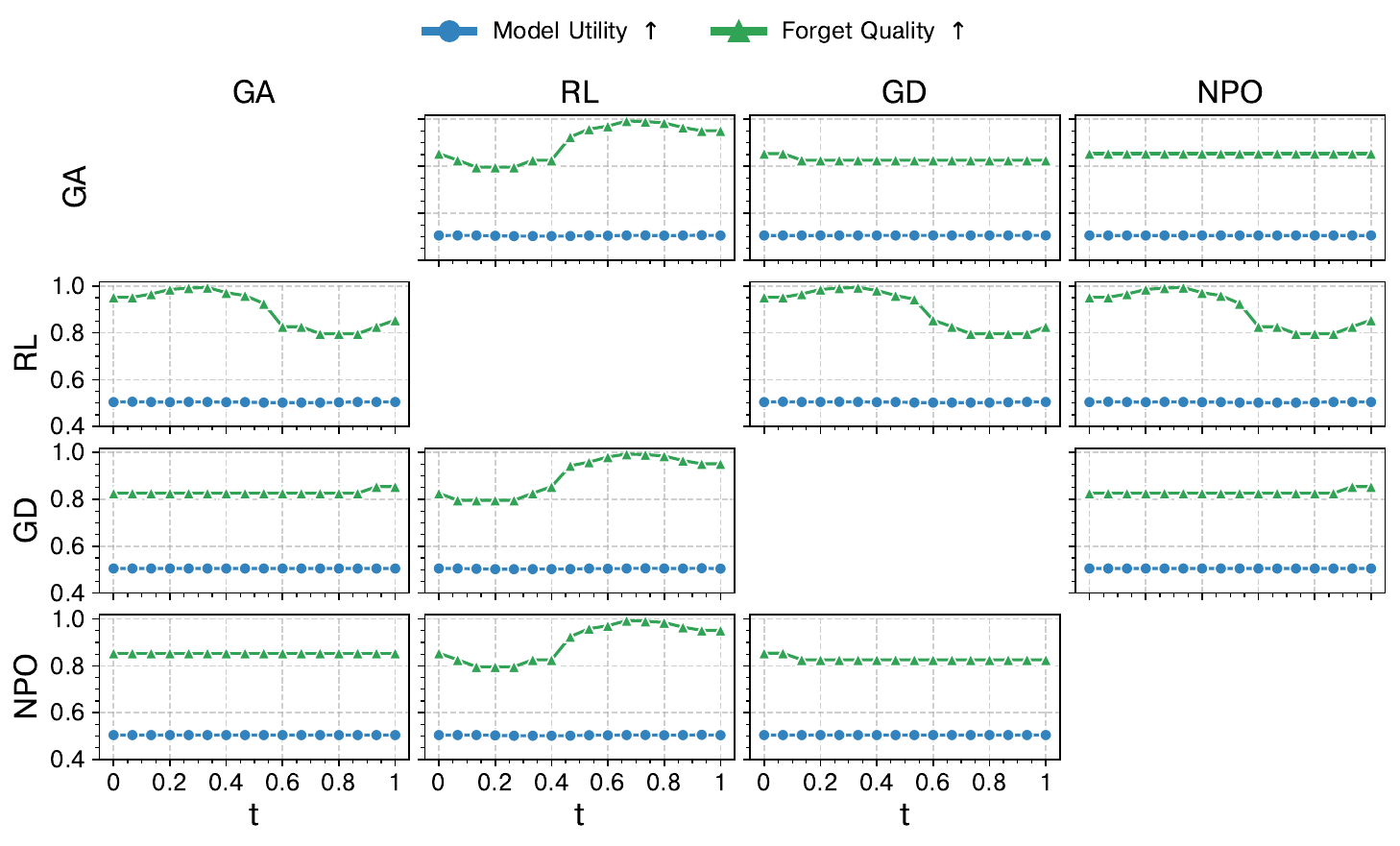}
        \caption{Linear MCU when $|D_f|=1\%$}
    \end{subfigure}
    \begin{subfigure}{0.49\textwidth}
        \includegraphics[width=\linewidth]{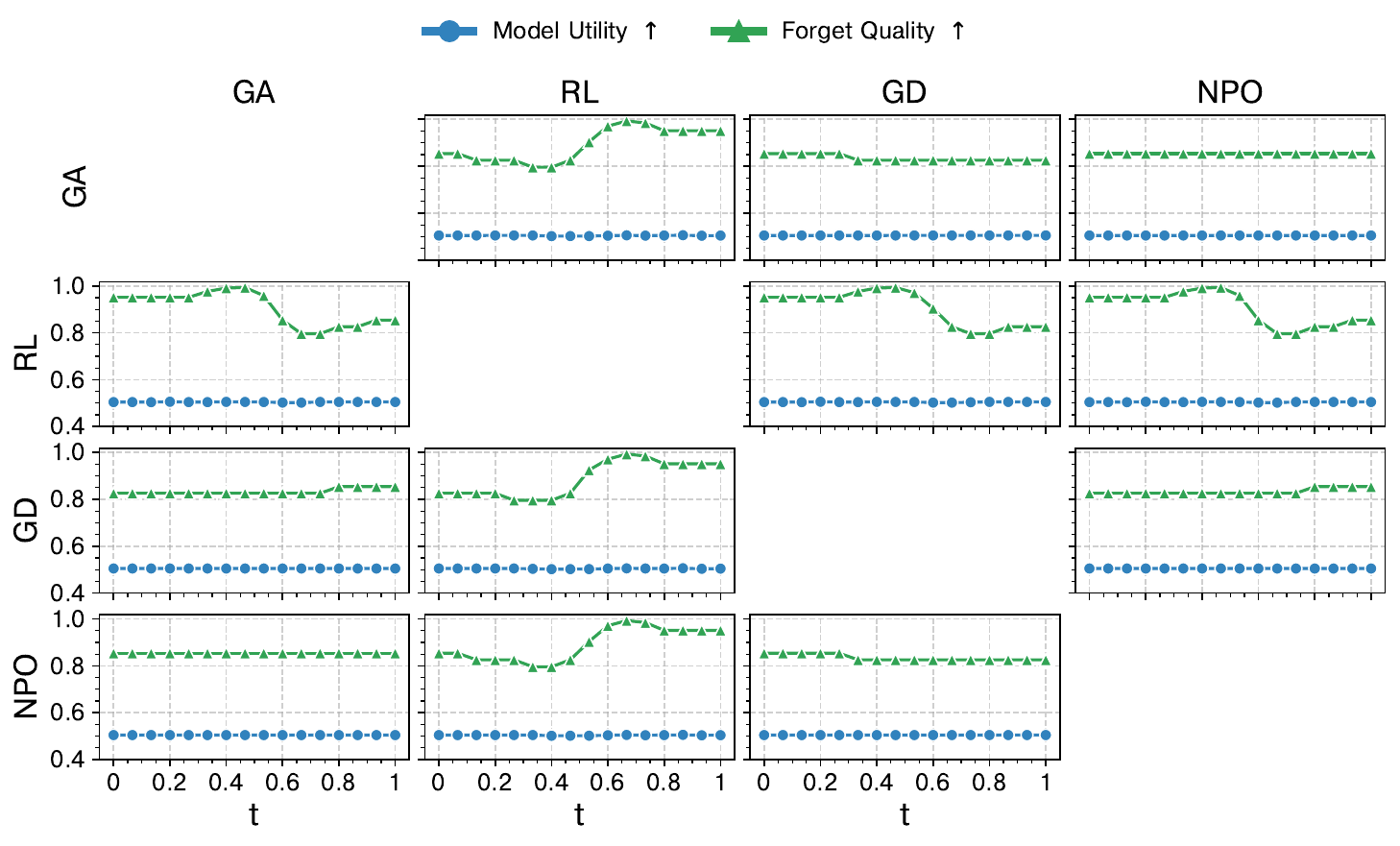}
        \caption{Bezier MCU when $|D_f|=1\%$}
    \end{subfigure}
    \begin{subfigure}{0.49\textwidth}
        \centering
        \includegraphics[width=\linewidth]{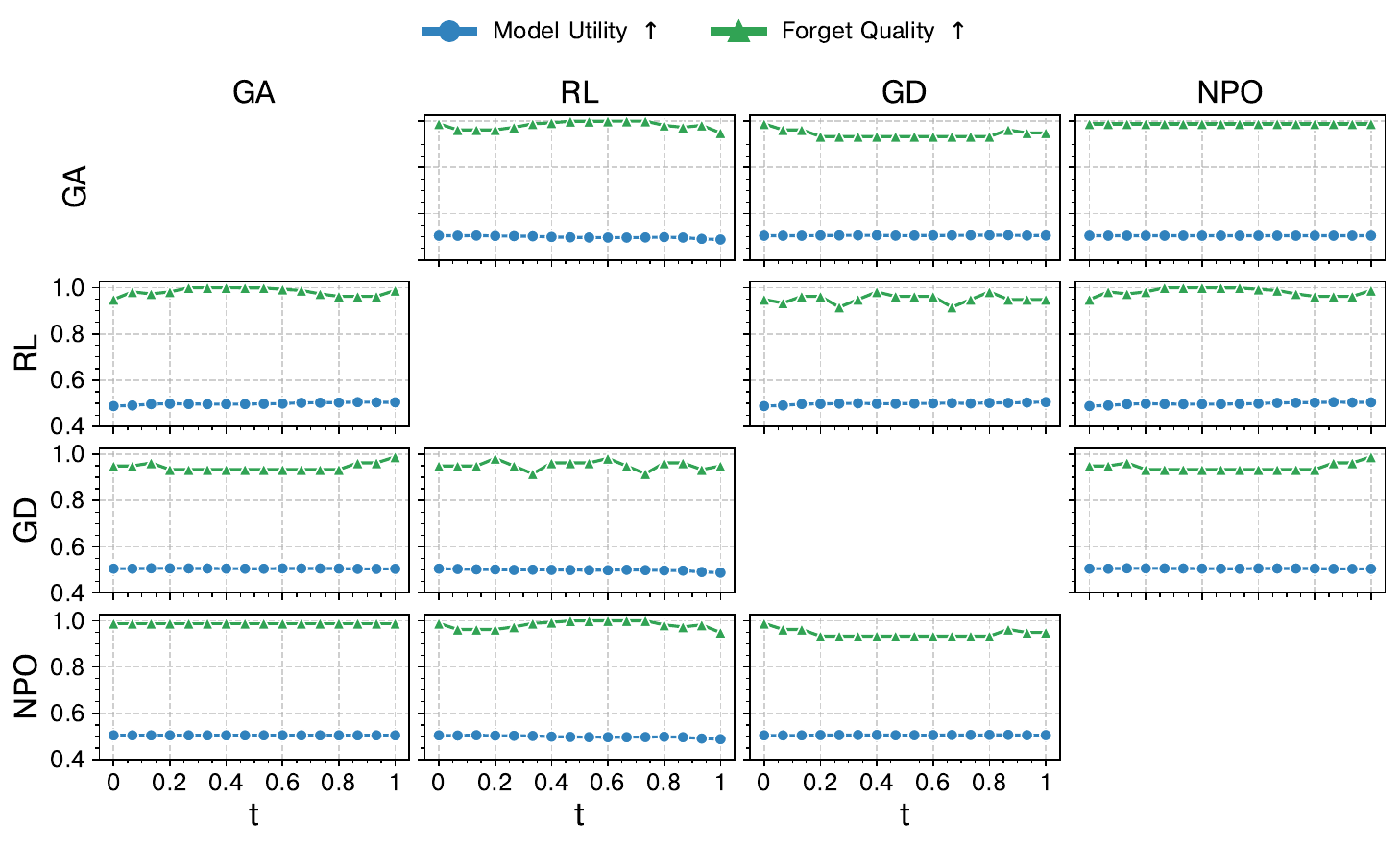}
        \caption{Linear MCU when $|D_f|=5\%$}
    \end{subfigure}
    \begin{subfigure}{0.49\textwidth}
        \centering
        \includegraphics[width=\linewidth]{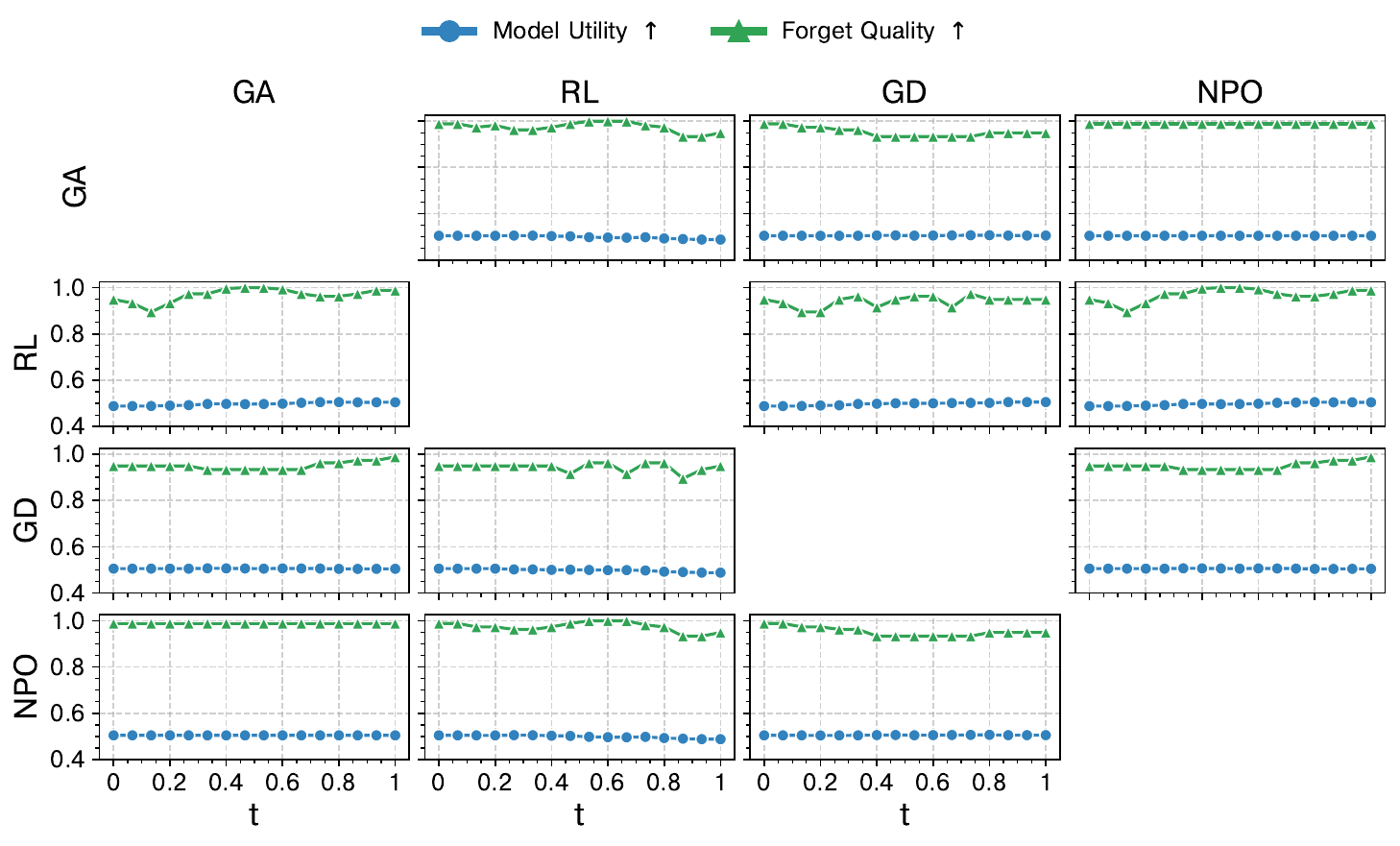}
        \caption{Bezier MCU when $|D_f|=5\%$}
    \end{subfigure}
    \begin{subfigure}{0.49\textwidth}
        \centering
        \includegraphics[width=\linewidth]{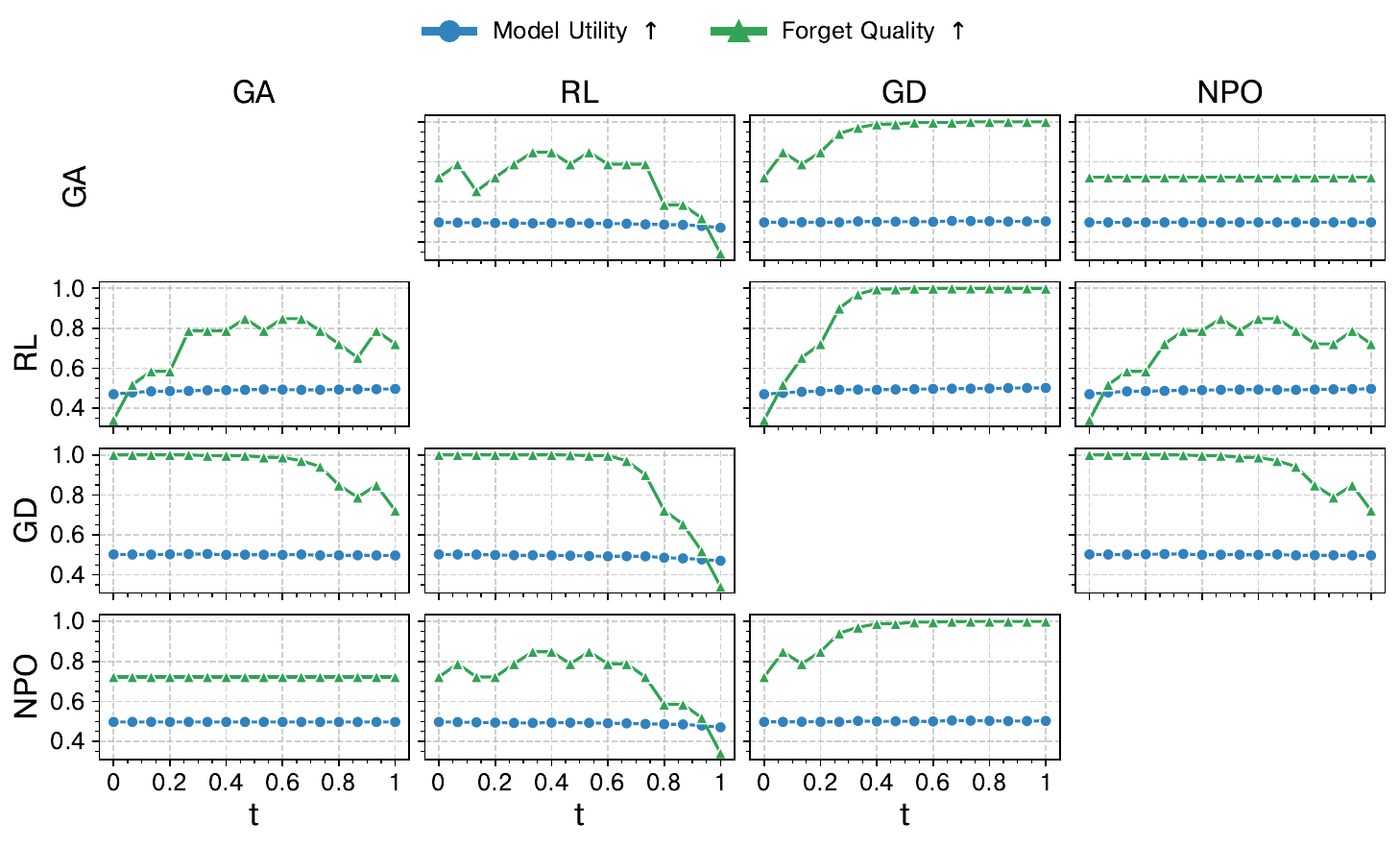}
        \caption{Linear MCU when $|D_f|=10\%$}
    \end{subfigure}
    \begin{subfigure}{0.49\textwidth}
        \centering
        \includegraphics[width=\linewidth]{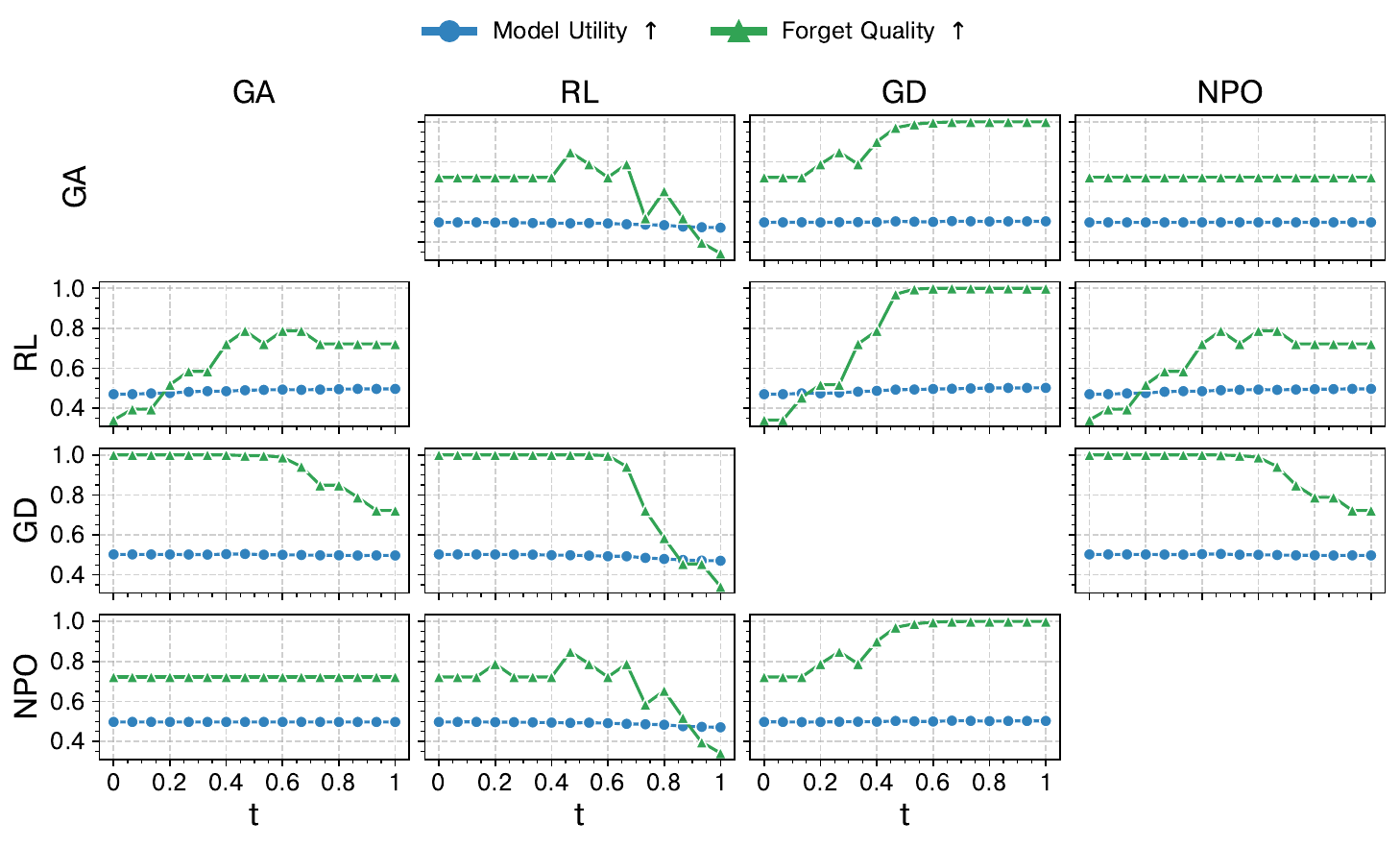}
        \caption{Bezier MCU when $|D_f|=10\%$}
    \end{subfigure}
\end{center}
\caption{MCU under \textbf{Met-SO} setting on \textbf{TOFU dataset}.}
\label{fig:tofu-met-so}
\end{figure}

\begin{figure}
\begin{center}
    \begin{subfigure}{0.49\textwidth}
        \includegraphics[width=\linewidth]{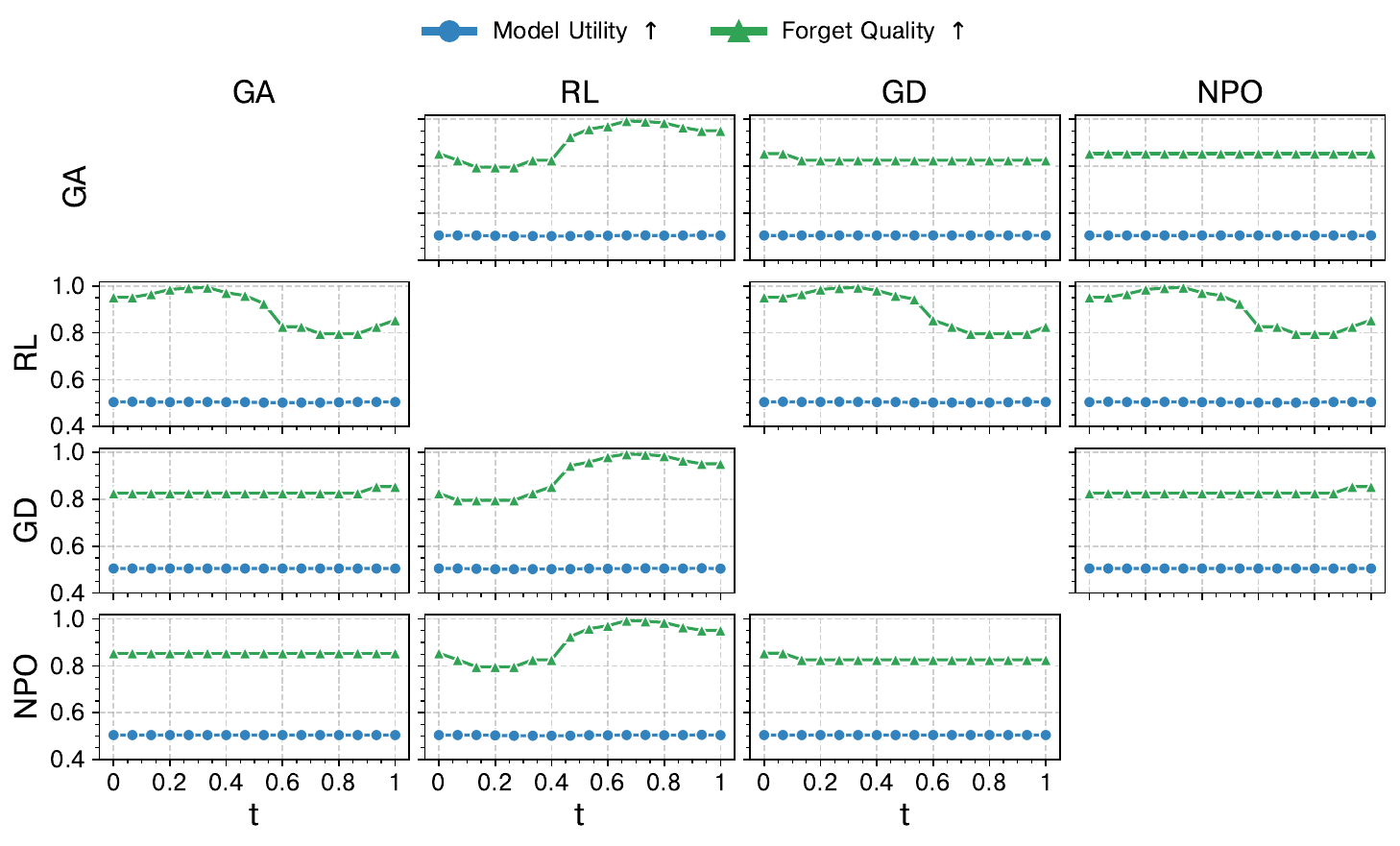}
        \caption{Linear MCU when $|D_f|=1\%$}
    \end{subfigure}
    \begin{subfigure}{0.49\textwidth}
        \includegraphics[width=\linewidth]{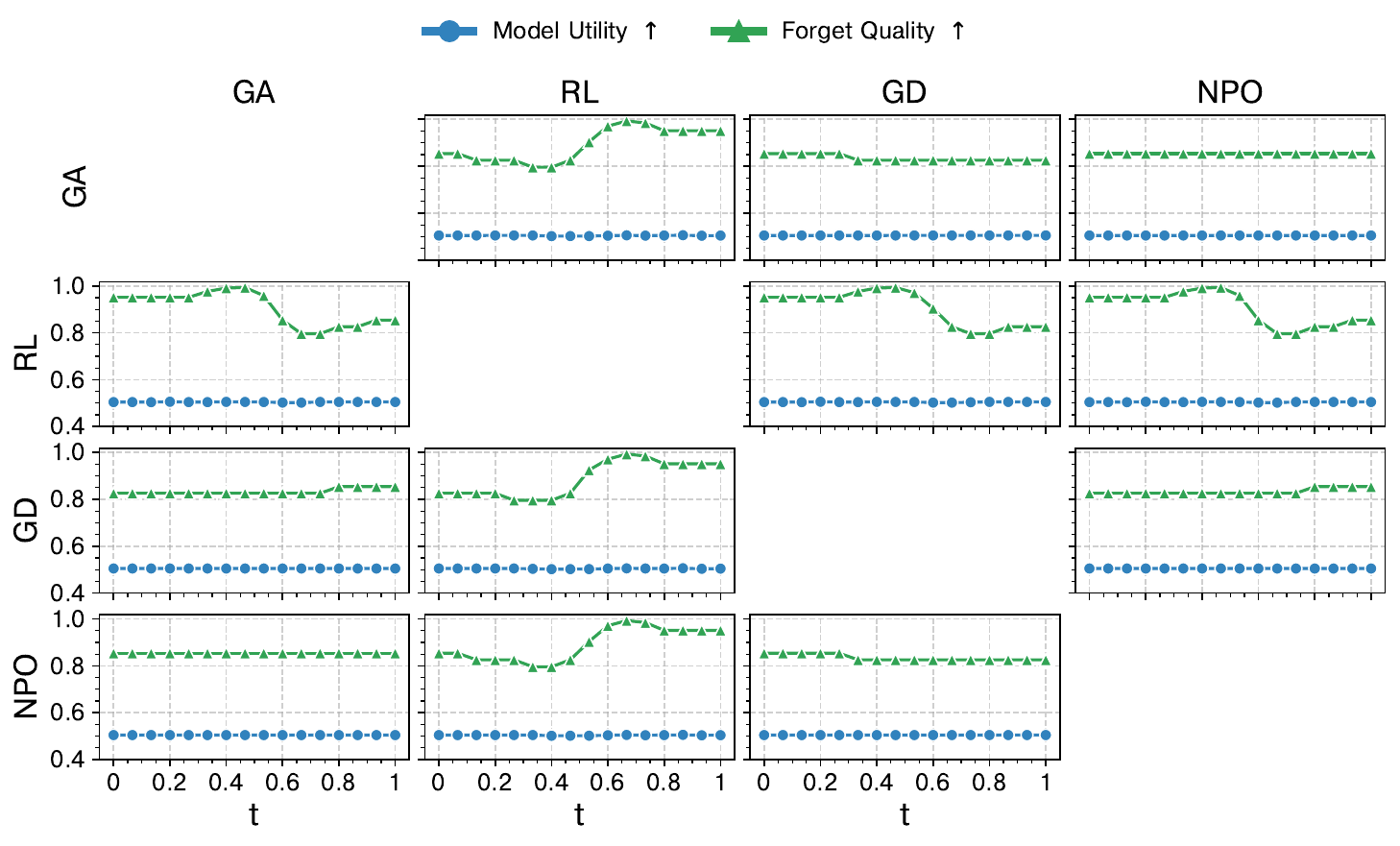}
        \caption{Bezier MCU when $|D_f|=1\%$}
    \end{subfigure}
    \begin{subfigure}{0.49\textwidth}
        \centering
        \includegraphics[width=\linewidth]{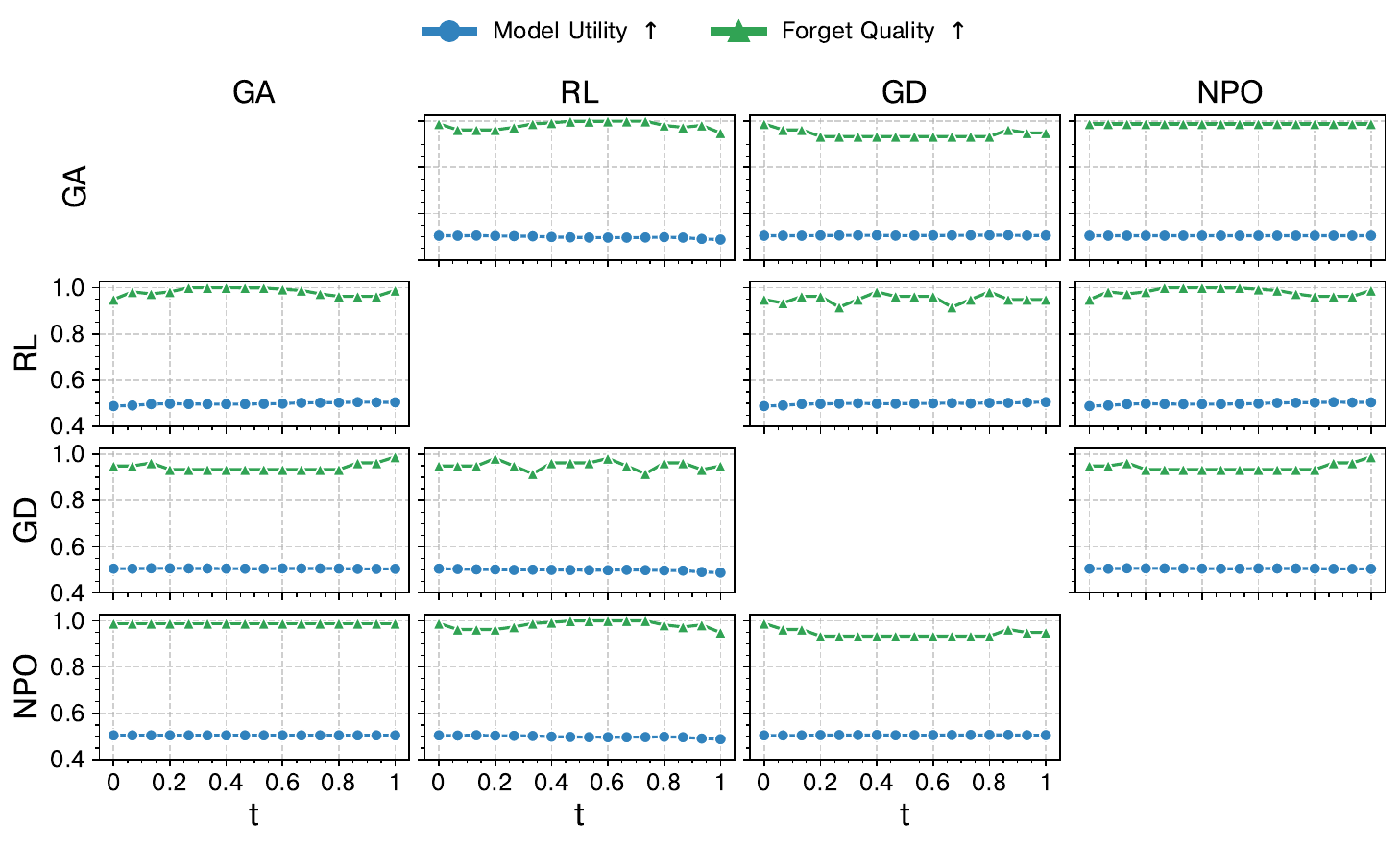}
        \caption{Linear MCU when $|D_f|=5\%$}
    \end{subfigure}
    \begin{subfigure}{0.49\textwidth}
        \centering
        \includegraphics[width=\linewidth]{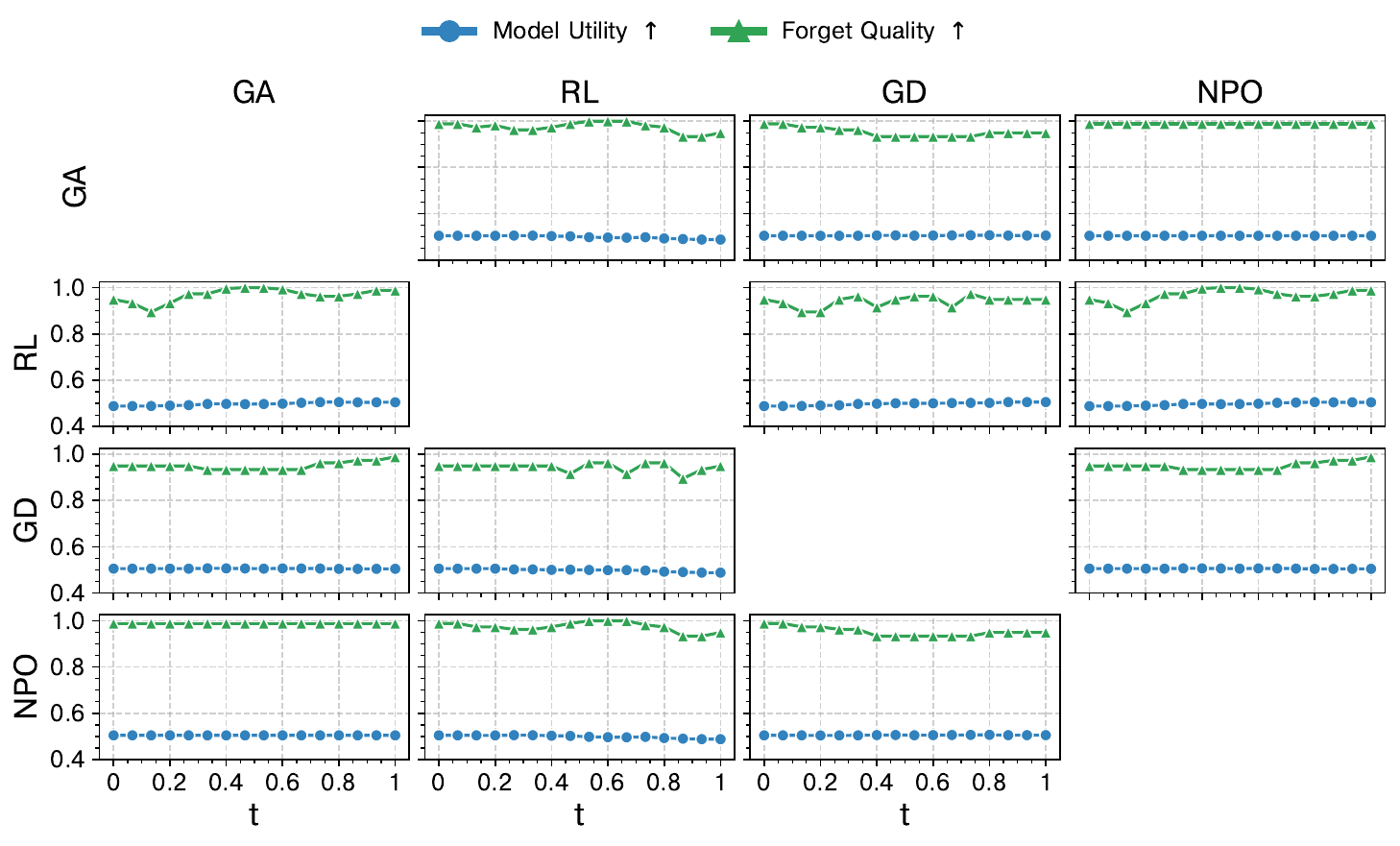}
        \caption{Bezier MCU when $|D_f|=5\%$}
    \end{subfigure}
    \begin{subfigure}{0.49\textwidth}
        \centering
        \includegraphics[width=\linewidth]{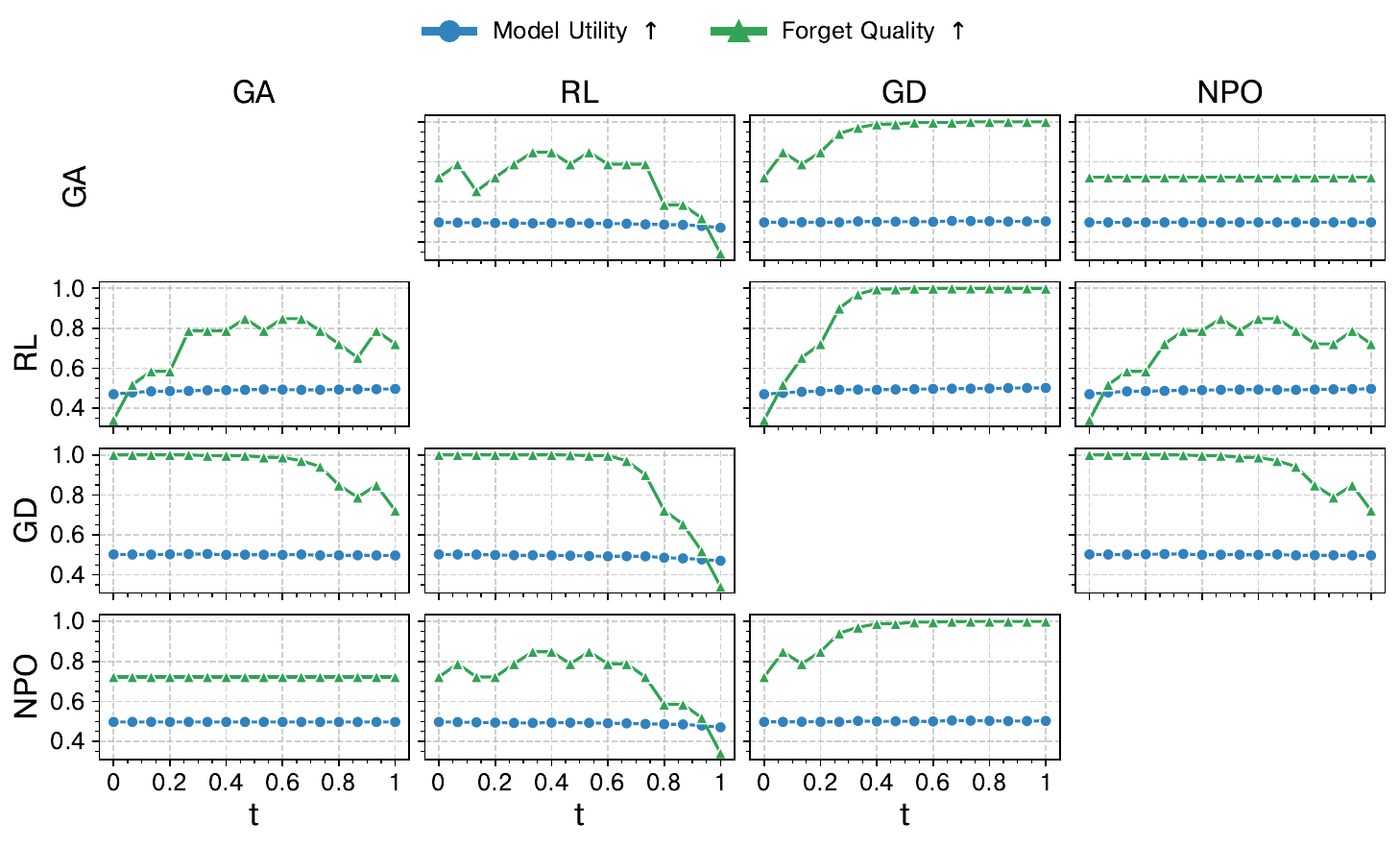}
        \caption{Linear MCU when $|D_f|=10\%$}
    \end{subfigure}
    \begin{subfigure}{0.49\textwidth}
        \centering
        \includegraphics[width=\linewidth]{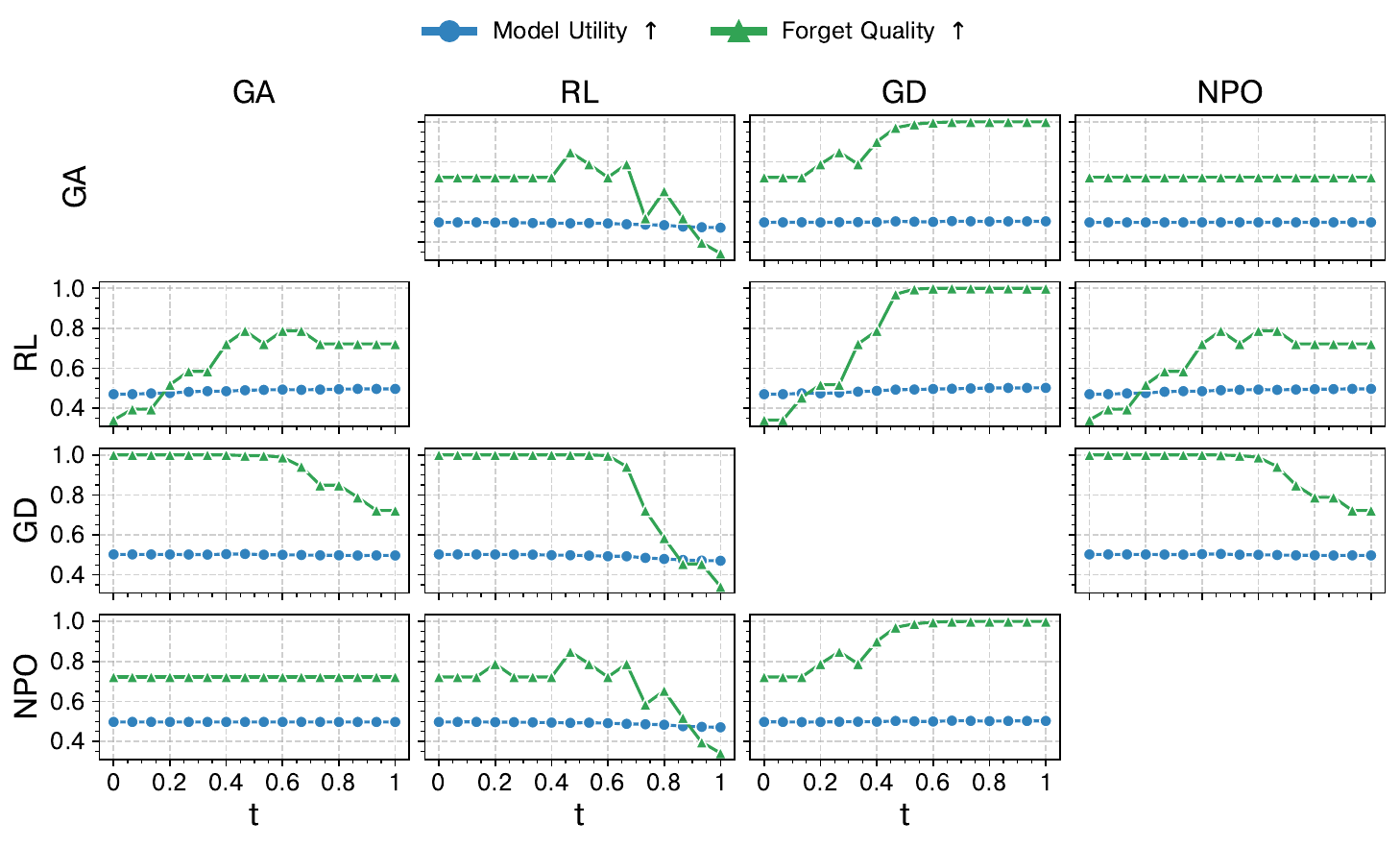}
        \caption{Bezier MCU when $|D_f|=10\%$}
    \end{subfigure}
\end{center}
\caption{MCU under \textbf{Met-CL-Non-CL} setting on \textbf{TOFU dataset}.}
\label{fig:tofu-met-cl-non-cl}
\end{figure}

\begin{figure}
\begin{center}
    \begin{subfigure}{0.49\textwidth}
        \includegraphics[width=\linewidth]{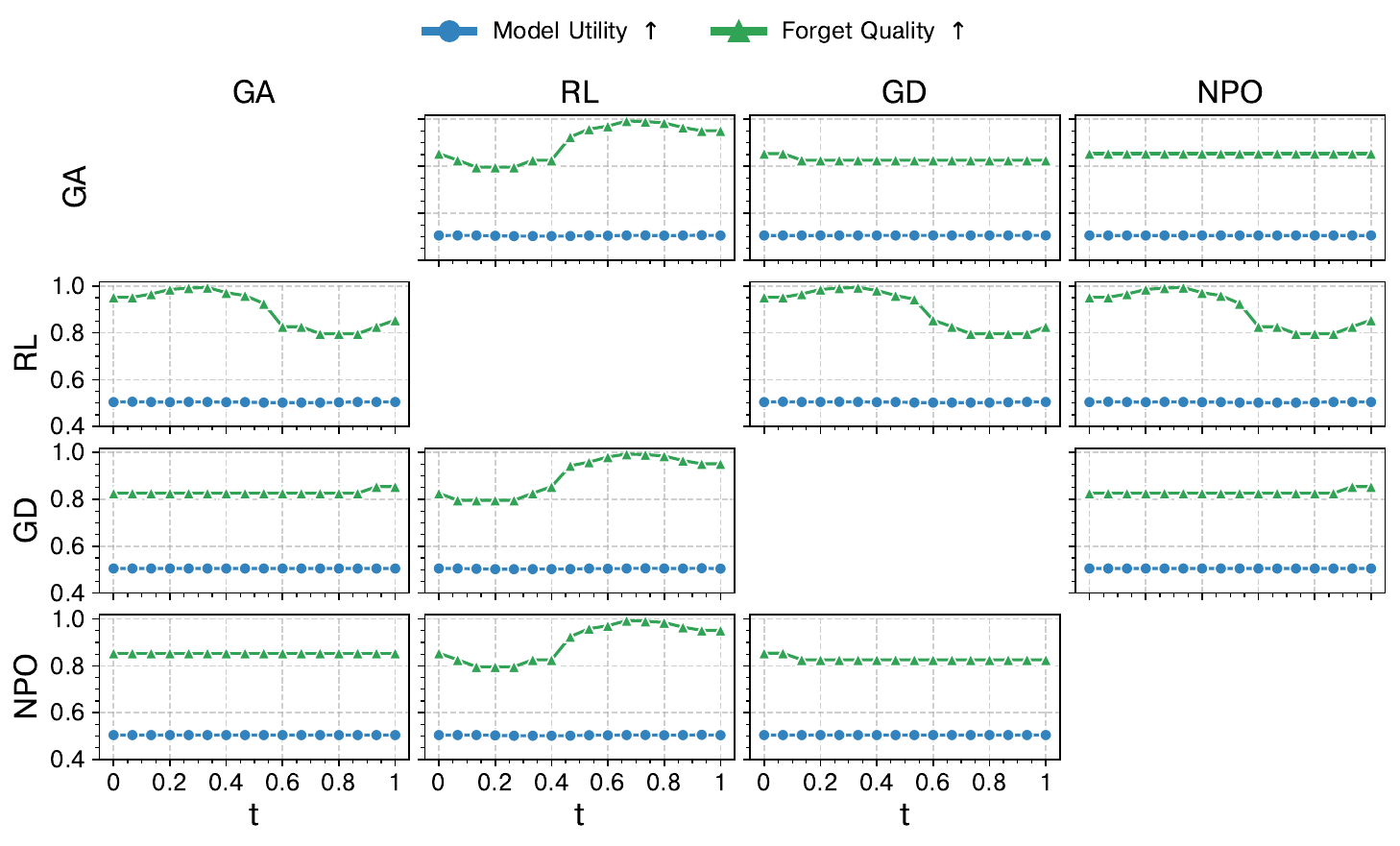}
        \caption{Linear MCU when $|D_f|=1\%$}
    \end{subfigure}
    \begin{subfigure}{0.49\textwidth}
        \includegraphics[width=\linewidth]{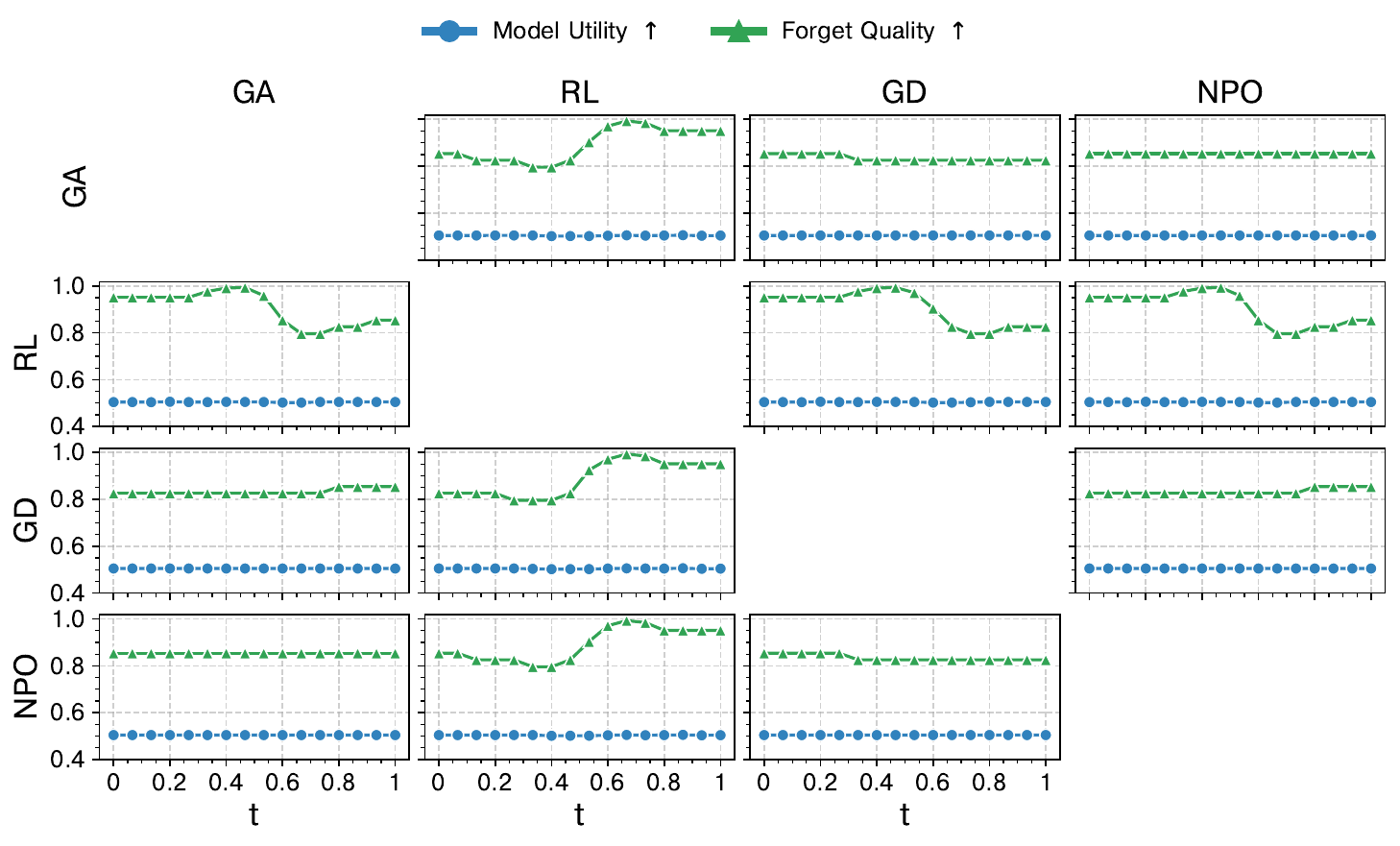}
        \caption{Bezier MCU when $|D_f|=1\%$}
    \end{subfigure}
    \begin{subfigure}{0.49\textwidth}
        \centering
        \includegraphics[width=\linewidth]{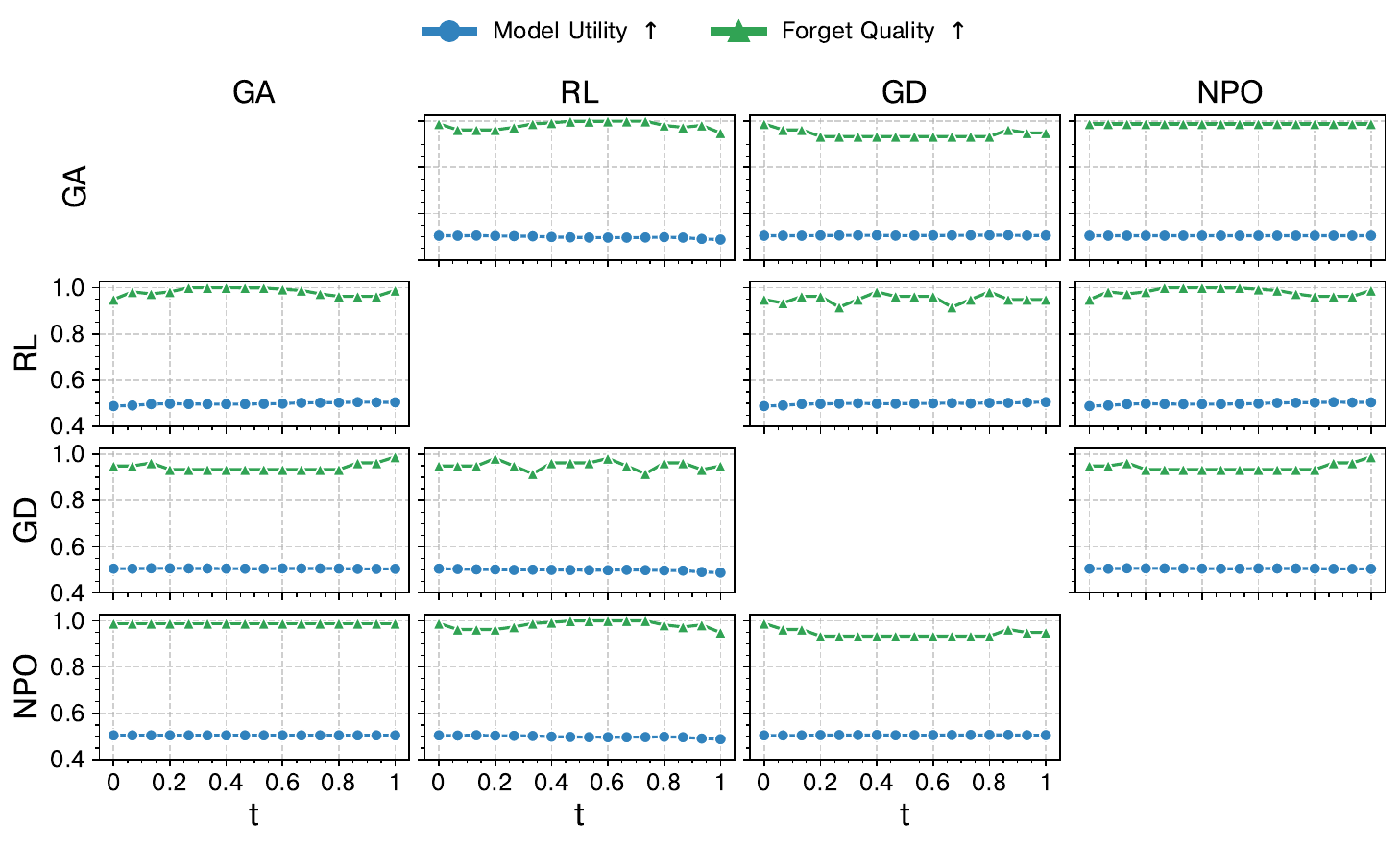}
        \caption{Linear MCU when $|D_f|=5\%$}
    \end{subfigure}
    \begin{subfigure}{0.49\textwidth}
        \centering
        \includegraphics[width=\linewidth]{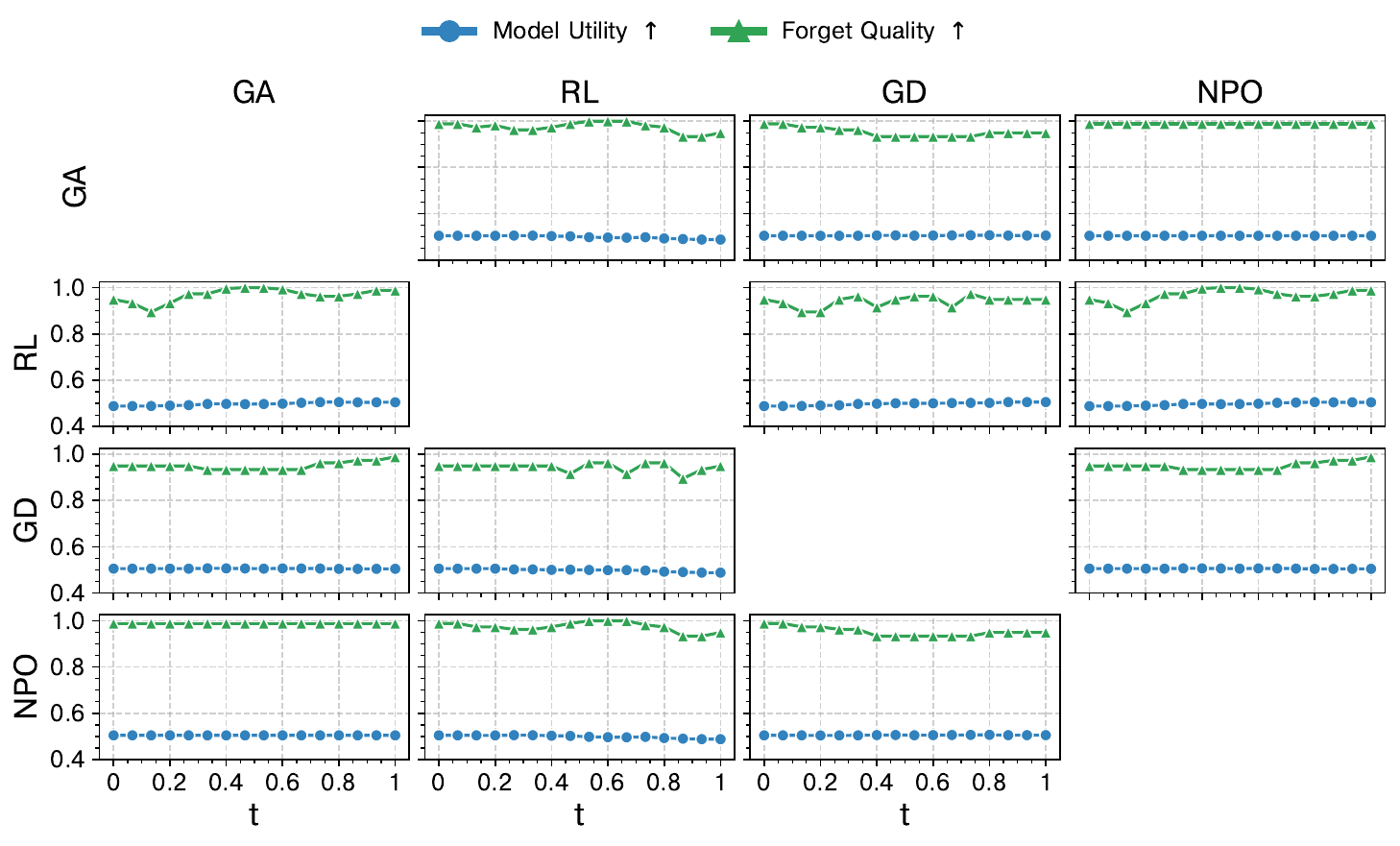}
        \caption{Bezier MCU when $|D_f|=5\%$}
    \end{subfigure}
    \begin{subfigure}{0.49\textwidth}
        \centering
        \includegraphics[width=\linewidth]{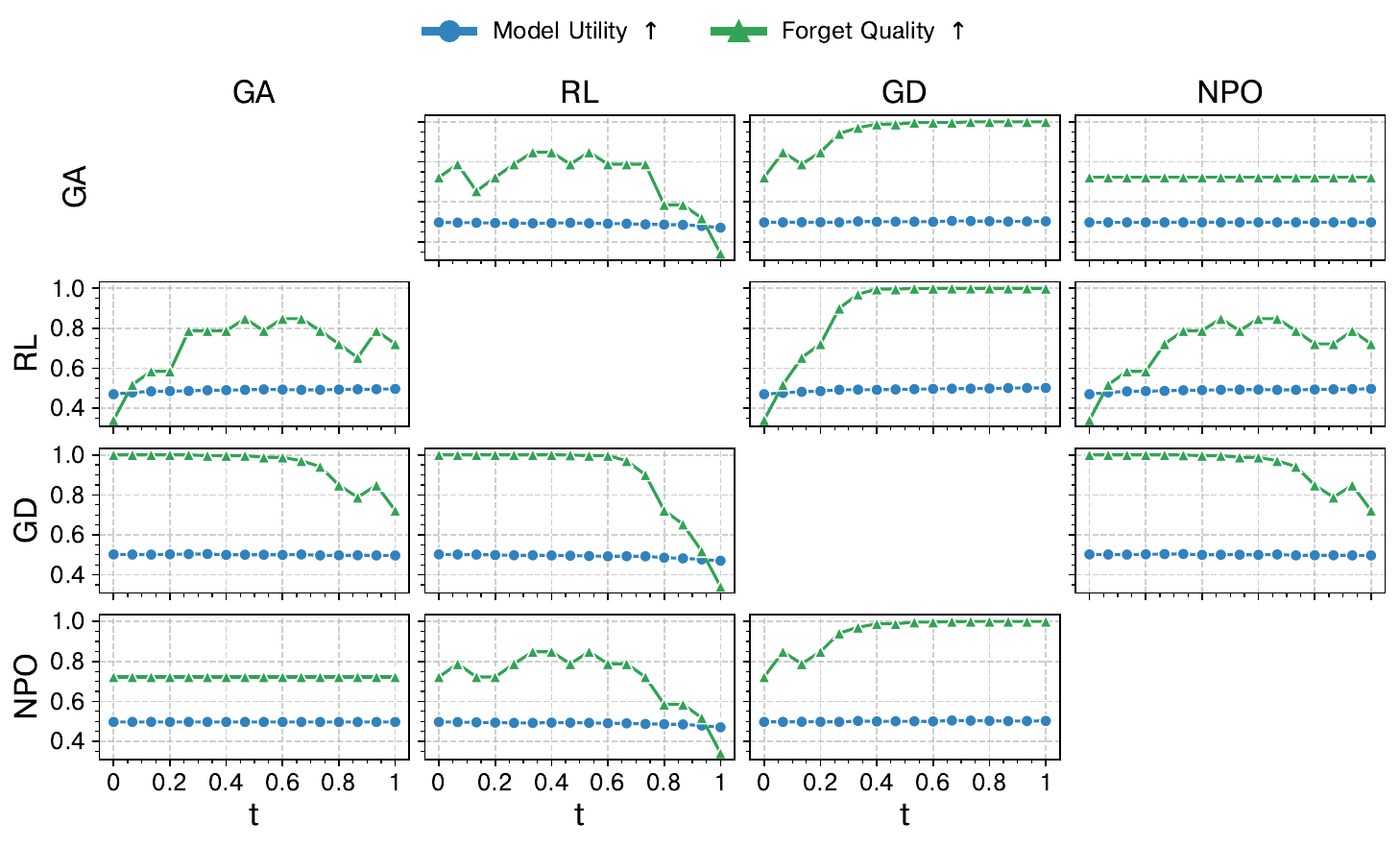}
        \caption{Linear MCU when $|D_f|=10\%$}
    \end{subfigure}
    \begin{subfigure}{0.49\textwidth}
        \centering
        \includegraphics[width=\linewidth]{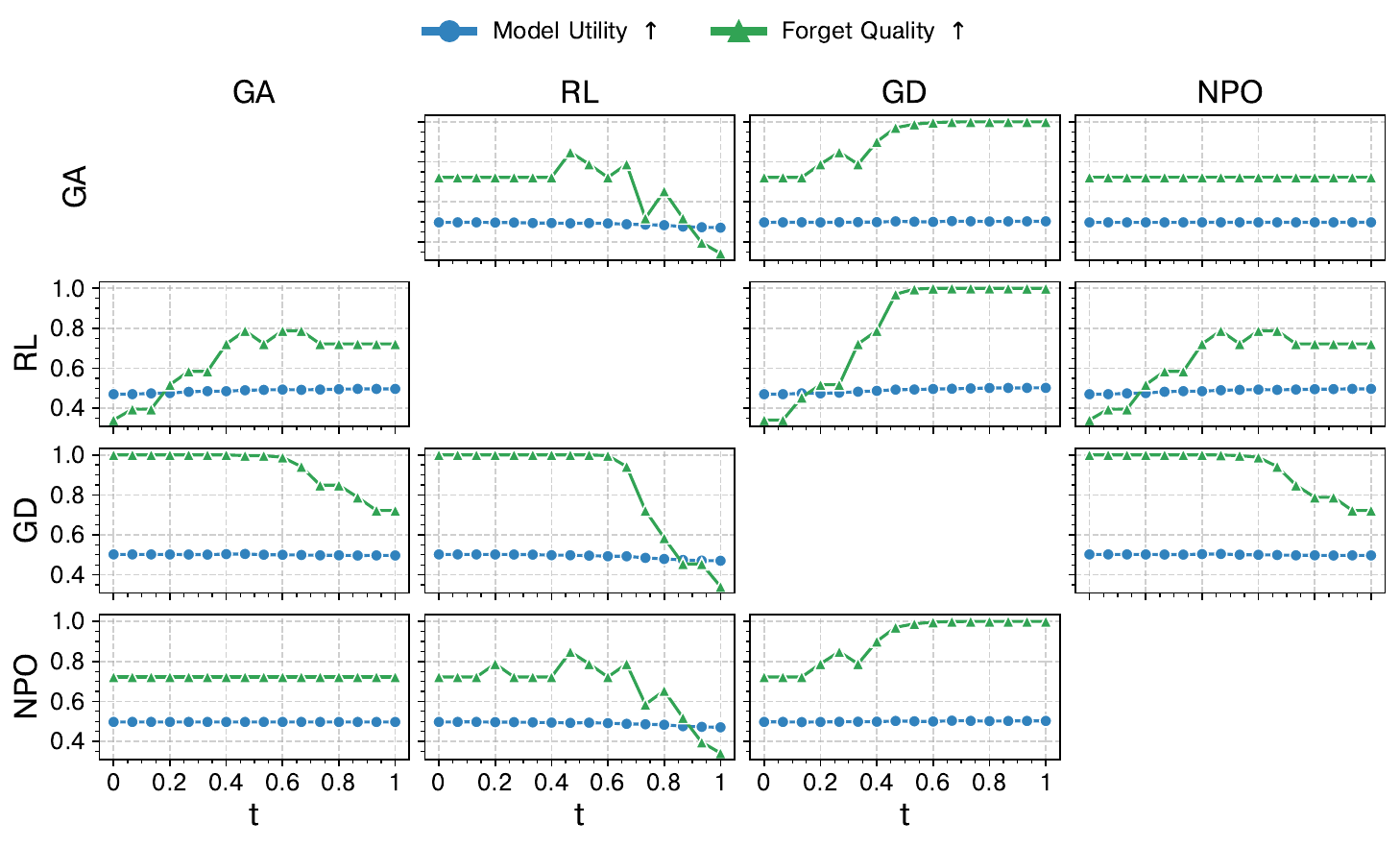}
        \caption{Bezier MCU when $|D_f|=10\%$}
    \end{subfigure}
\end{center}
\caption{MCU under \textbf{Met-FO-SO} setting on \textbf{TOFU dataset}.}
\label{fig:tofu-met-fo-so}
\end{figure}

\begin{figure}
    \centering
    \includegraphics[width=0.99\linewidth]{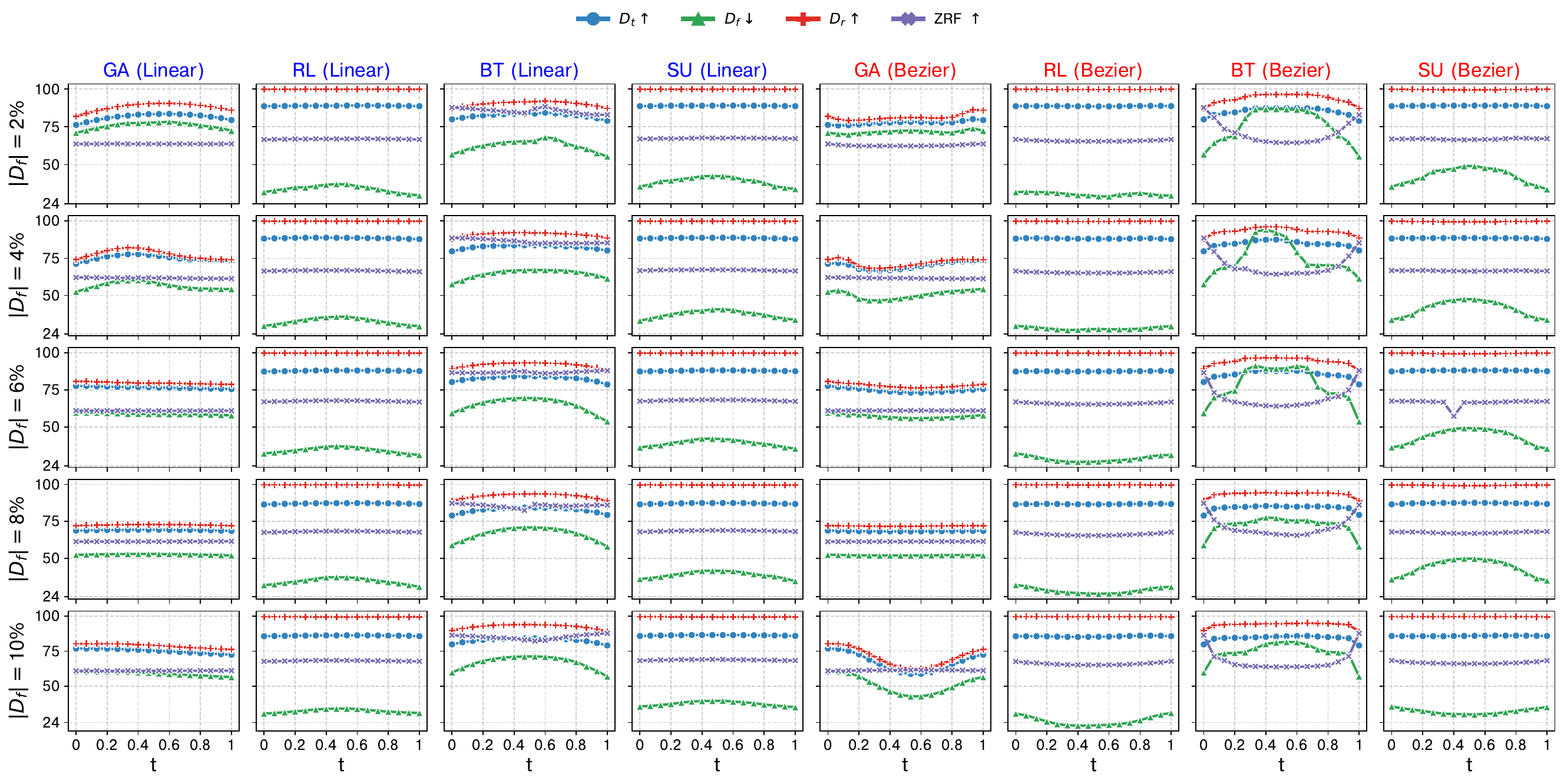}
    \caption{MCU under \textbf{Rand} setting on \textbf{classification dataset}.}
    \label{fig:cls-rand}
\end{figure}

\begin{figure}
    \centering
    \includegraphics[width=0.99\linewidth]{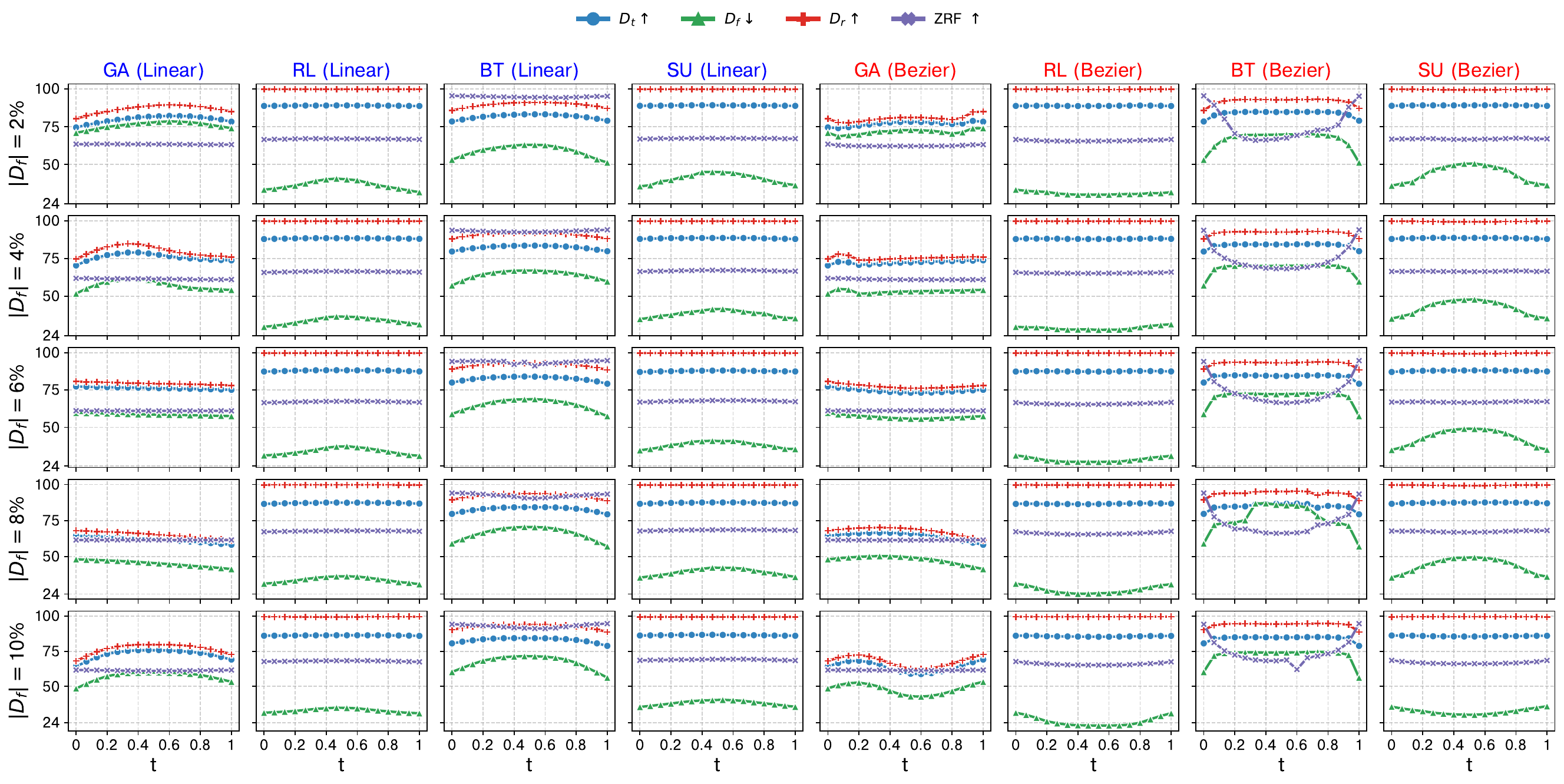}
    \caption{MCU under \textbf{Rand-CL} setting on \textbf{classification dataset}.}
    \label{fig:cls-rand-cl}
\end{figure}

\begin{figure}
    \centering
    \includegraphics[width=0.99\linewidth]{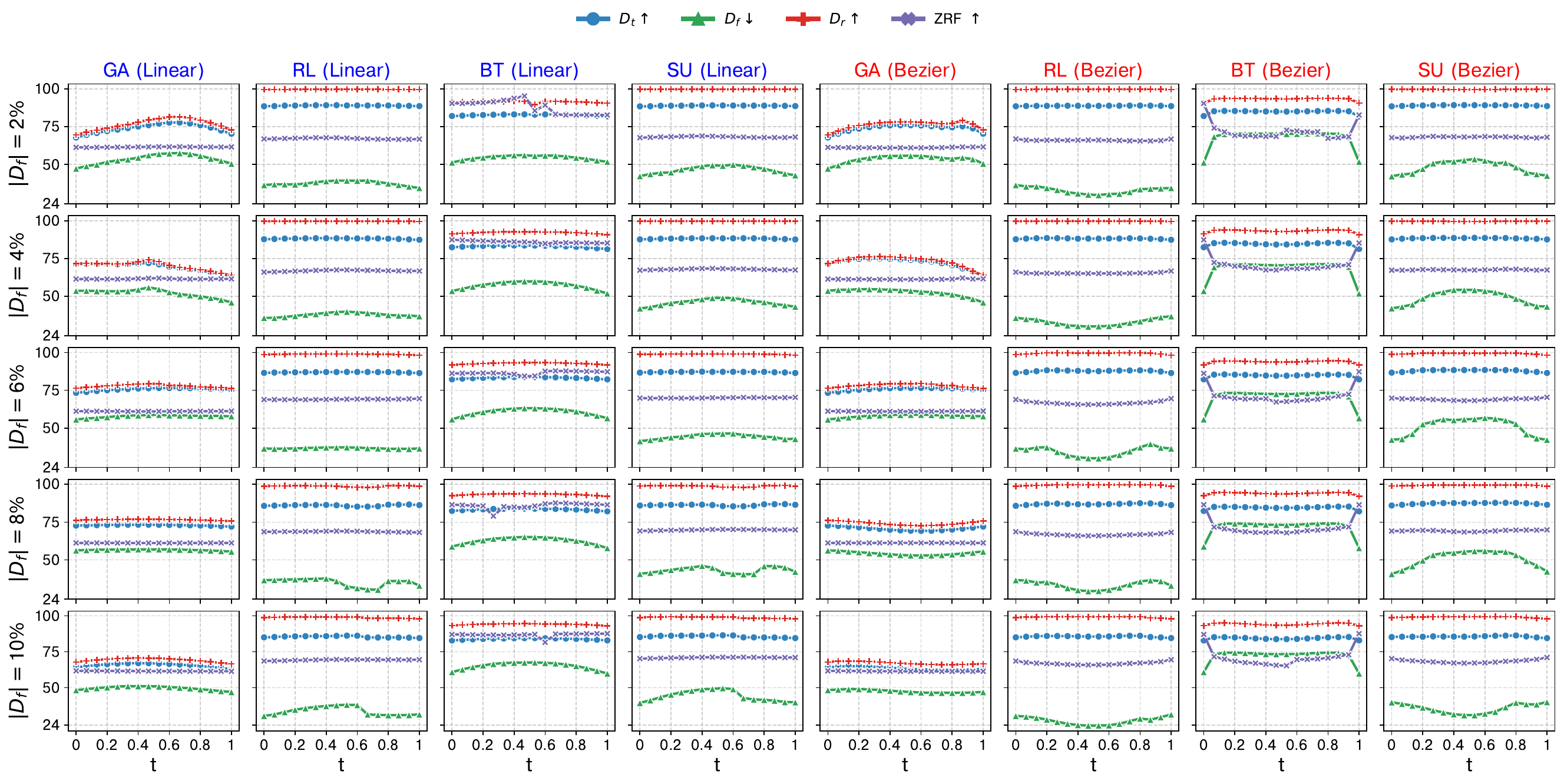}
    \caption{MCU under \textbf{Rand-SO} setting on \textbf{classification dataset}.}
    \label{fig:cls-rand-so}
\end{figure}

\begin{figure}
    \centering
    \includegraphics[width=0.99\linewidth]{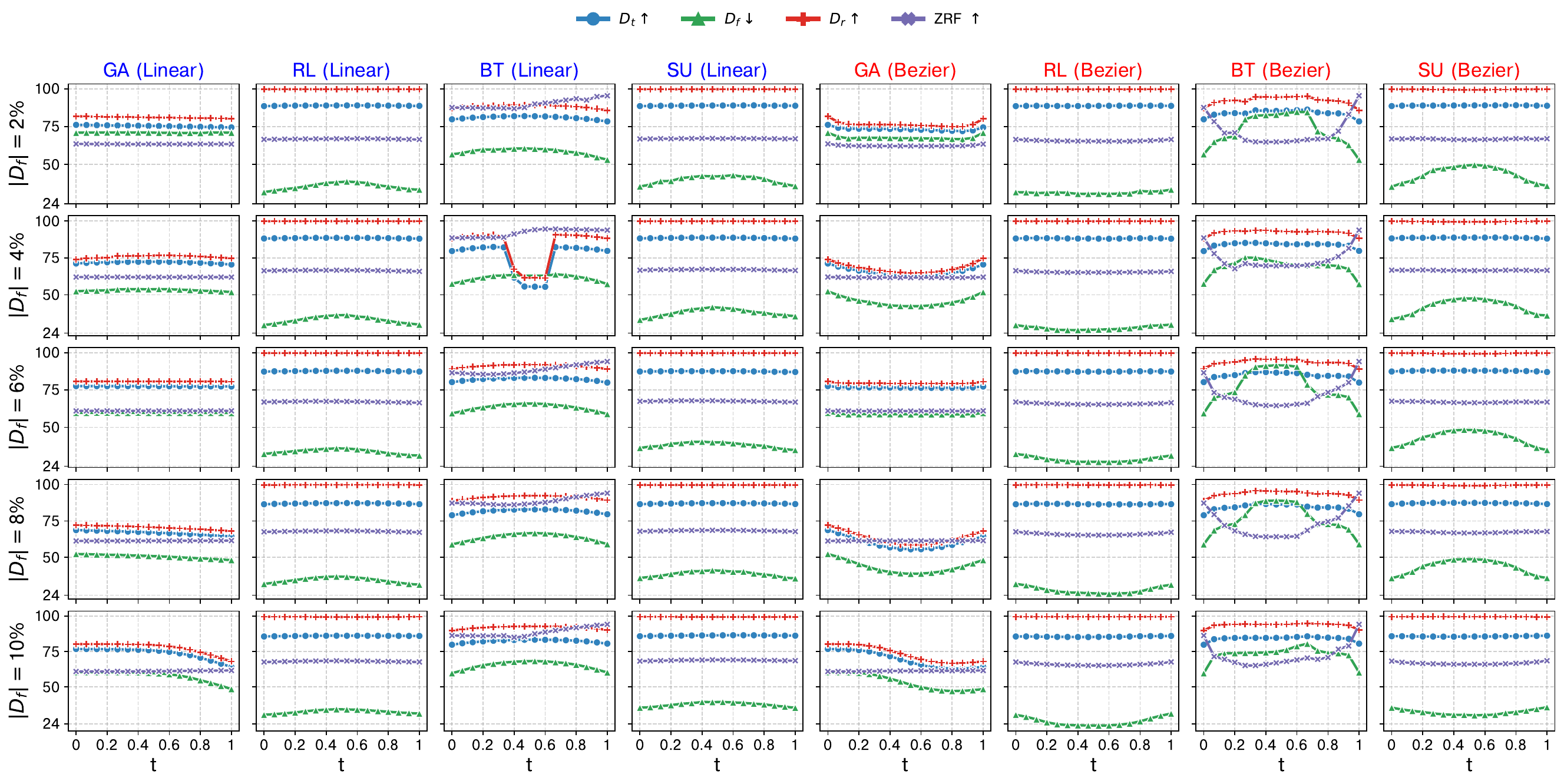}
    \caption{MCU under \textbf{CL-Non-CL} setting on \textbf{classification dataset}.}
    \label{fig:cls-cl-non-cl}
\end{figure}

\begin{figure}
    \centering
    \includegraphics[width=0.99\linewidth]{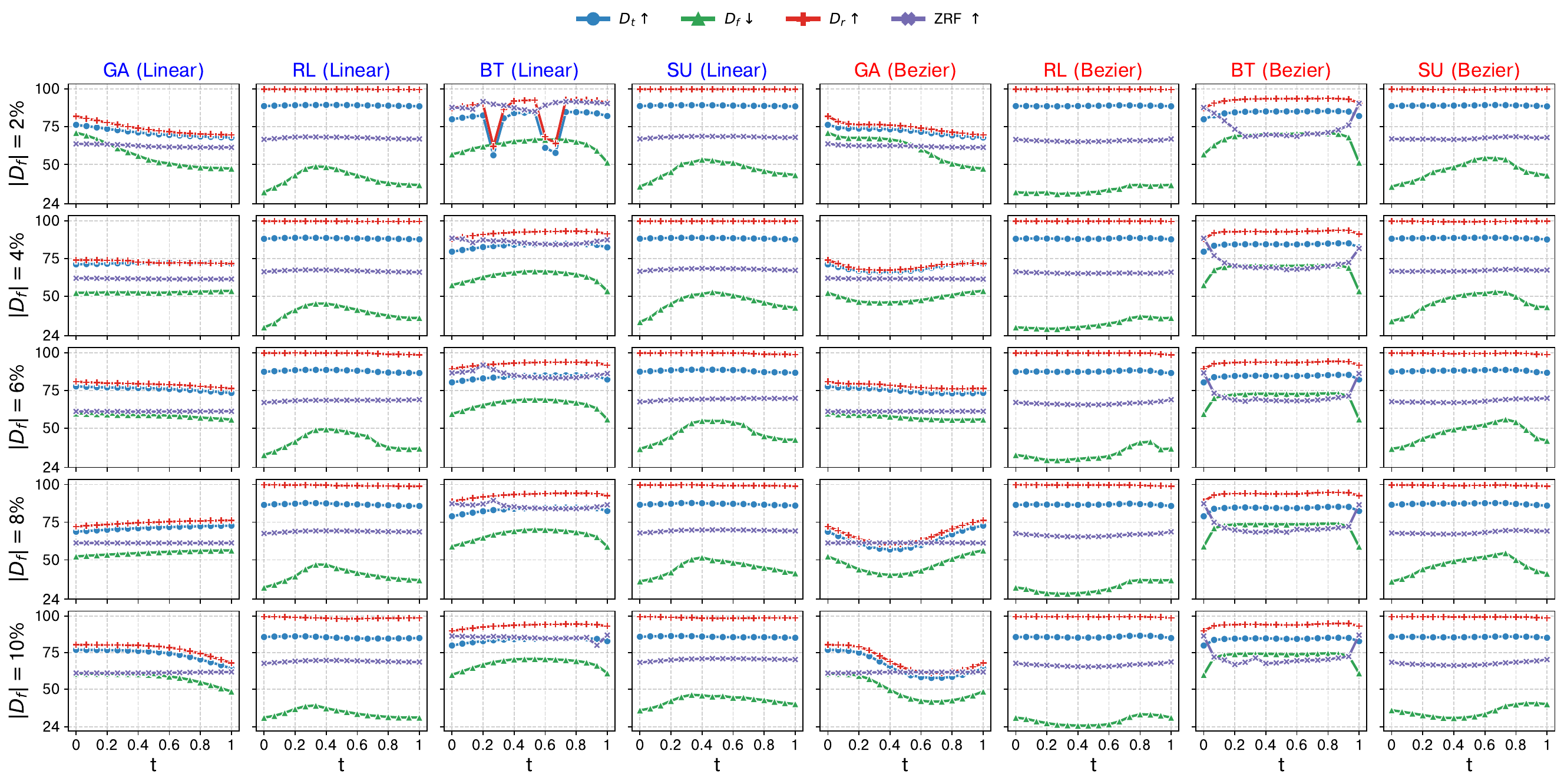}
    \caption{MCU under \textbf{FO-SO} setting on \textbf{classification dataset}.}
    \label{fig:cls-fo-so}
\end{figure}

\begin{figure}
\begin{center}
    \begin{subfigure}{0.49\textwidth}
        \includegraphics[width=\linewidth]{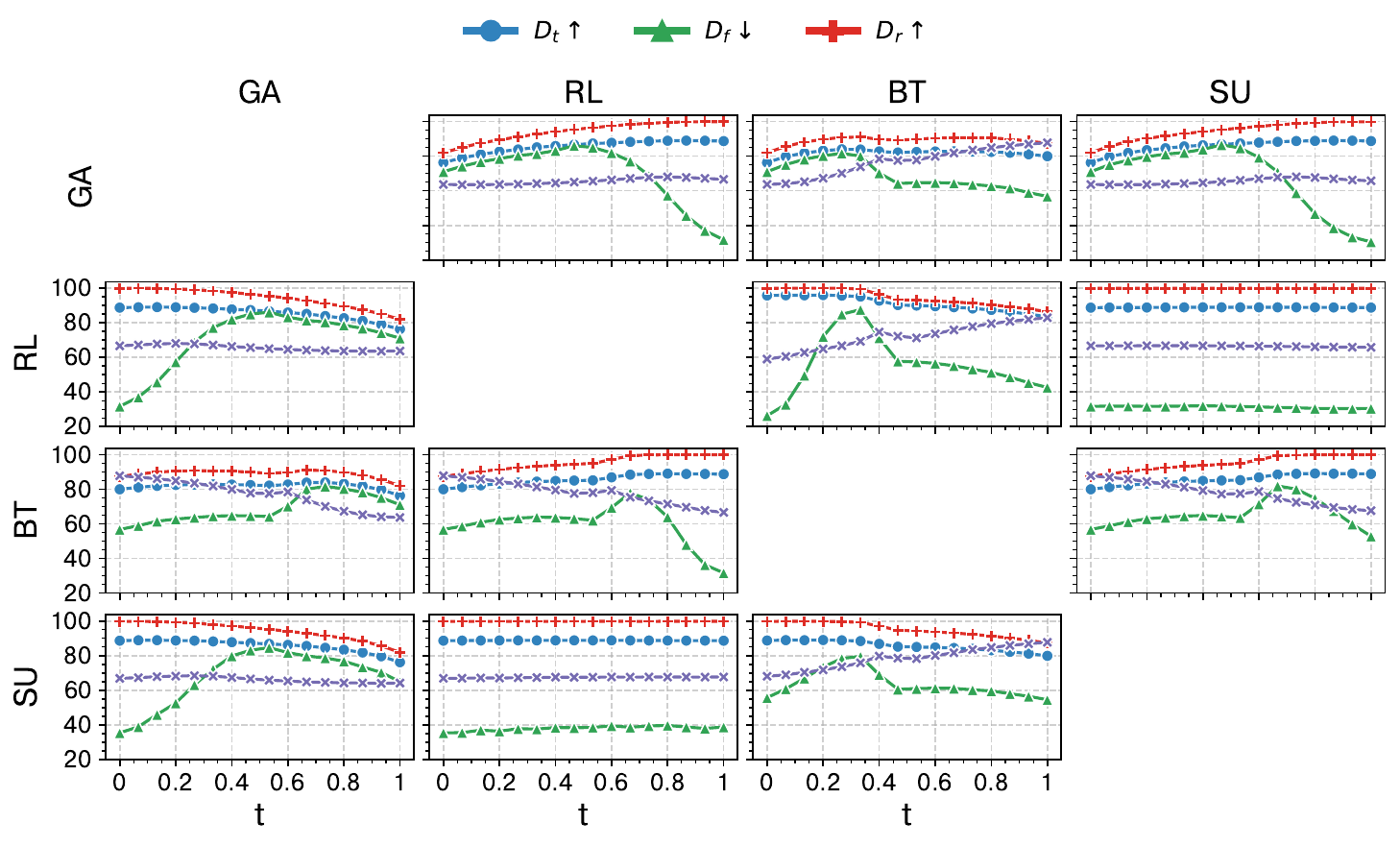}
        \caption{Linear MCU when $|D_f|=2.0\%$}
    \end{subfigure}
    \begin{subfigure}{0.49\textwidth}
        \includegraphics[width=\linewidth]{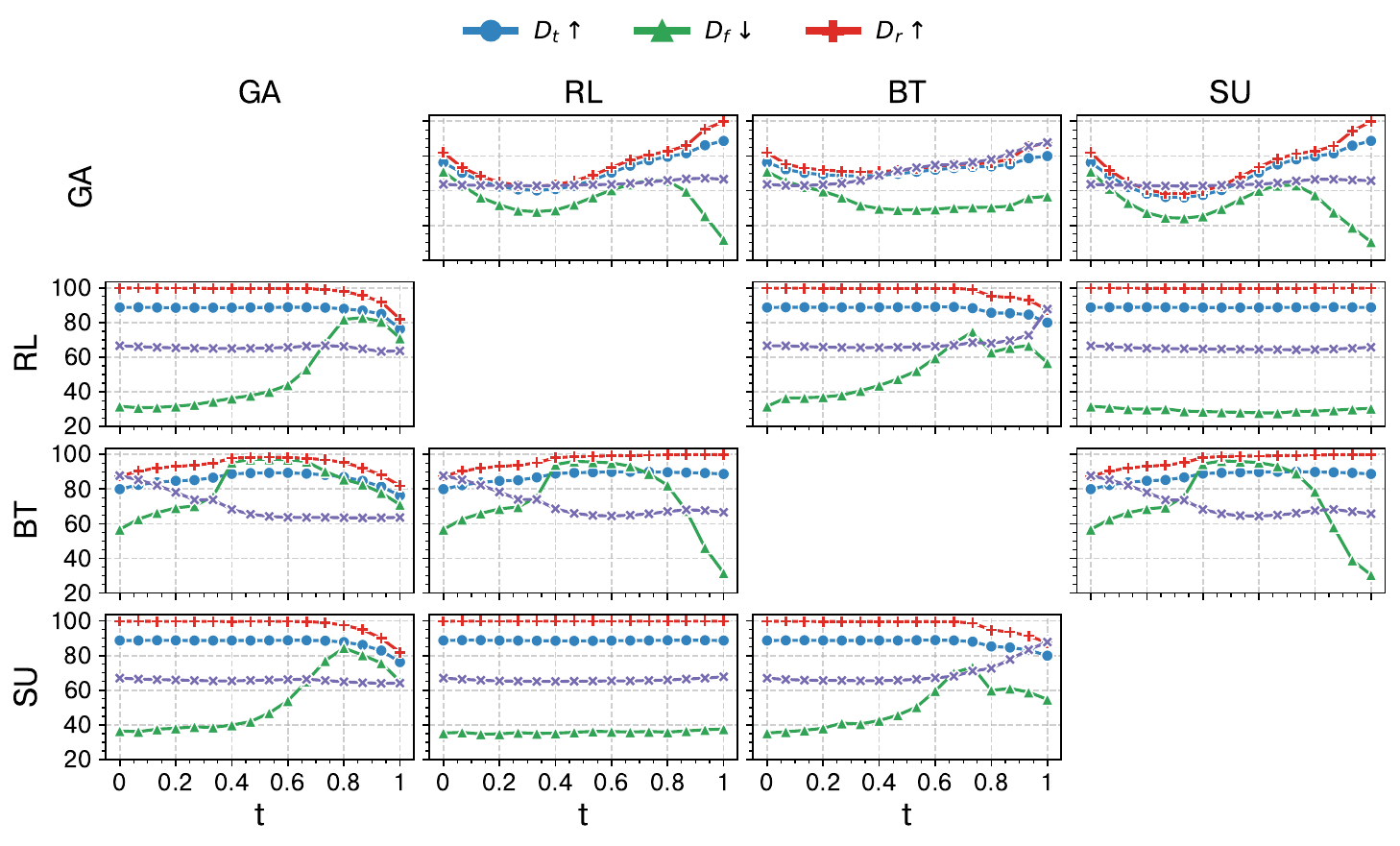}
        \caption{Quadratic MCU when $|D_f|=2.0\%$}
    \end{subfigure}
    \begin{subfigure}{0.49\textwidth}
        \centering
        \includegraphics[width=\linewidth]{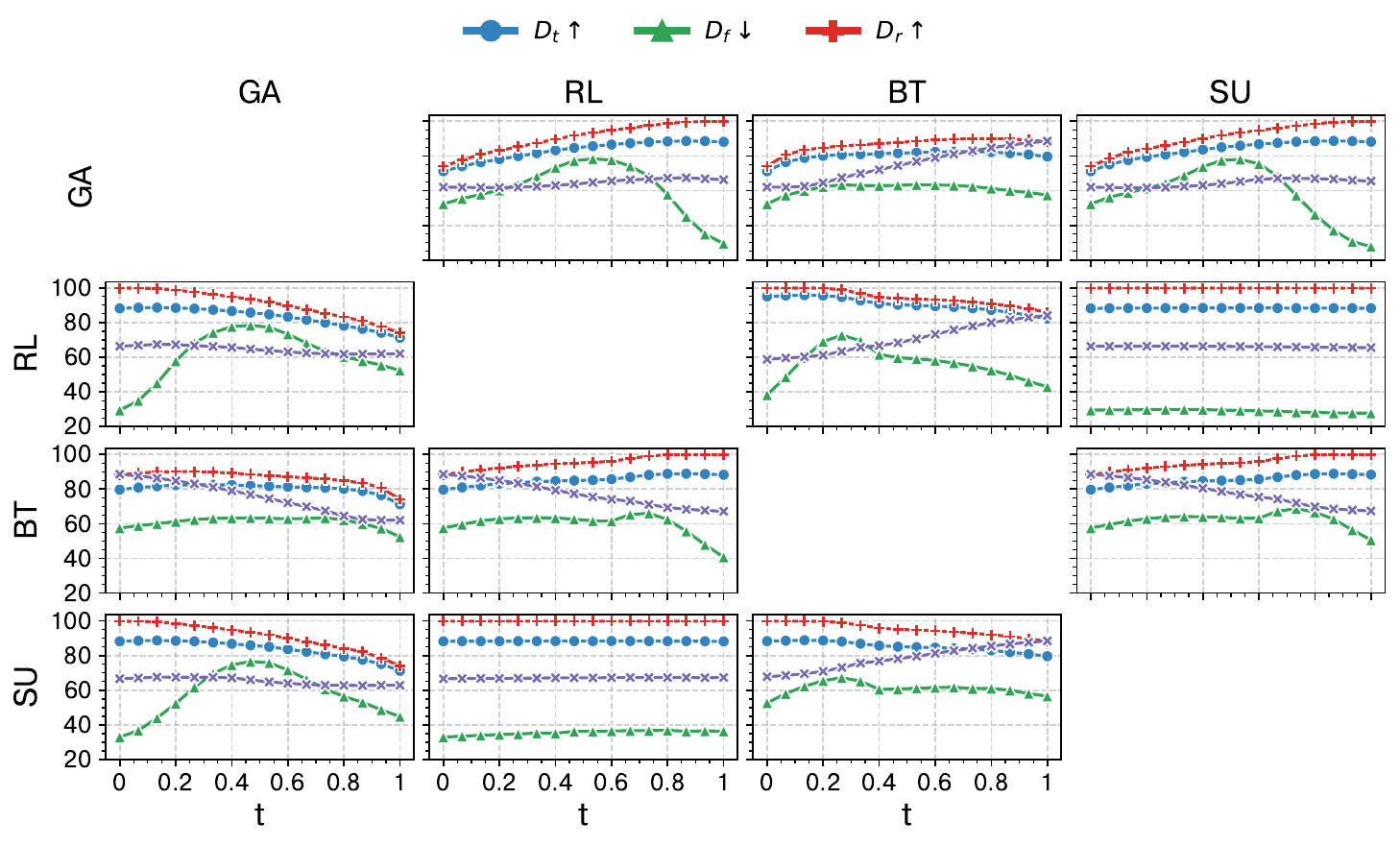}
        \caption{Linear MCU when $|D_f|=4.0\%$}
    \end{subfigure}
    \begin{subfigure}{0.49\textwidth}
        \centering
        \includegraphics[width=\linewidth]{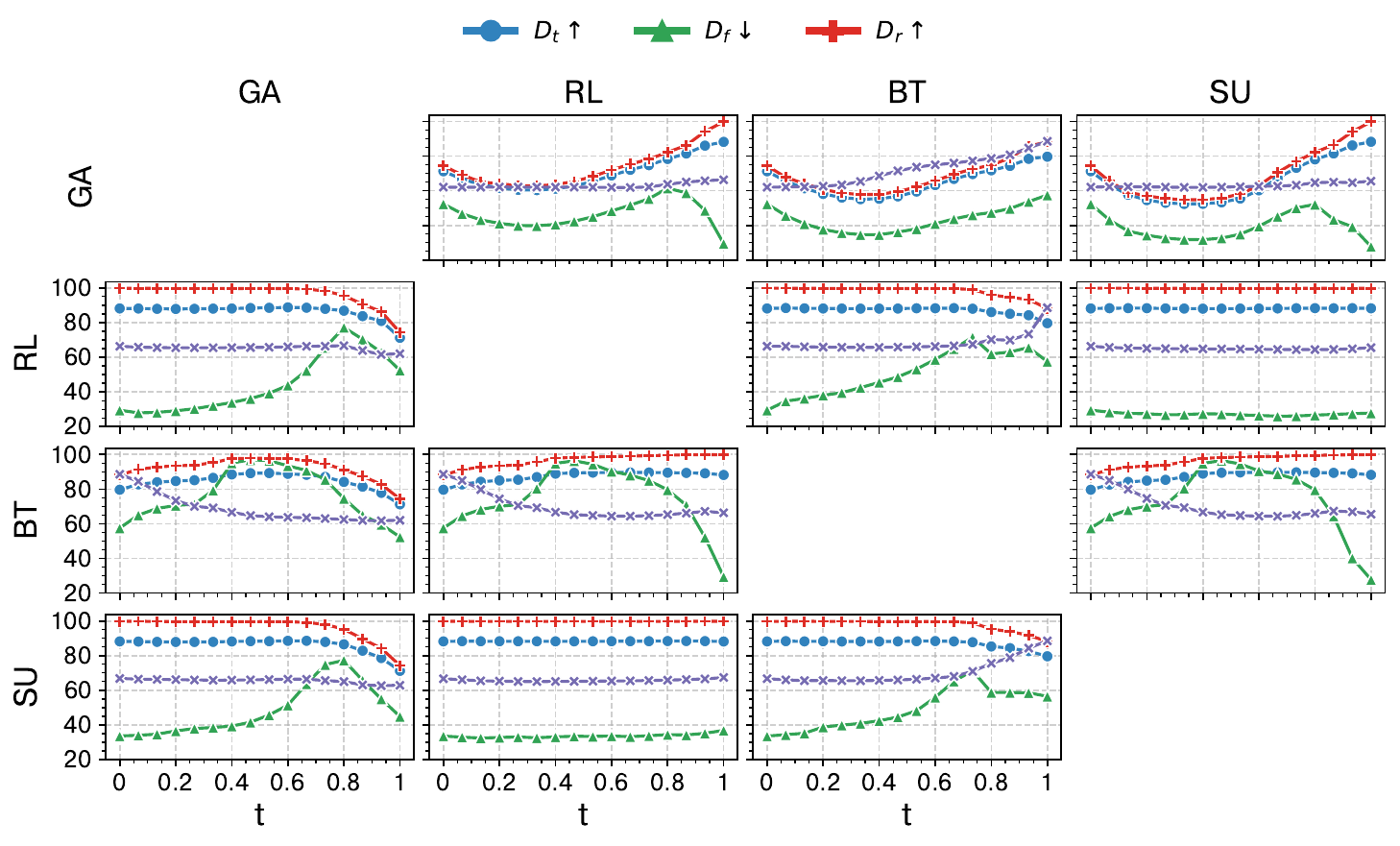}
        \caption{Quadratic MCU when $|D_f|=4.0\%$}
    \end{subfigure}
\end{center}
\end{figure}
\begin{figure}
\begin{center}
    \begin{subfigure}{0.49\textwidth}
        \centering
        \includegraphics[width=\linewidth]{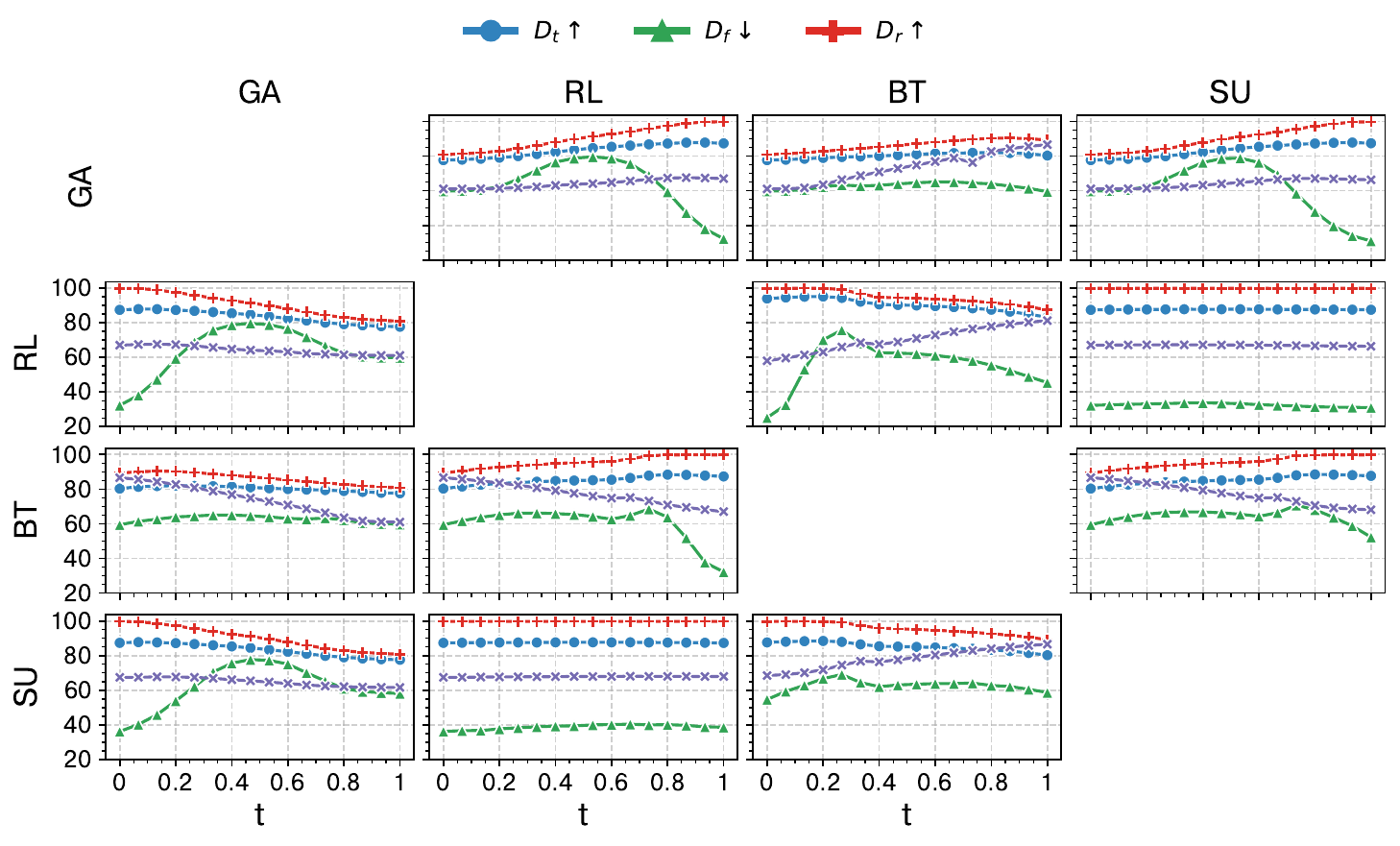}
        \caption{Linear MCU when $|D_f|=6.0\%$}
    \end{subfigure}
    \begin{subfigure}{0.49\textwidth}
        \centering
        \includegraphics[width=\linewidth]{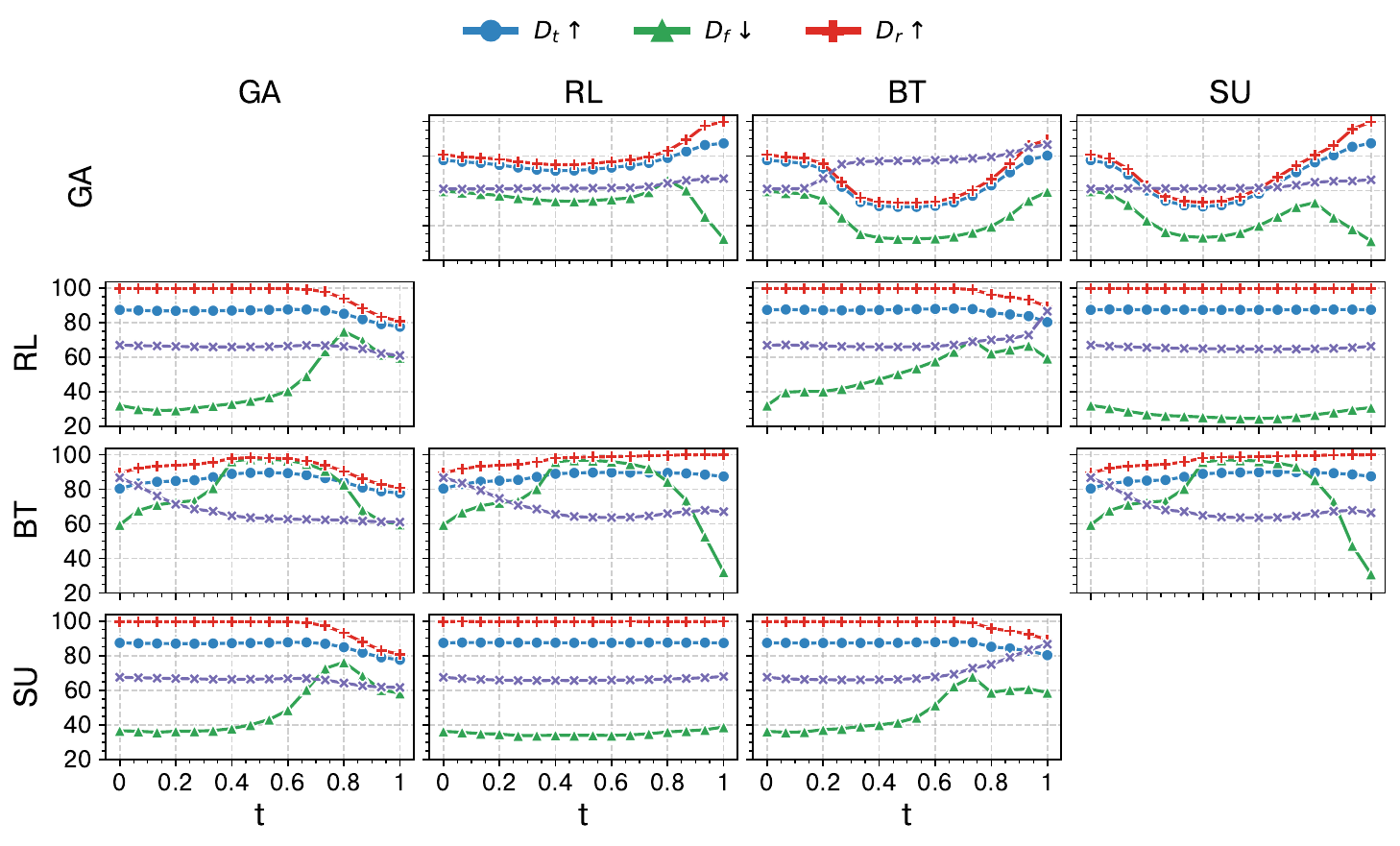}
        \caption{Quadratic MCU when $|D_f|=6.0\%$}
    \end{subfigure}
    \begin{subfigure}{0.49\textwidth}
        \centering
        \includegraphics[width=\linewidth]{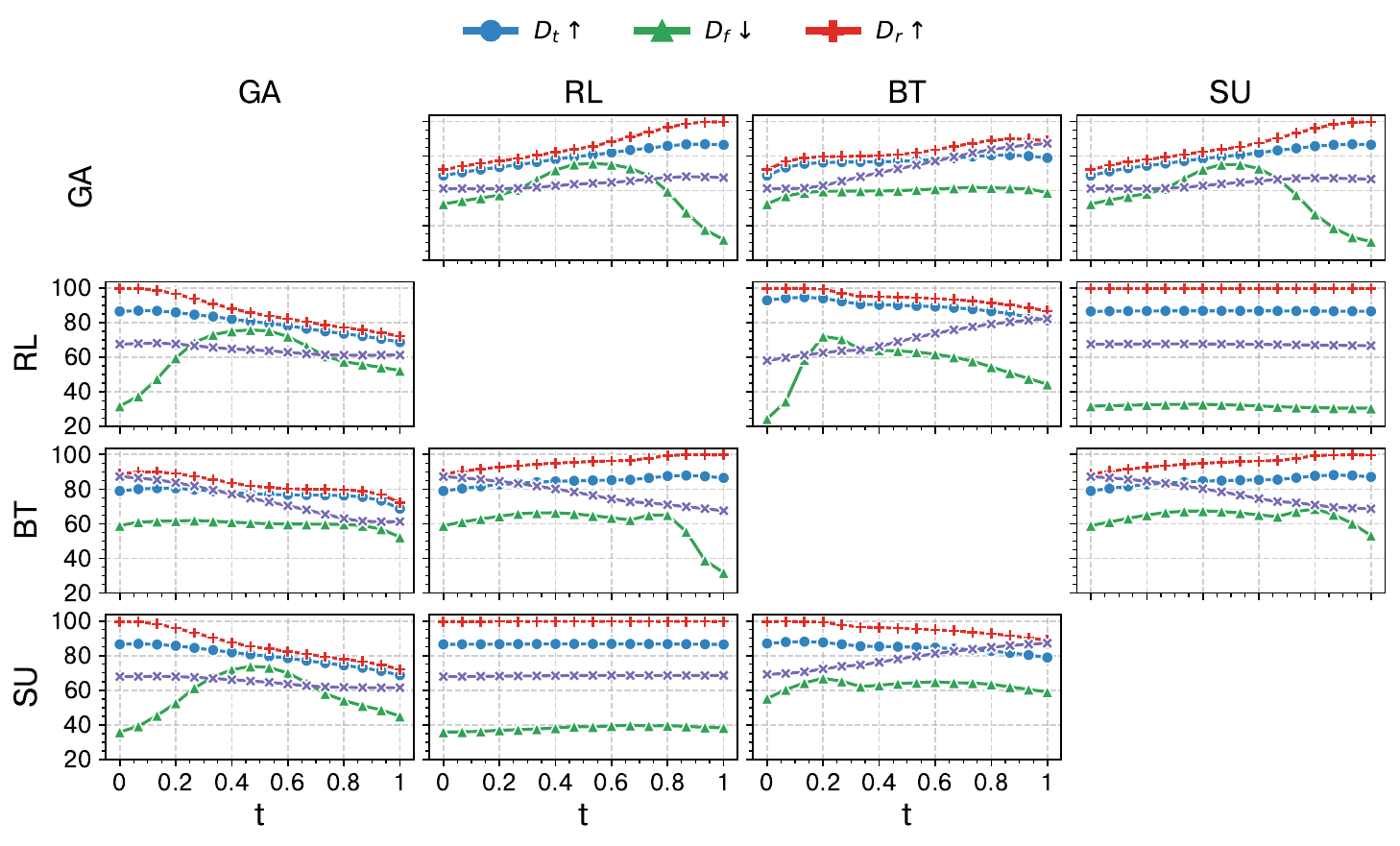}
        \caption{Linear MCU when $|D_f|=8.0\%$}
    \end{subfigure}
    \begin{subfigure}{0.49\textwidth}
        \centering
        \includegraphics[width=\linewidth]{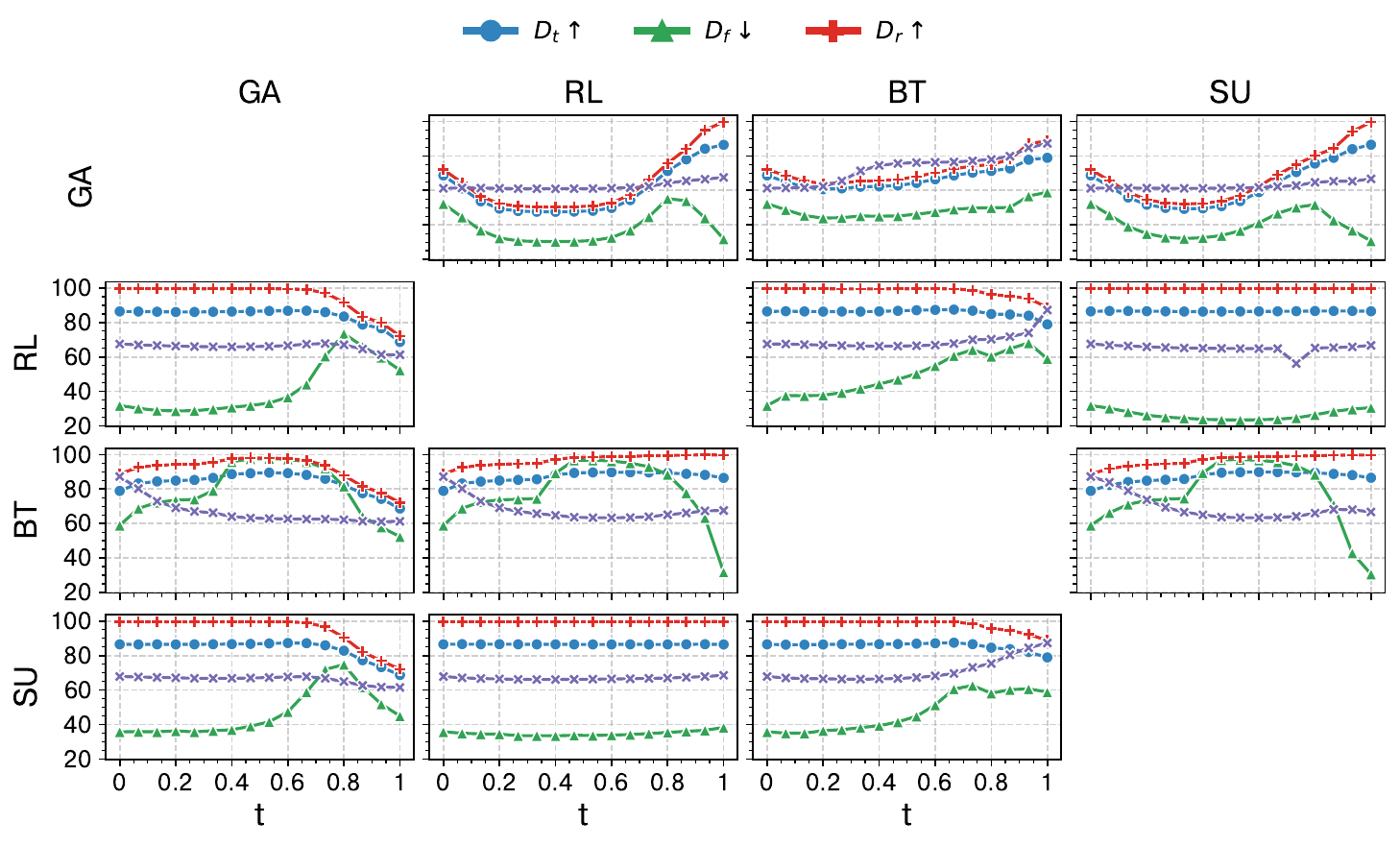}
        \caption{Quadratic MCU when $|D_f|=8.0\%$}
    \end{subfigure}
    \begin{subfigure}{0.49\textwidth}
        \centering
        \includegraphics[width=\linewidth]{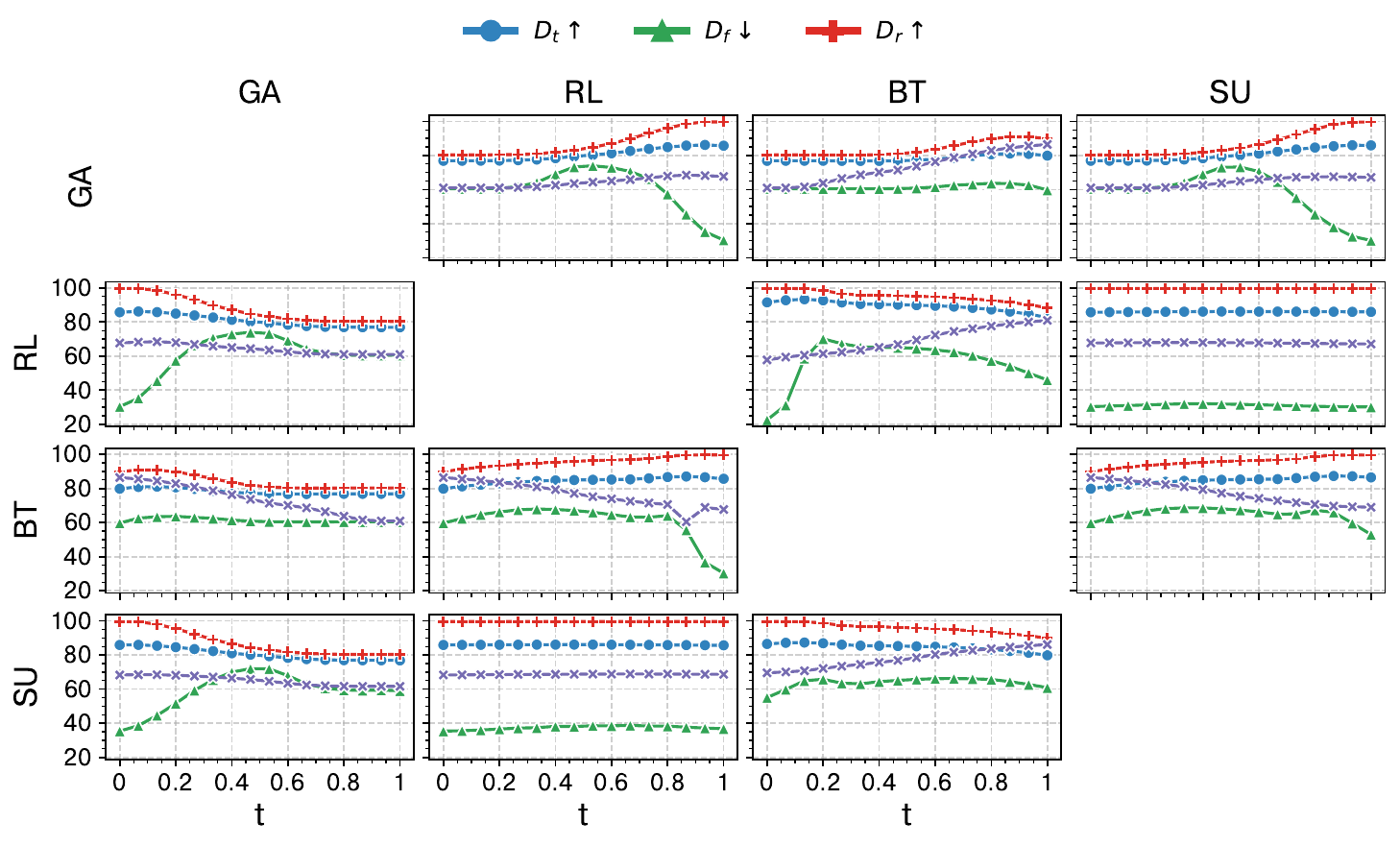}
        \caption{Linear MCU when $|D_f|=10.0\%$}
    \end{subfigure}
    \begin{subfigure}{0.49\textwidth}
        \centering
        \includegraphics[width=\linewidth]{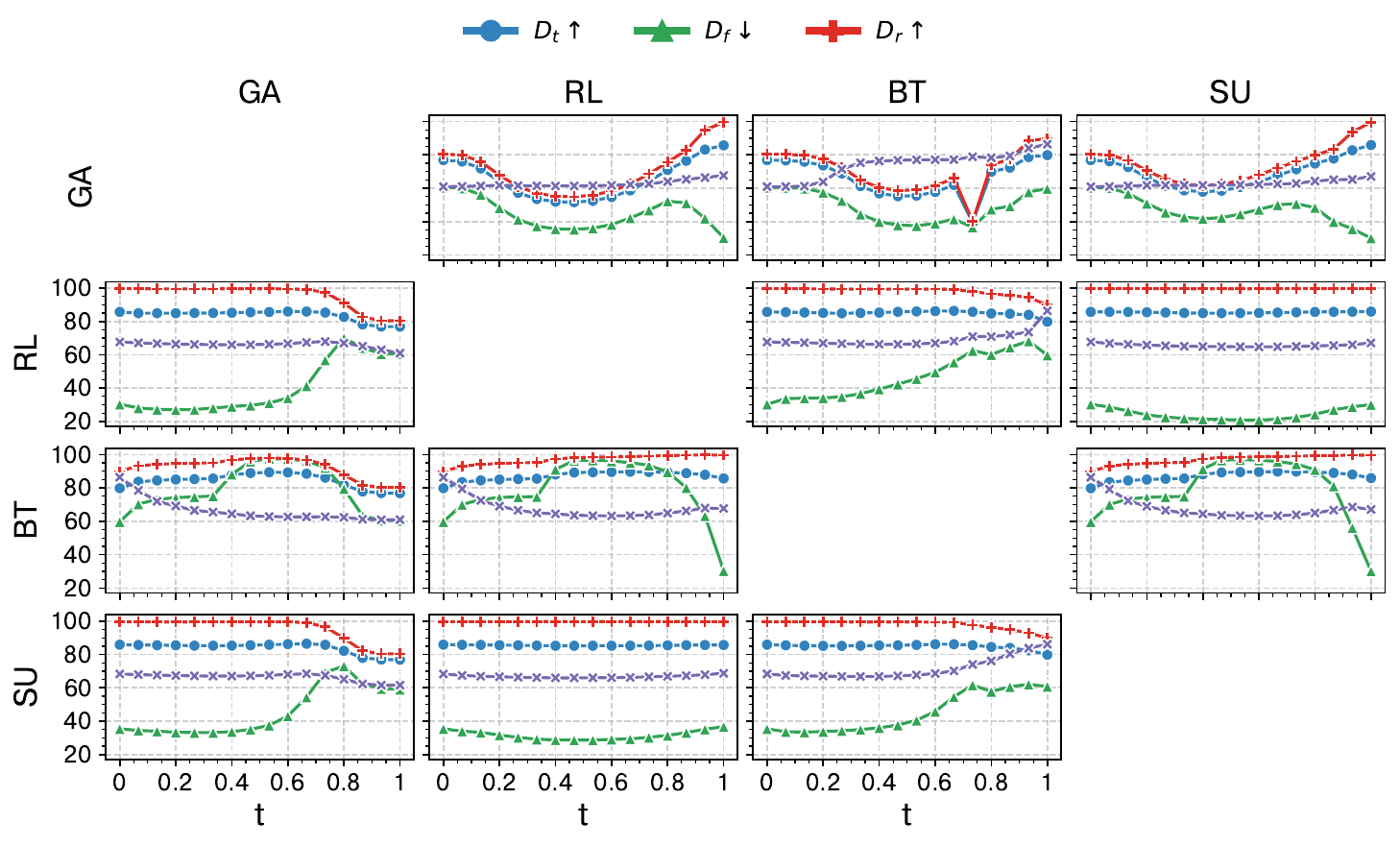}
        \caption{Quadratic MCU when $|D_f|=10.0\%$}
    \end{subfigure}
\end{center}
\caption{MCU under \textbf{Met} setting on \textbf{classification datasets}.}
\label{fig:cls-met}
\end{figure}

\begin{figure}
\begin{center}
    \begin{subfigure}{0.49\textwidth}
        \includegraphics[width=\linewidth]{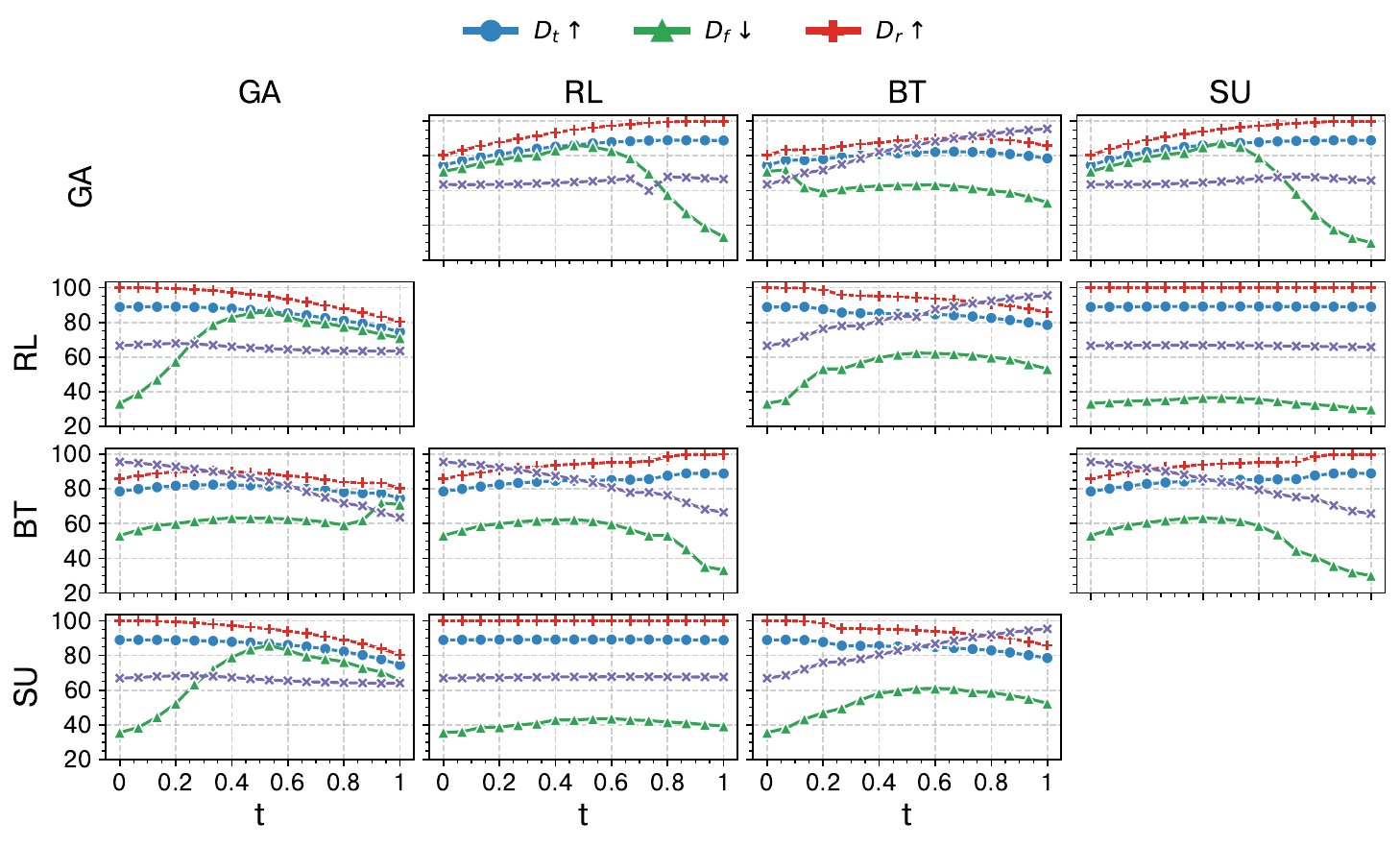}
        \caption{Linear MCU when $|D_f|=2.0\%$}
    \end{subfigure}
    \begin{subfigure}{0.49\textwidth}
        \includegraphics[width=\linewidth]{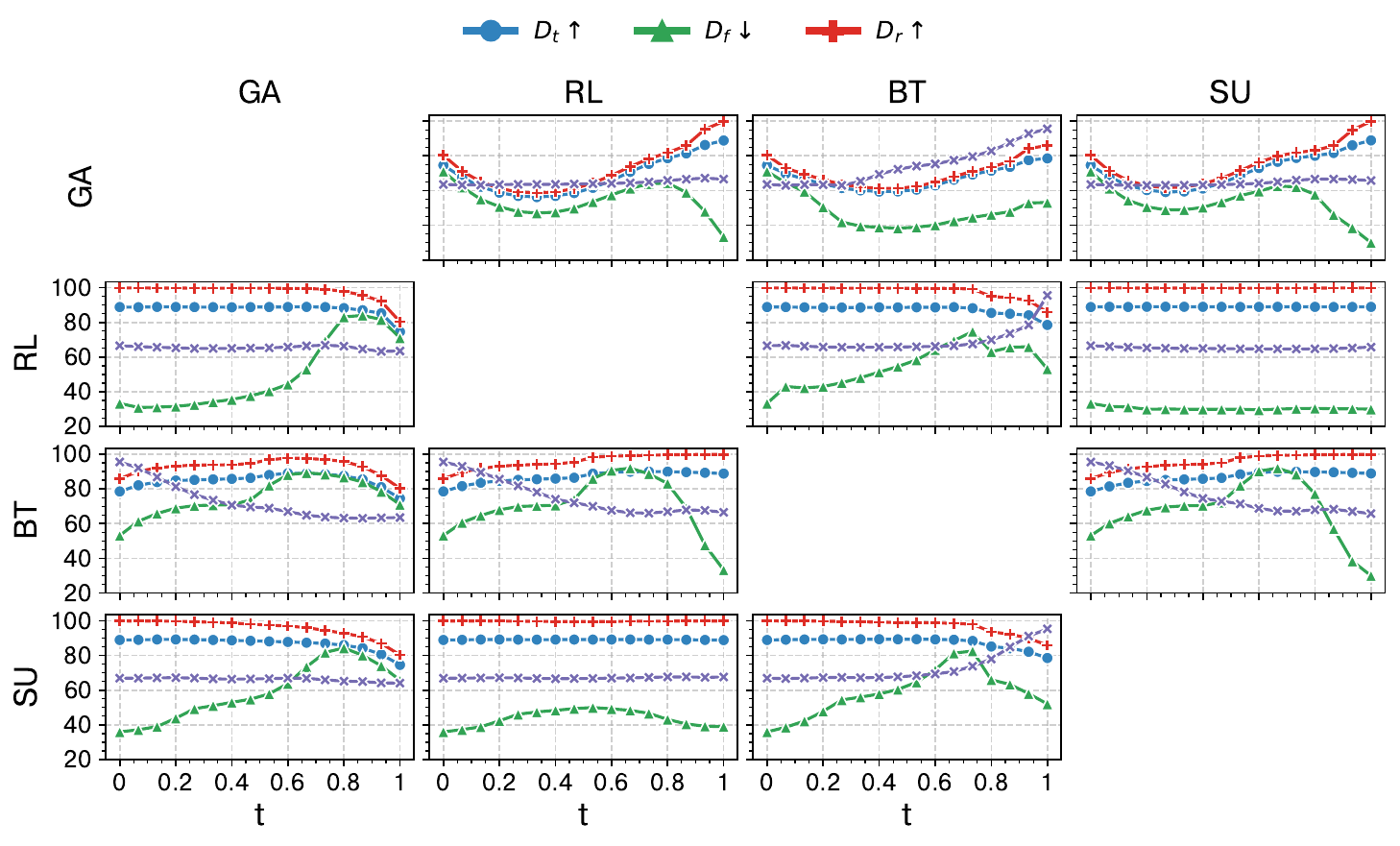}
        \caption{Quadratic MCU when $|D_f|=2.0\%$}
    \end{subfigure}
    \begin{subfigure}{0.49\textwidth}
        \centering
        \includegraphics[width=\linewidth]{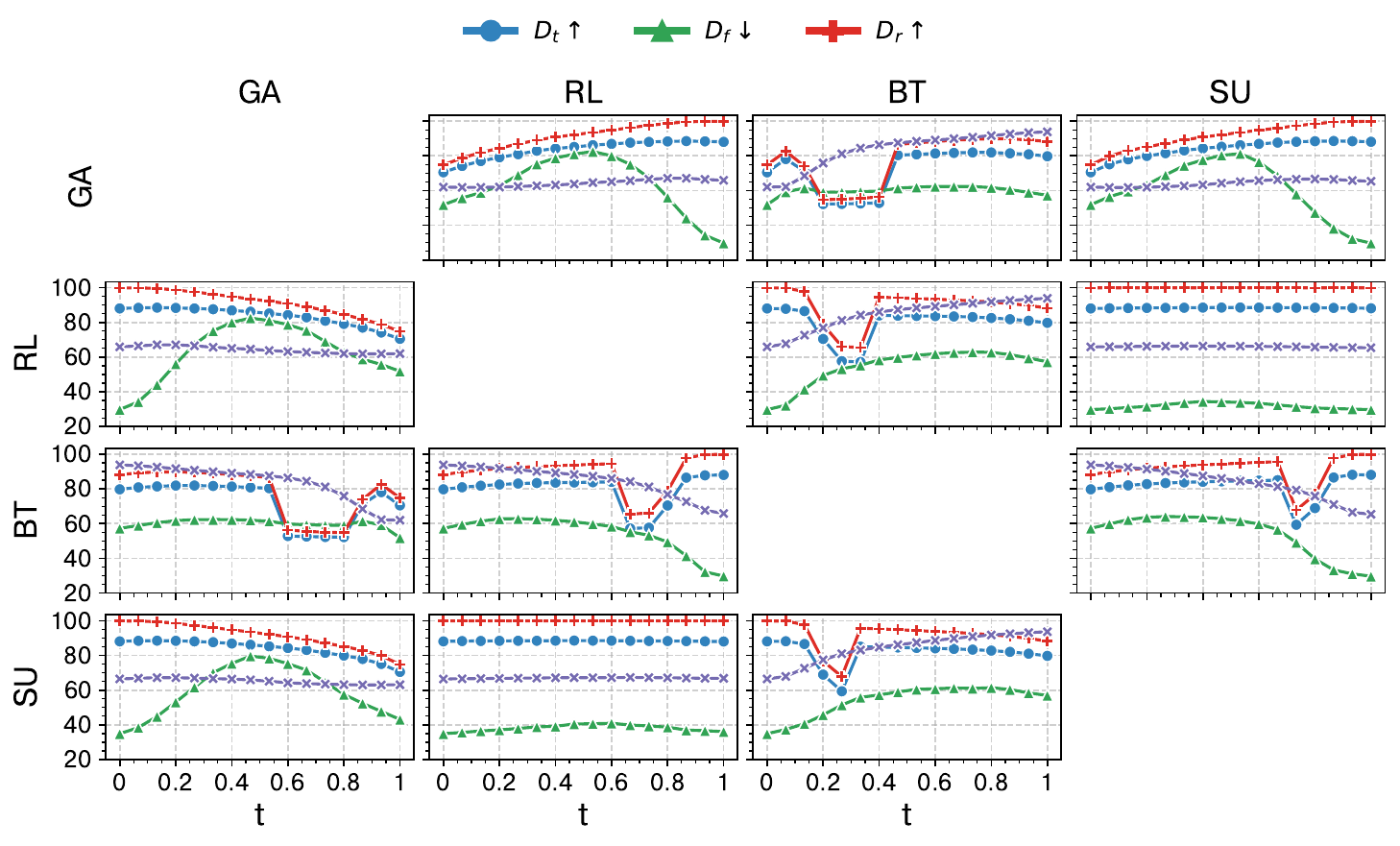}
        \caption{Linear MCU when $|D_f|=4.0\%$}
    \end{subfigure}
    \begin{subfigure}{0.49\textwidth}
        \centering
        \includegraphics[width=\linewidth]{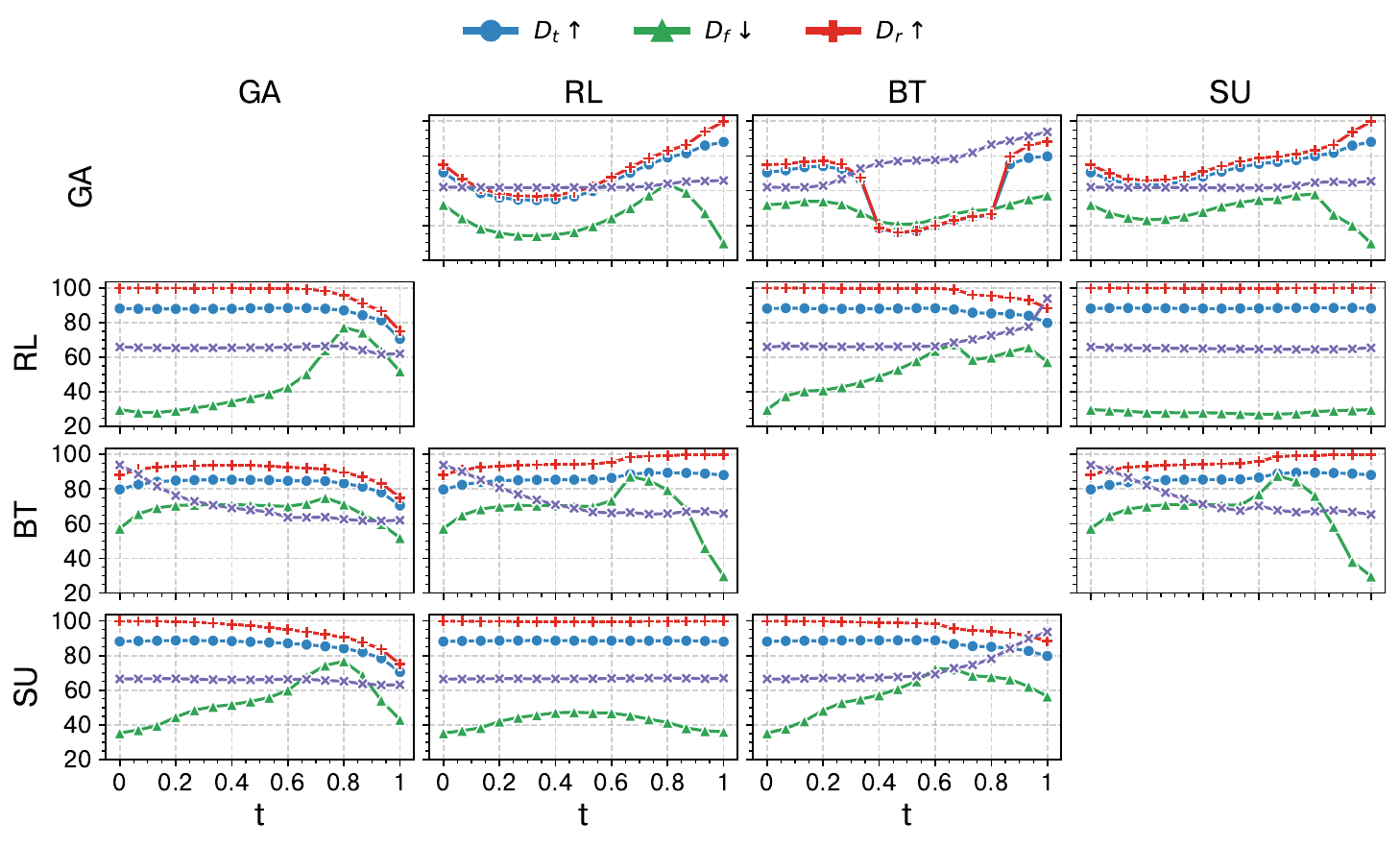}
        \caption{Quadratic MCU when $|D_f|=4.0\%$}
    \end{subfigure}
\end{center}
\end{figure}
\begin{figure}
\begin{center}
    \begin{subfigure}{0.49\textwidth}
        \centering
        \includegraphics[width=\linewidth]{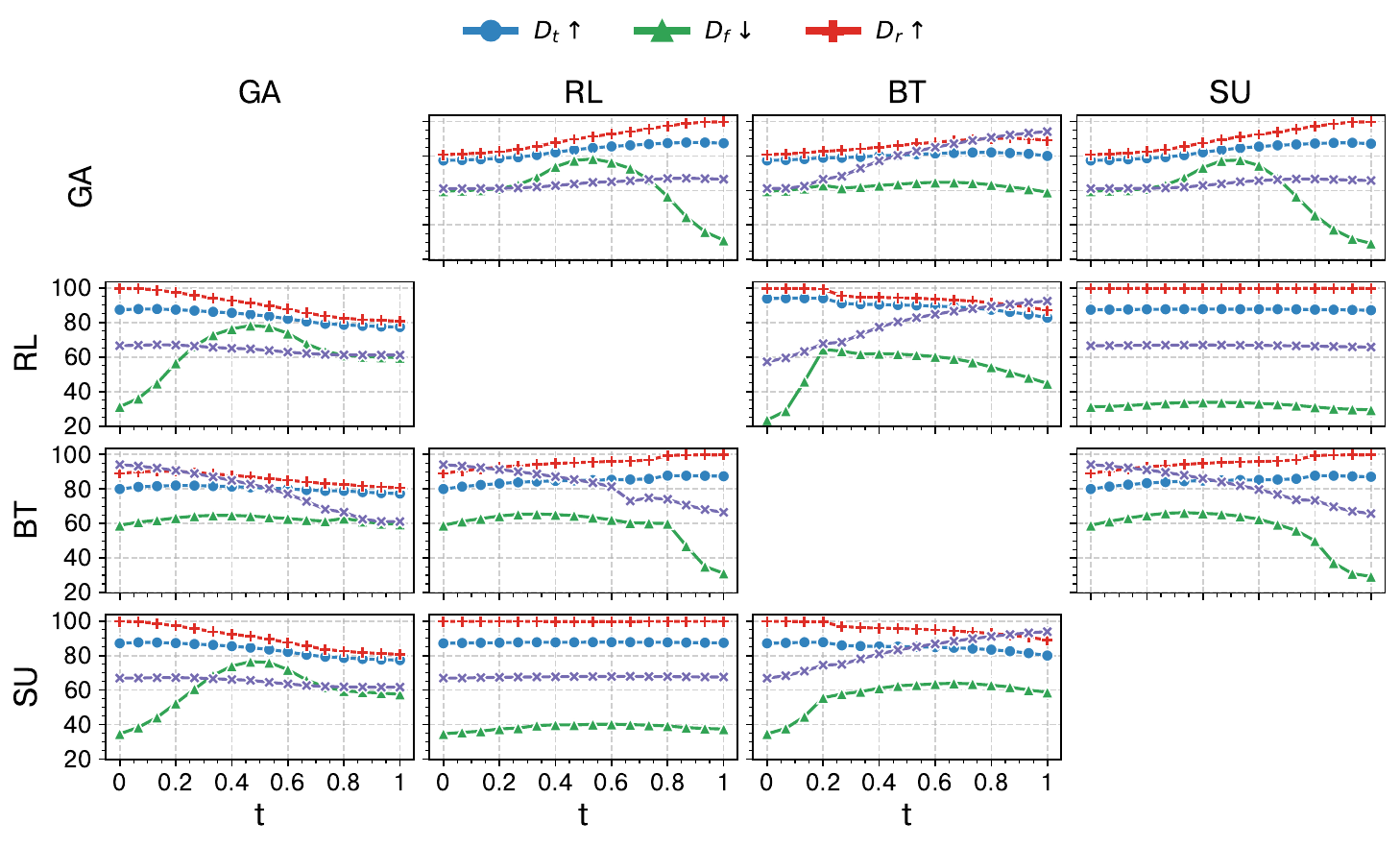}
        \caption{Linear MCU when $|D_f|=6.0\%$}
    \end{subfigure}
    \begin{subfigure}{0.49\textwidth}
        \centering
        \includegraphics[width=\linewidth]{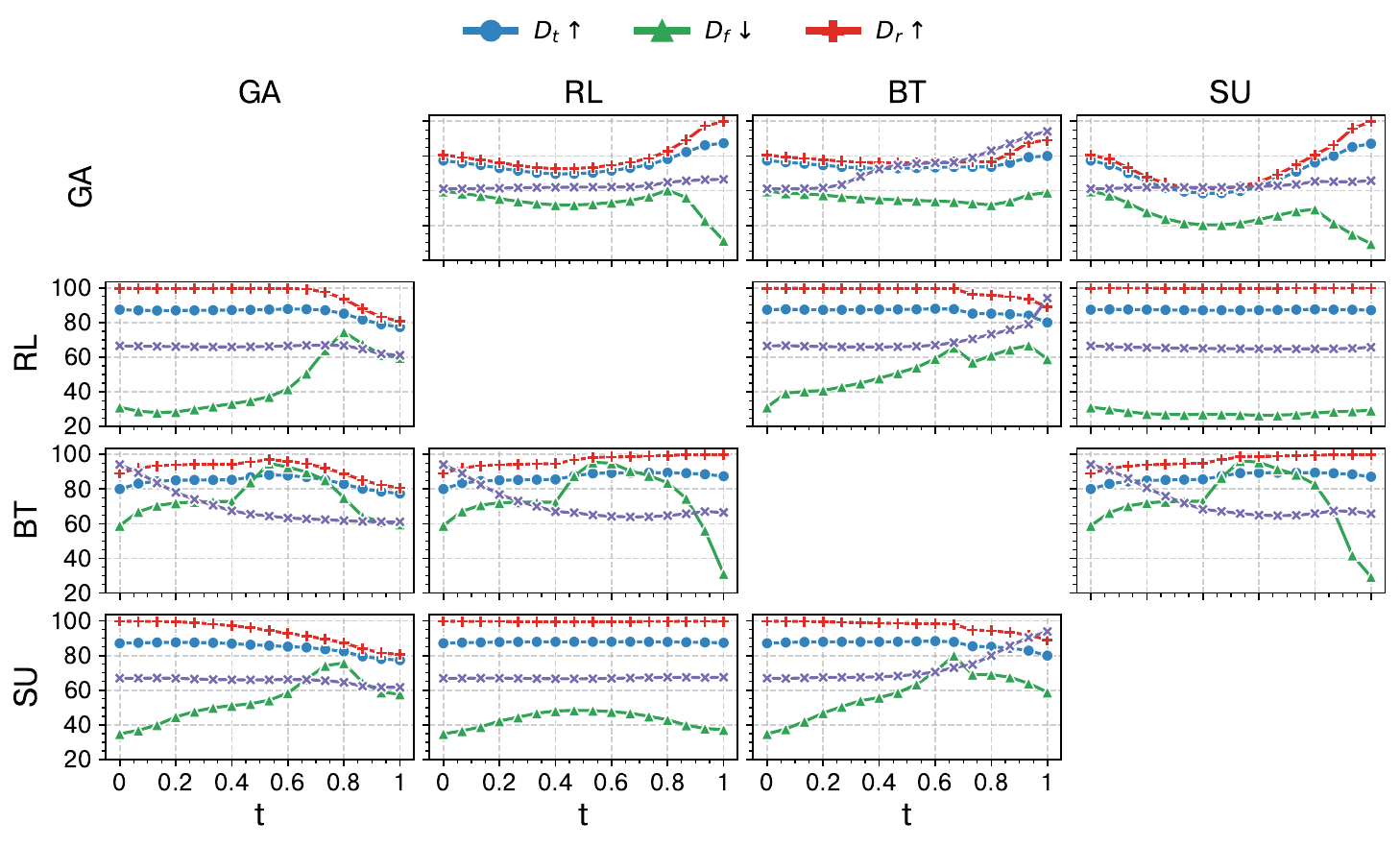}
        \caption{Quadratic MCU when $|D_f|=6.0\%$}
    \end{subfigure}
    \begin{subfigure}{0.49\textwidth}
        \centering
        \includegraphics[width=\linewidth]{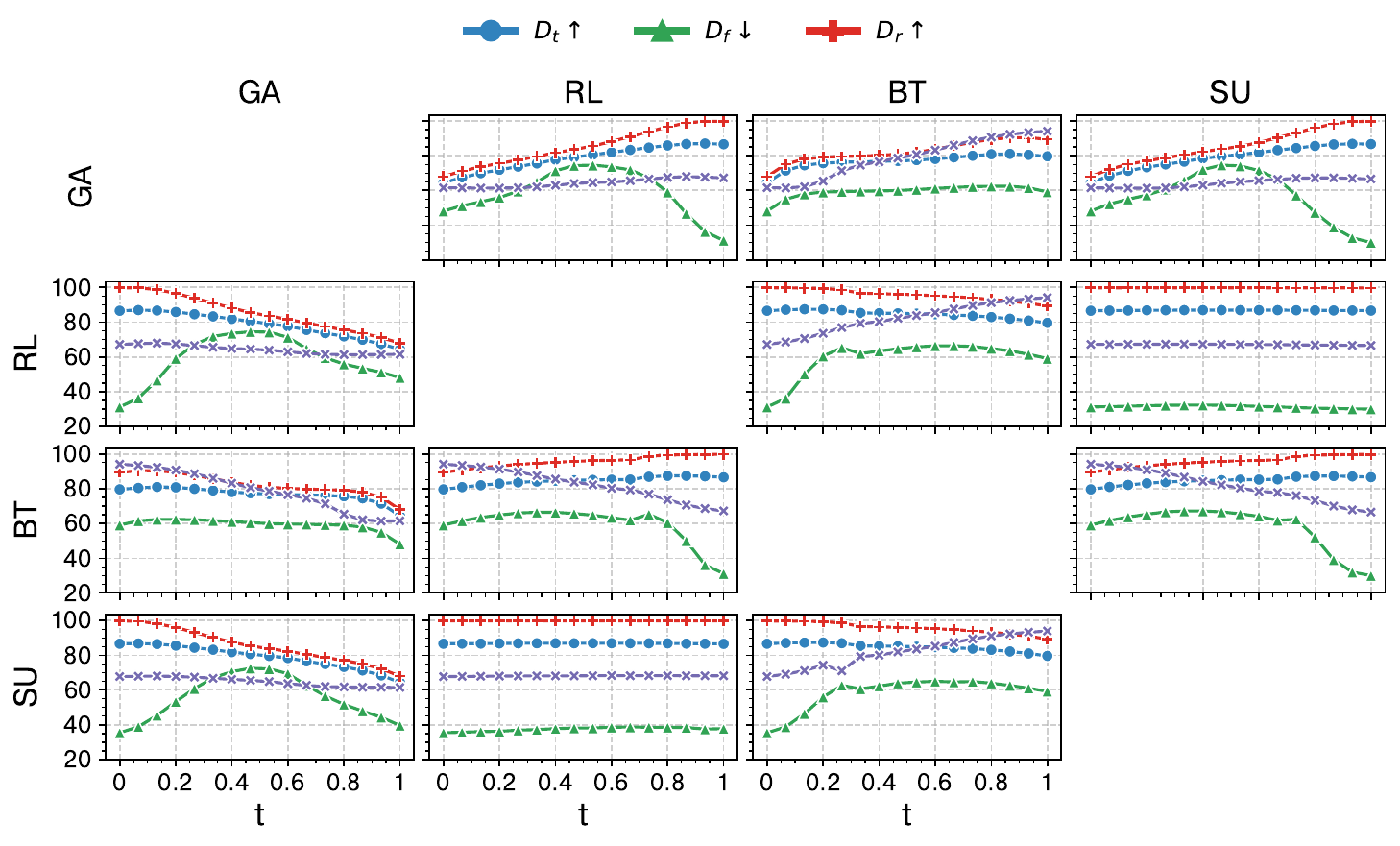}
        \caption{Linear MCU when $|D_f|=8.0\%$}
    \end{subfigure}
    \begin{subfigure}{0.49\textwidth}
        \centering
        \includegraphics[width=\linewidth]{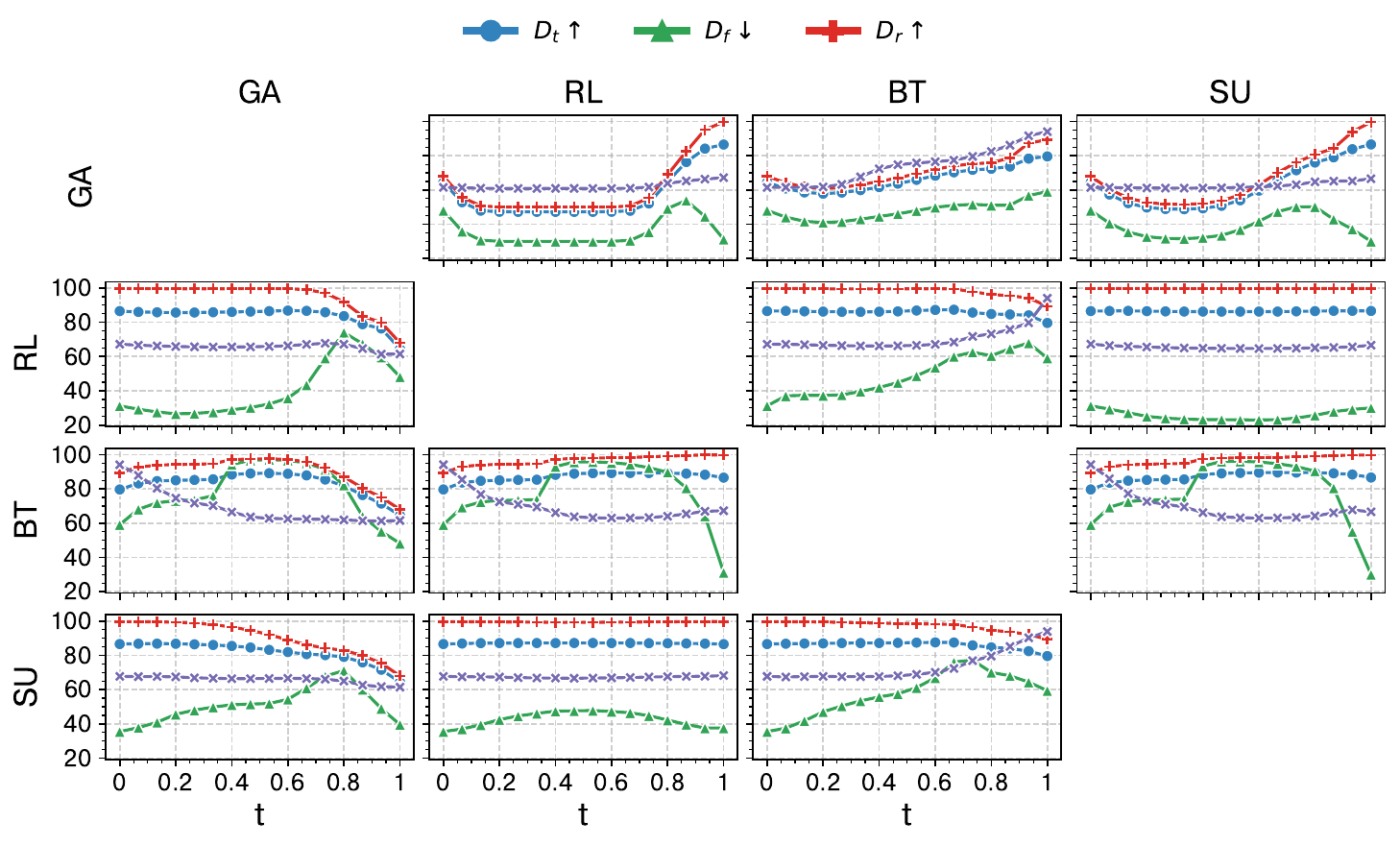}
        \caption{Quadratic MCU when $|D_f|=8.0\%$}
    \end{subfigure}
    \begin{subfigure}{0.49\textwidth}
        \centering
        \includegraphics[width=\linewidth]{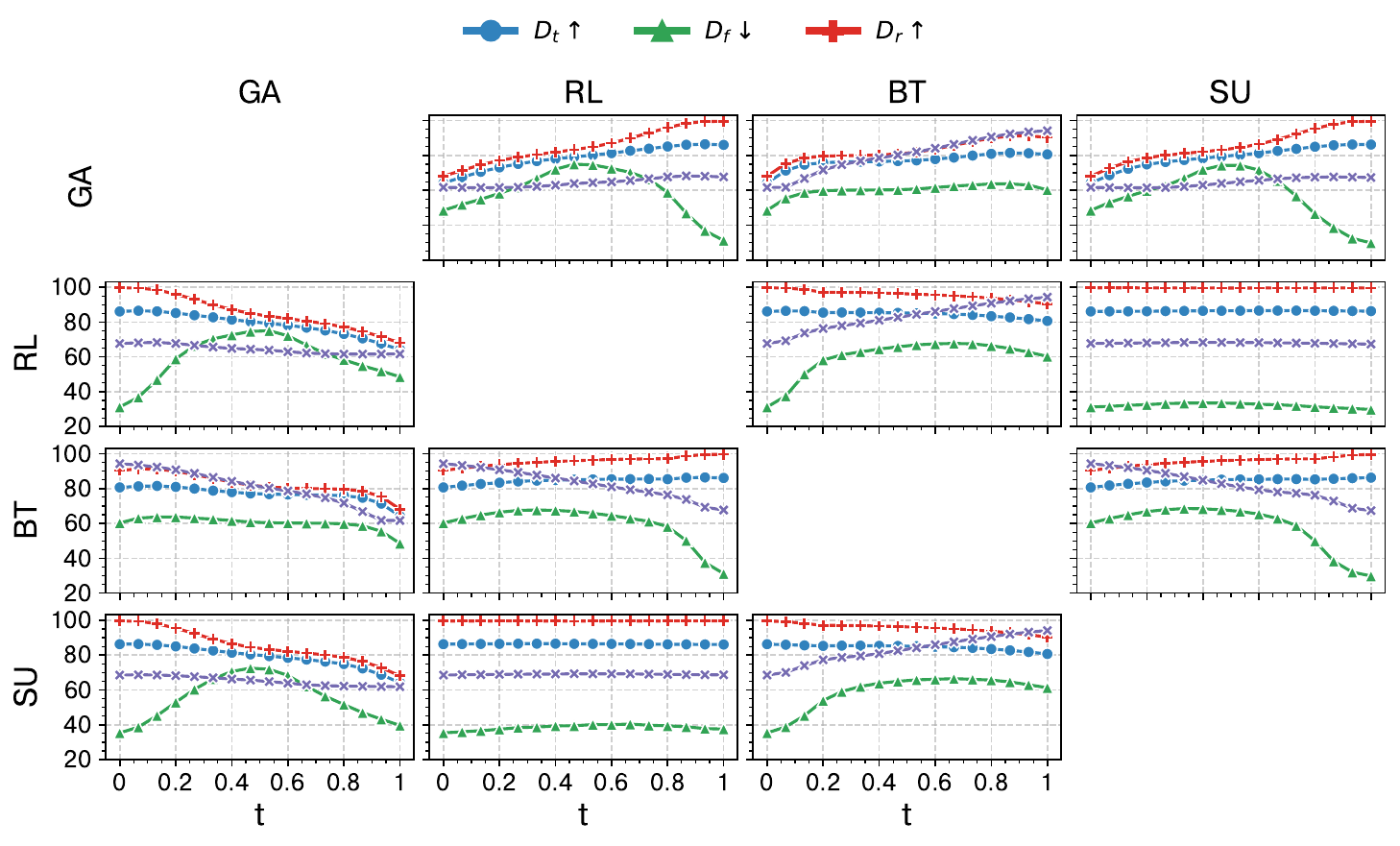}
        \caption{Linear MCU when $|D_f|=10.0\%$}
    \end{subfigure}
    \begin{subfigure}{0.49\textwidth}
        \centering
        \includegraphics[width=\linewidth]{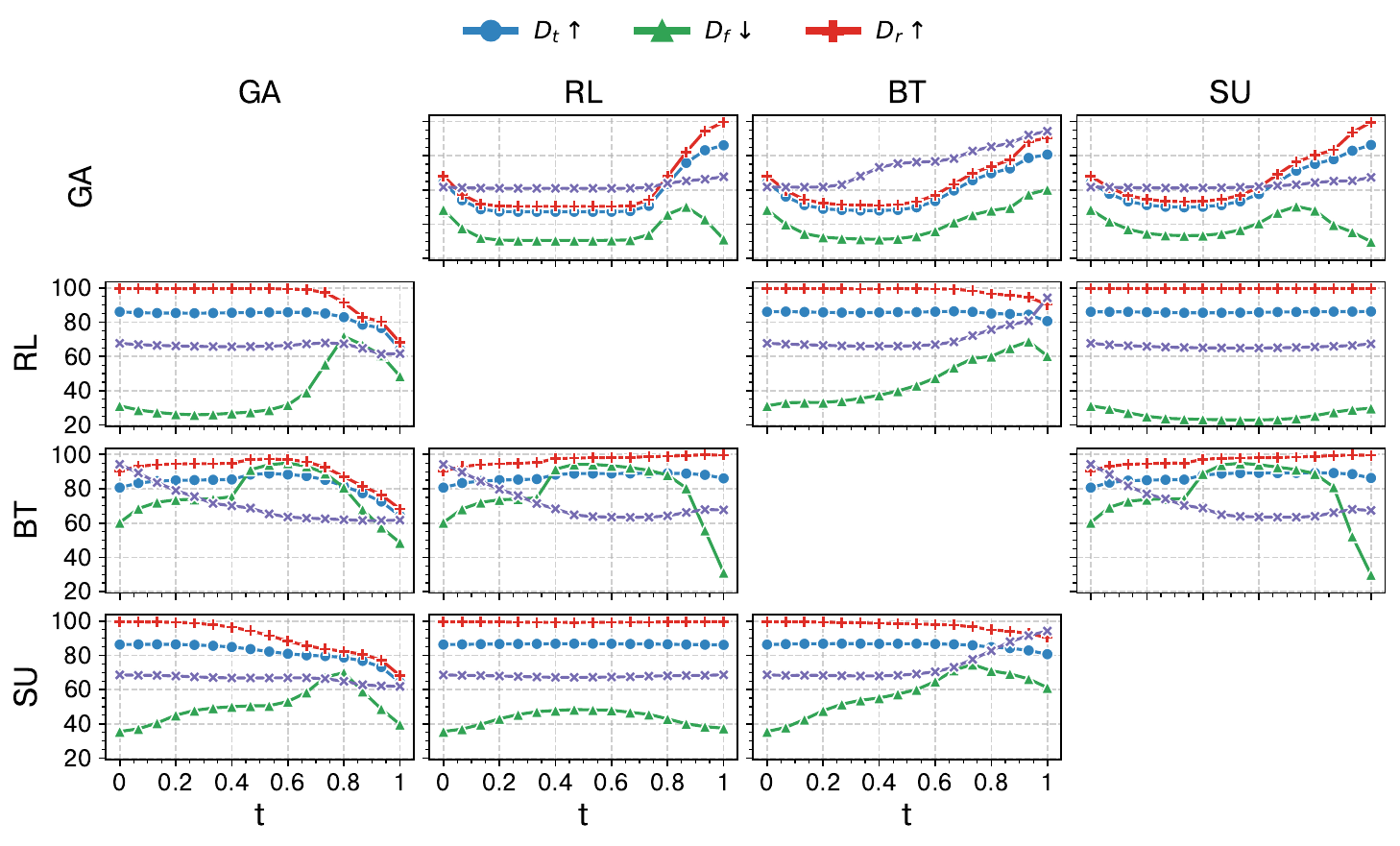}
        \caption{Quadratic MCU when $|D_f|=10.0\%$}
    \end{subfigure}
\end{center}
\caption{MCU under \textbf{Met-CL} setting on \textbf{classification datasets}.}
\label{fig:cls-met-cl}
\end{figure}

\begin{figure}
\begin{center}
    \begin{subfigure}{0.49\textwidth}
        \includegraphics[width=\linewidth]{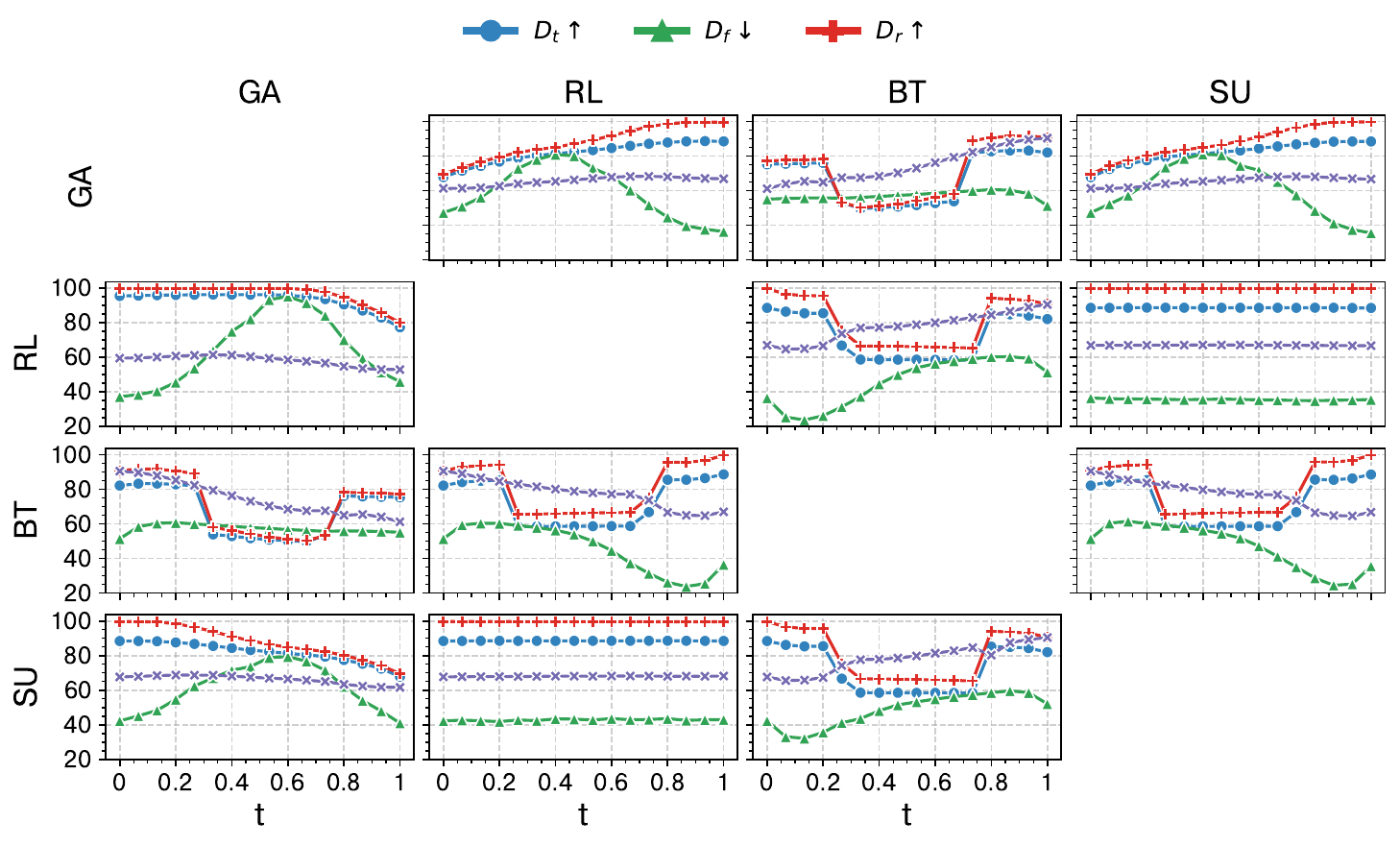}
        \caption{Linear MCU when $|D_f|=2.0\%$}
    \end{subfigure}
    \begin{subfigure}{0.49\textwidth}
        \includegraphics[width=\linewidth]{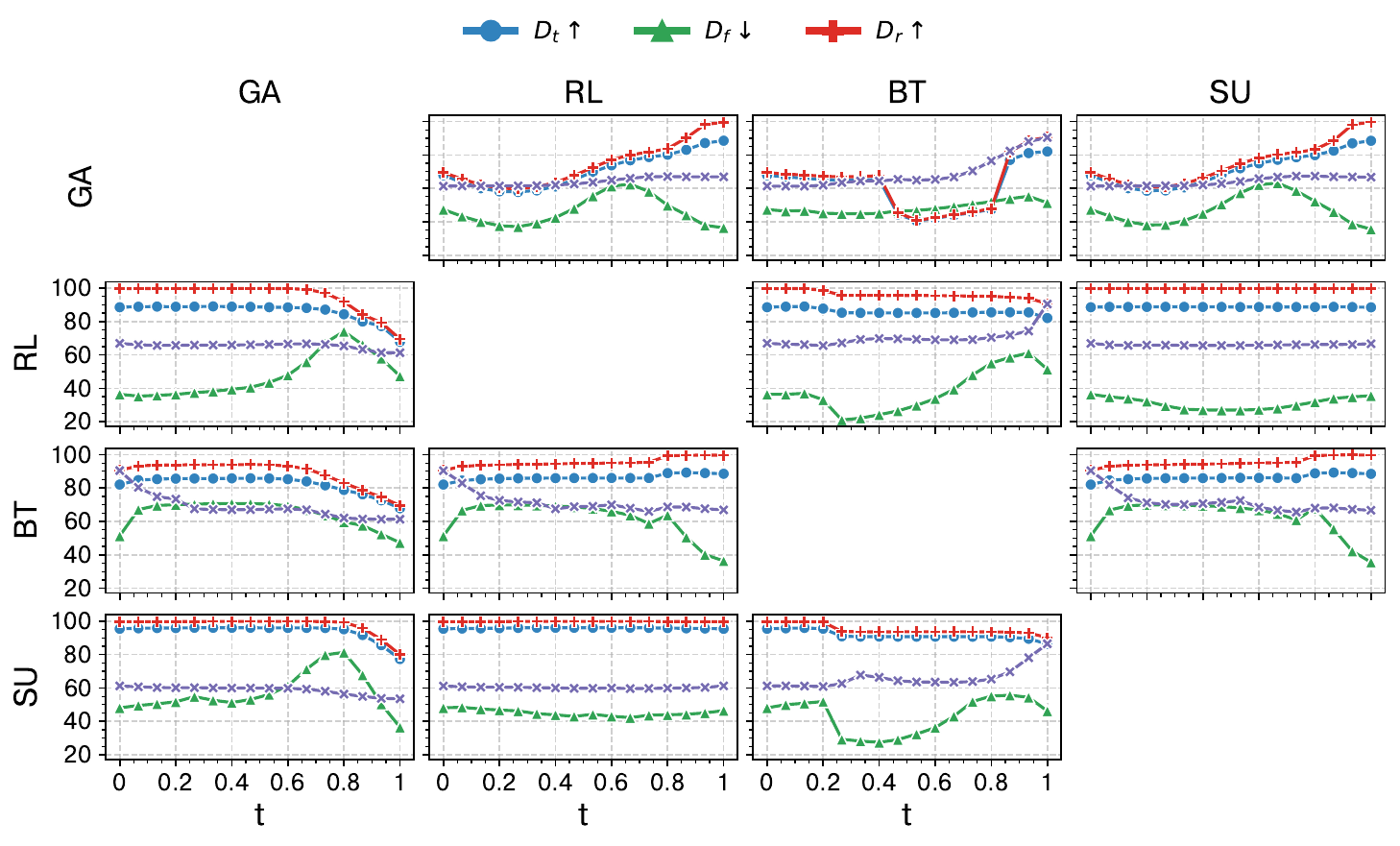}
        \caption{Quadratic MCU when $|D_f|=2.0\%$}
    \end{subfigure}
    \begin{subfigure}{0.49\textwidth}
        \centering
        \includegraphics[width=\linewidth]{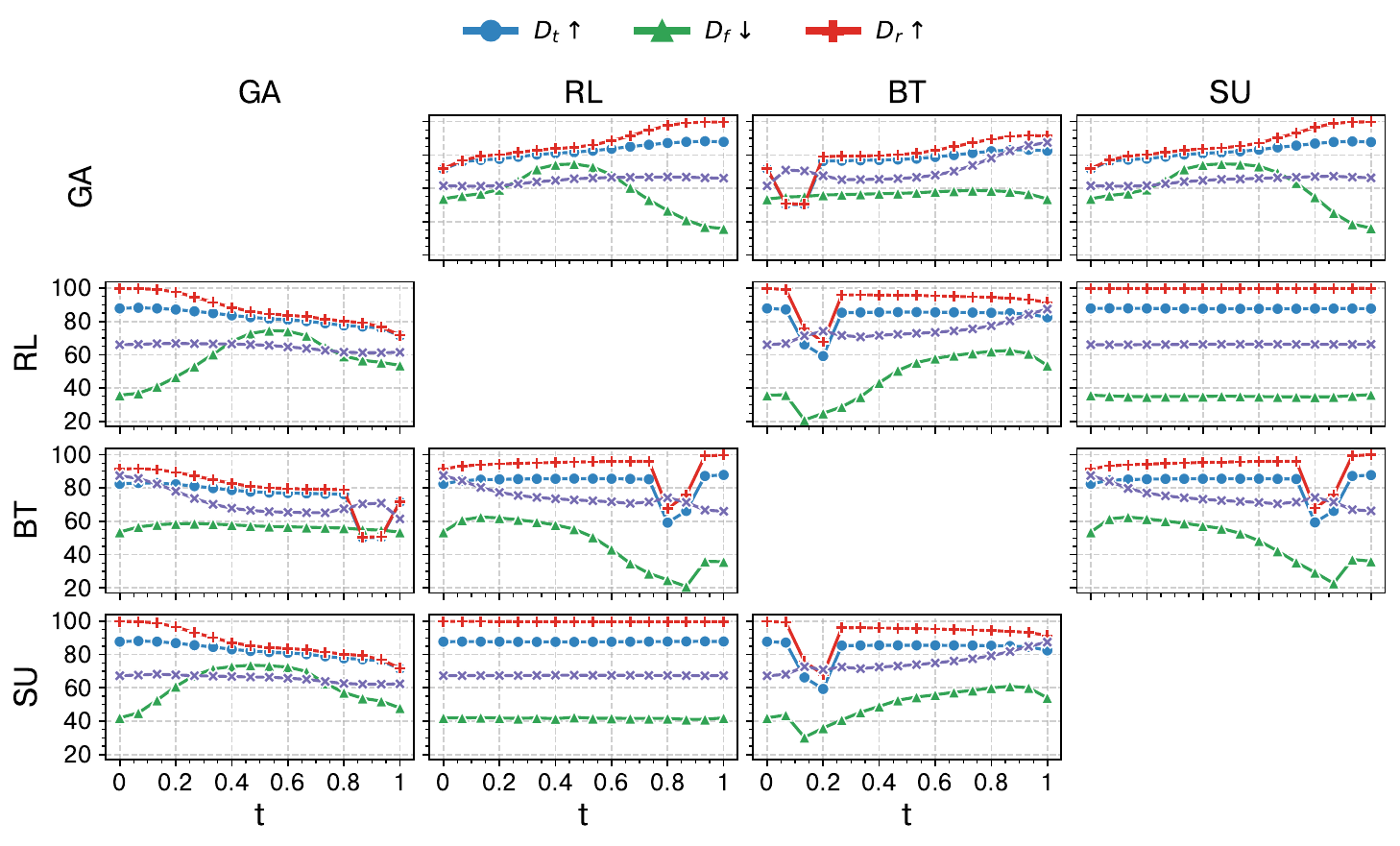}
        \caption{Linear MCU when $|D_f|=4.0\%$}
    \end{subfigure}
    \begin{subfigure}{0.49\textwidth}
        \centering
        \includegraphics[width=\linewidth]{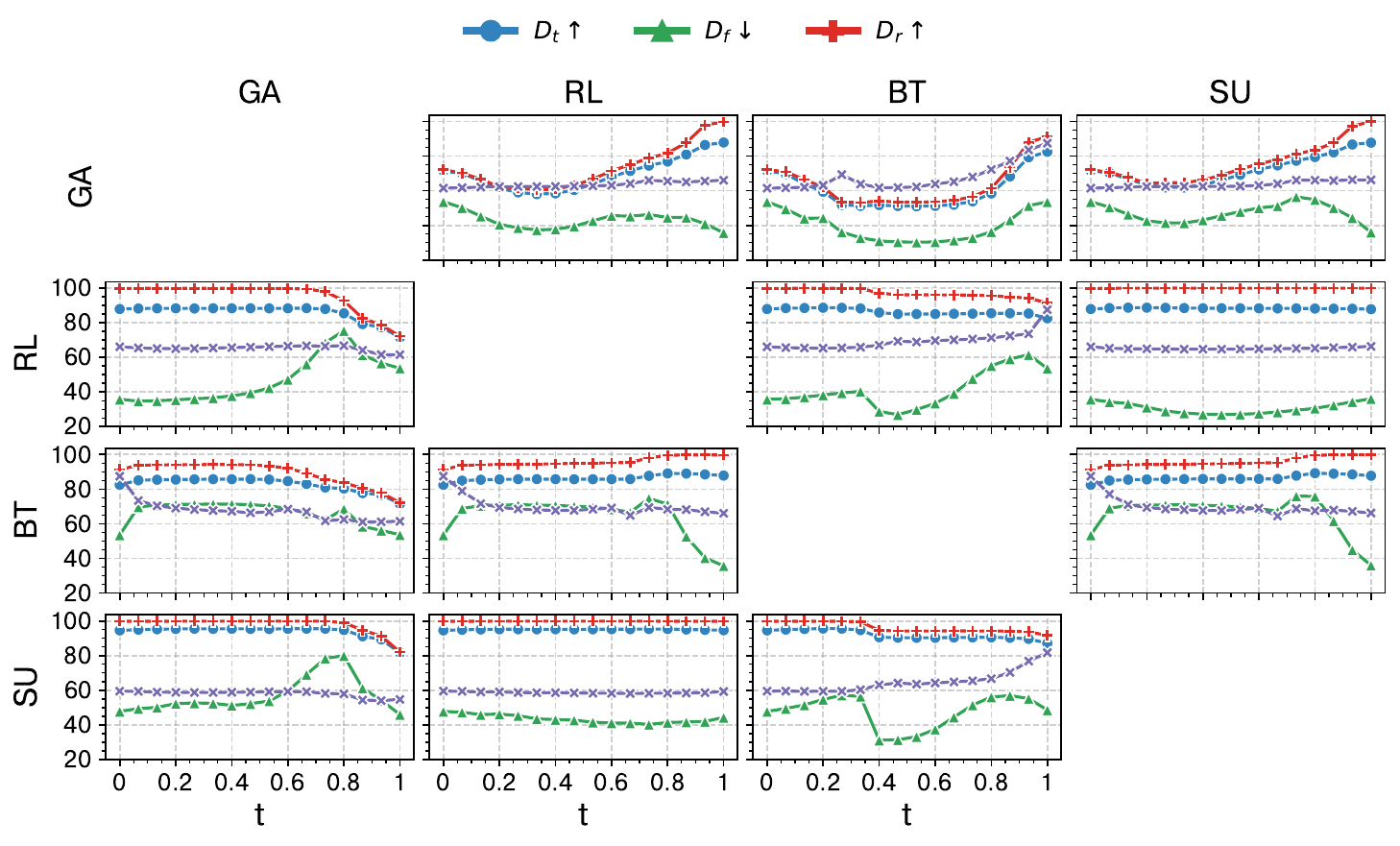}
        \caption{Quadratic MCU when $|D_f|=4.0\%$}
    \end{subfigure}
\end{center}
\end{figure}
\begin{figure}
\begin{center}
    \begin{subfigure}{0.49\textwidth}
        \centering
        \includegraphics[width=\linewidth]{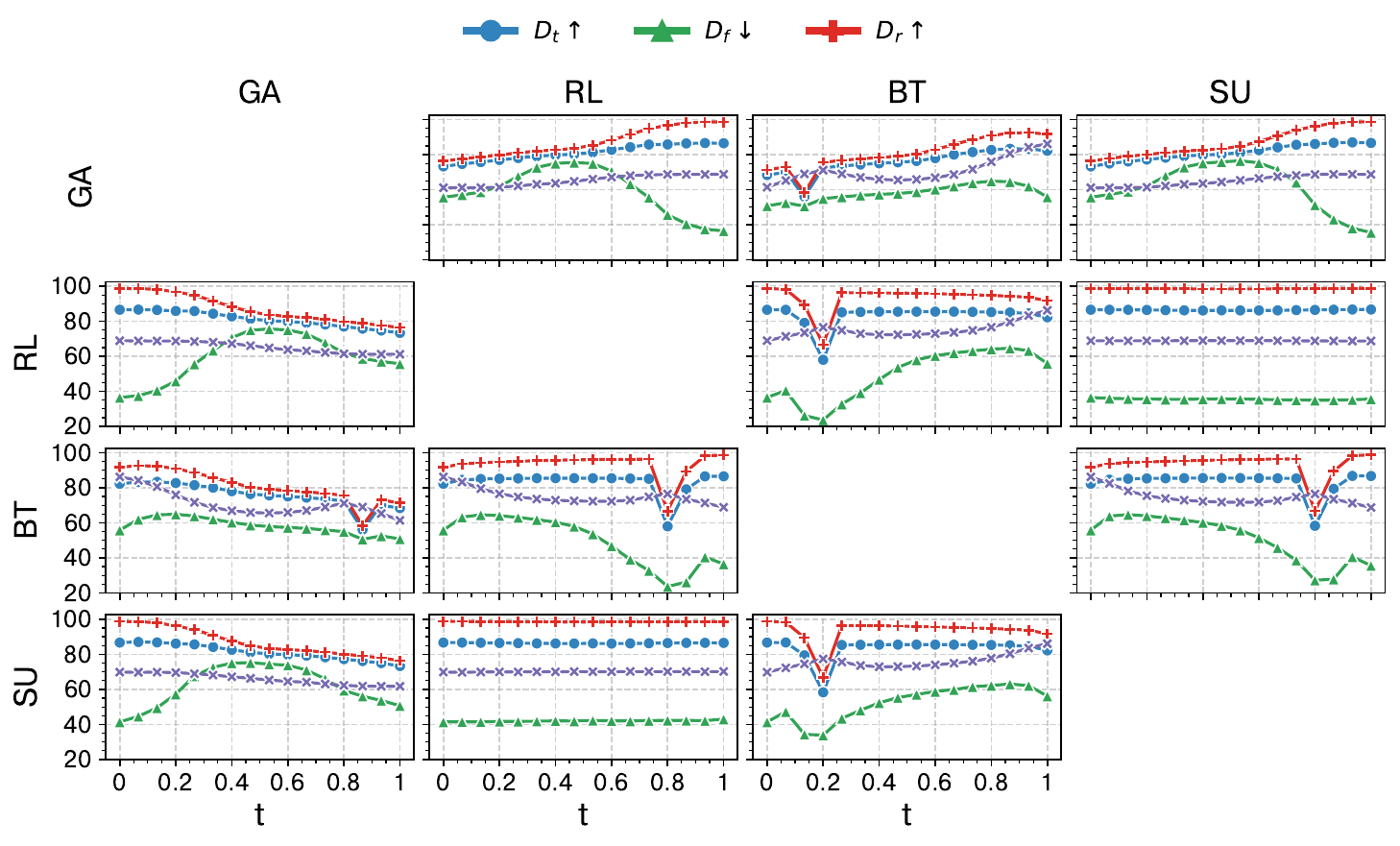}
        \caption{Linear MCU when $|D_f|=6.0\%$}
    \end{subfigure}
    \begin{subfigure}{0.49\textwidth}
        \centering
        \includegraphics[width=\linewidth]{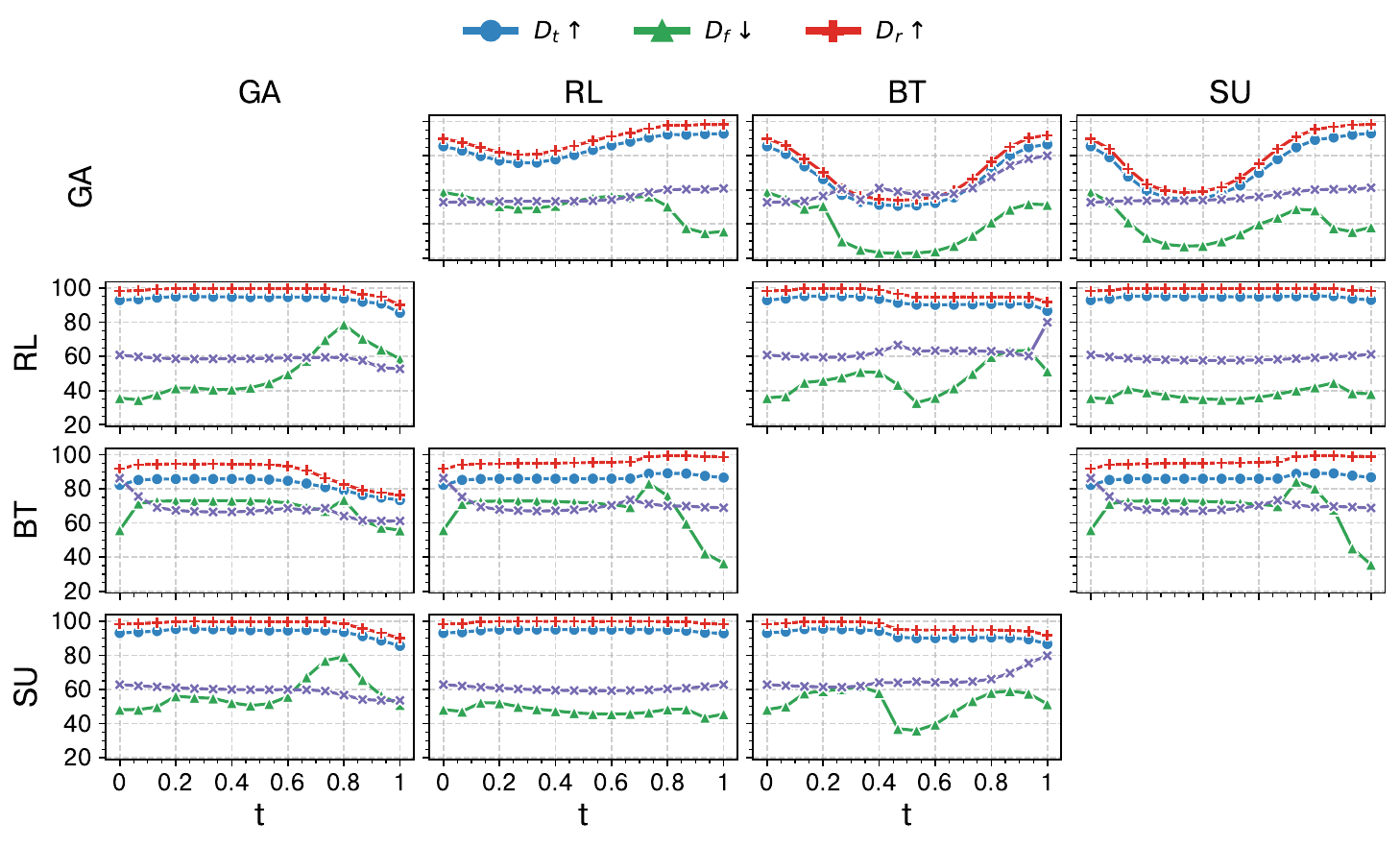}
        \caption{Quadratic MCU when $|D_f|=6.0\%$}
    \end{subfigure}
    \begin{subfigure}{0.49\textwidth}
        \centering
        \includegraphics[width=\linewidth]{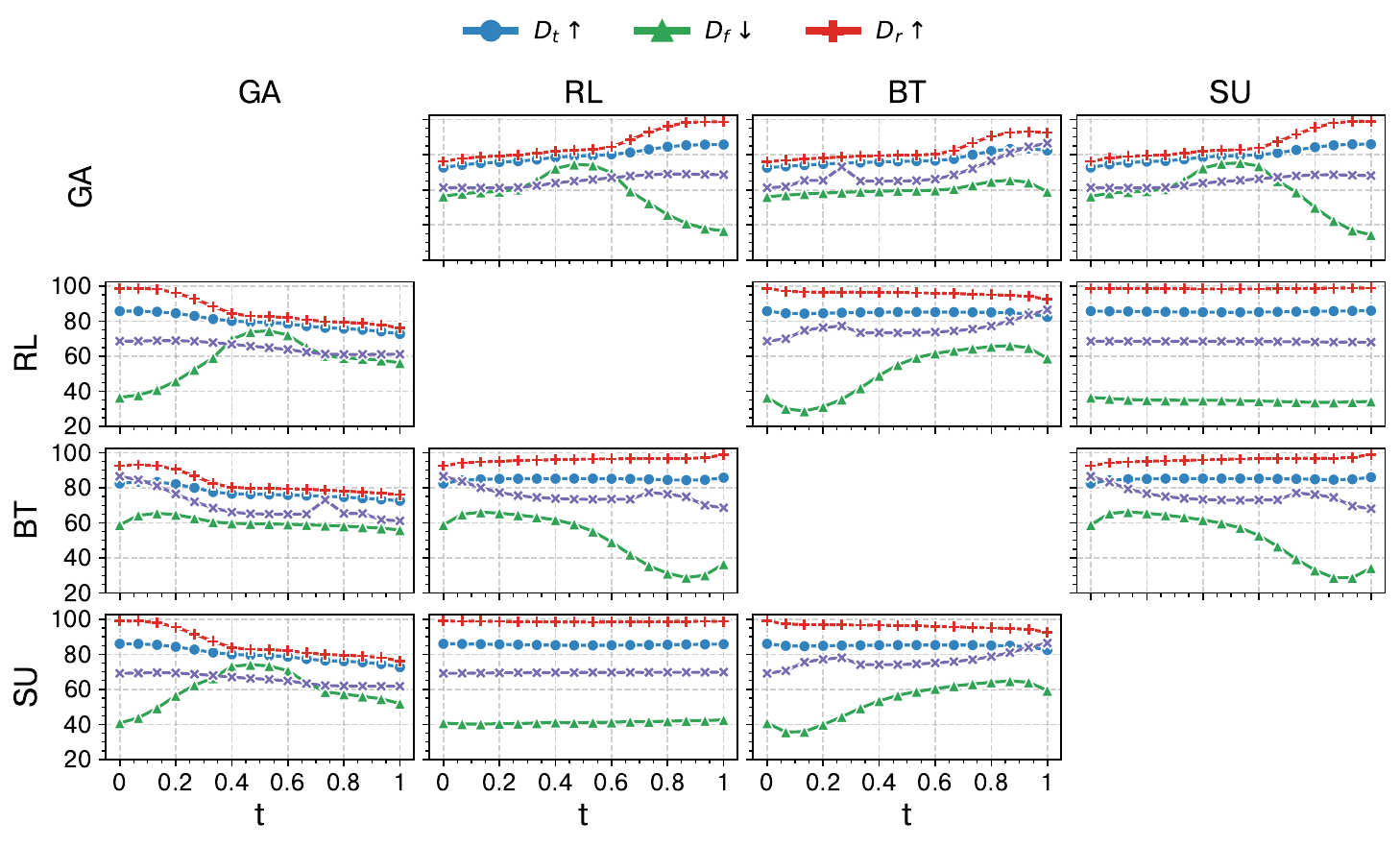}
        \caption{Linear MCU when $|D_f|=8.0\%$}
    \end{subfigure}
    \begin{subfigure}{0.49\textwidth}
        \centering
        \includegraphics[width=\linewidth]{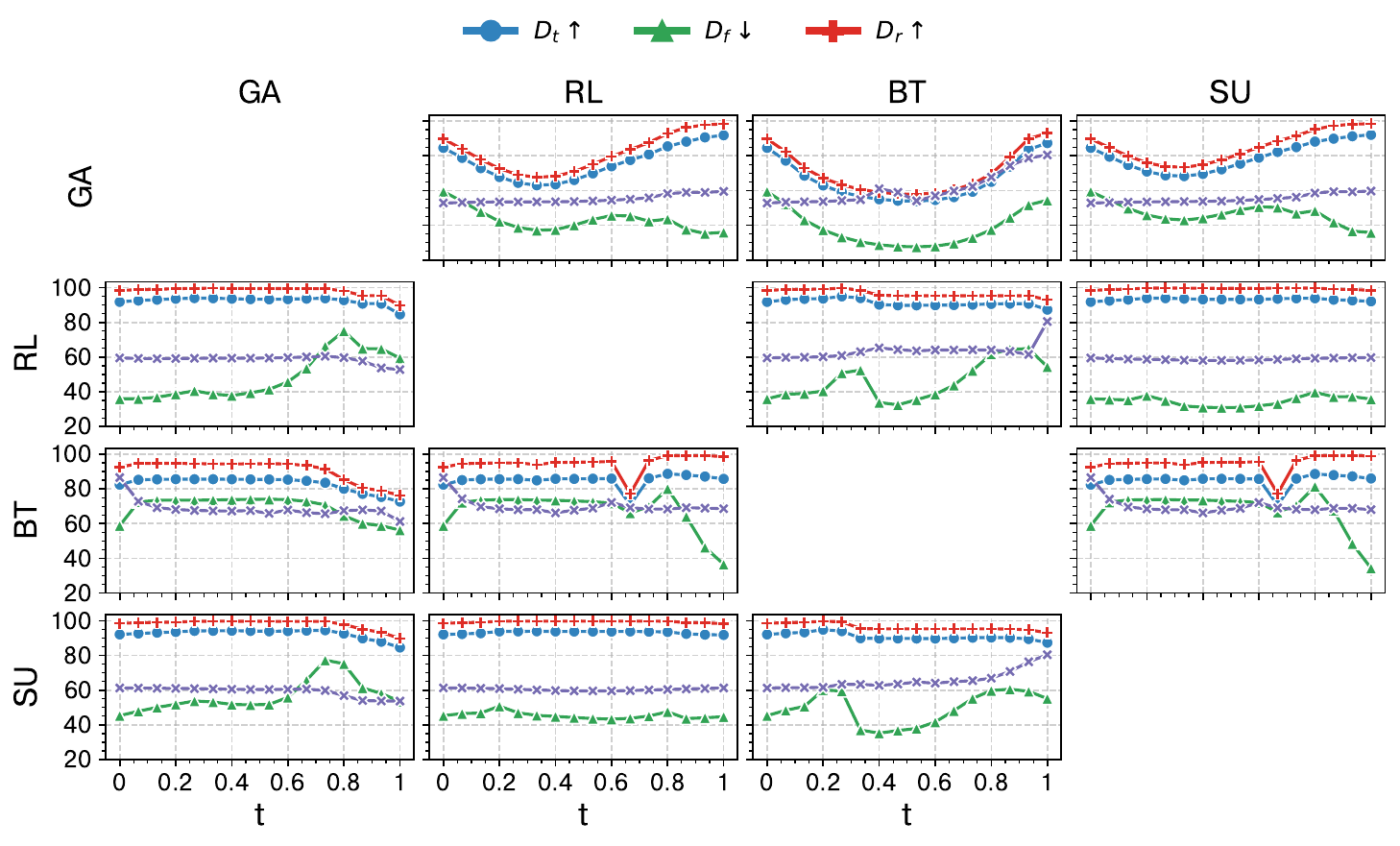}
        \caption{Quadratic MCU when $|D_f|=8.0\%$}
    \end{subfigure}
    \begin{subfigure}{0.49\textwidth}
        \centering
        \includegraphics[width=\linewidth]{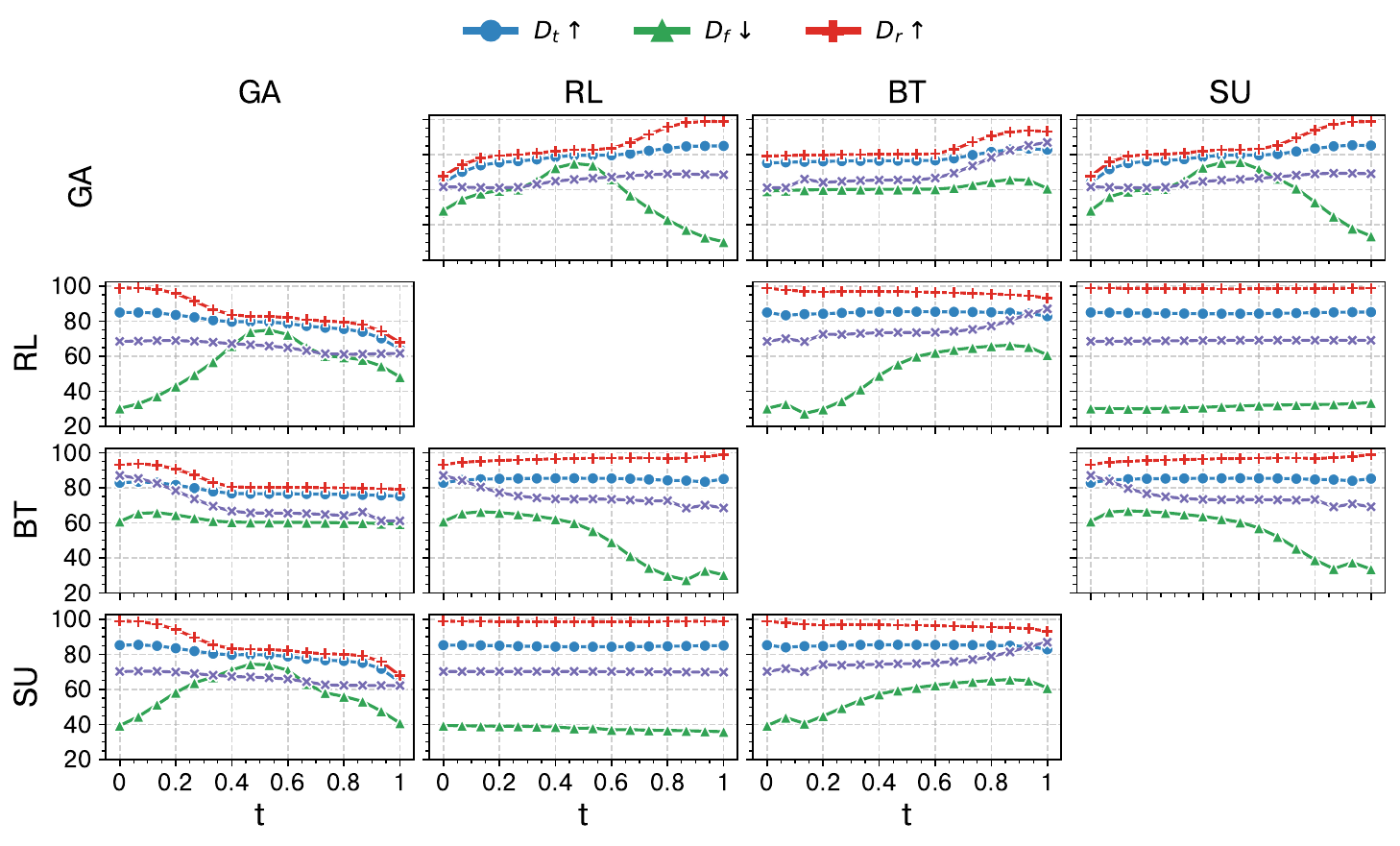}
        \caption{Linear MCU when $|D_f|=10.0\%$}
    \end{subfigure}
    \begin{subfigure}{0.49\textwidth}
        \centering
        \includegraphics[width=\linewidth]{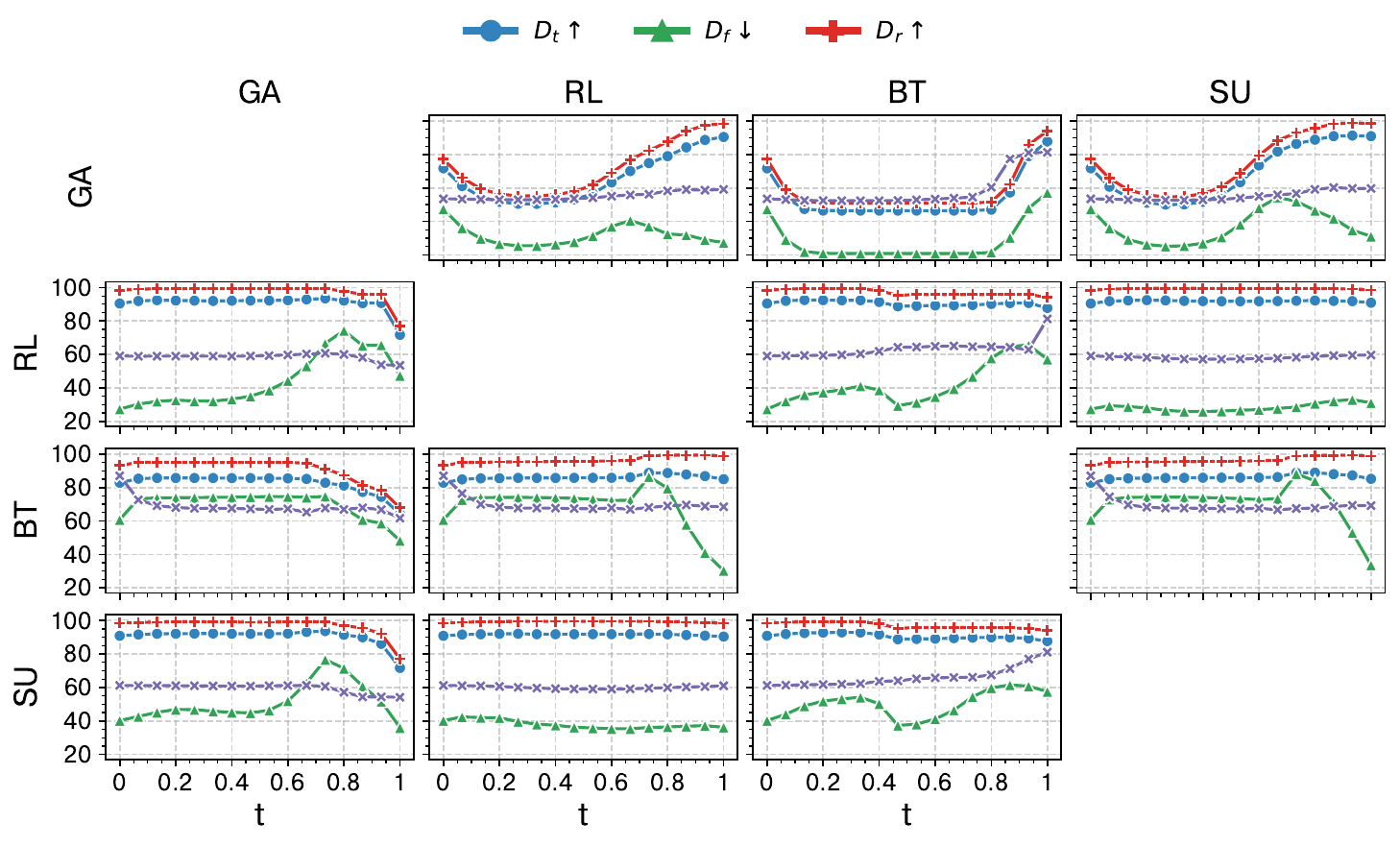}
        \caption{Quadratic MCU when $|D_f|=10.0\%$}
    \end{subfigure}
\end{center}
\caption{MCU under \textbf{Met-SO} setting on \textbf{classification datasets}.}
\label{fig:cls-met-so}
\end{figure}

\begin{figure}
\begin{center}
    \begin{subfigure}{0.49\textwidth}
        \includegraphics[width=\linewidth]{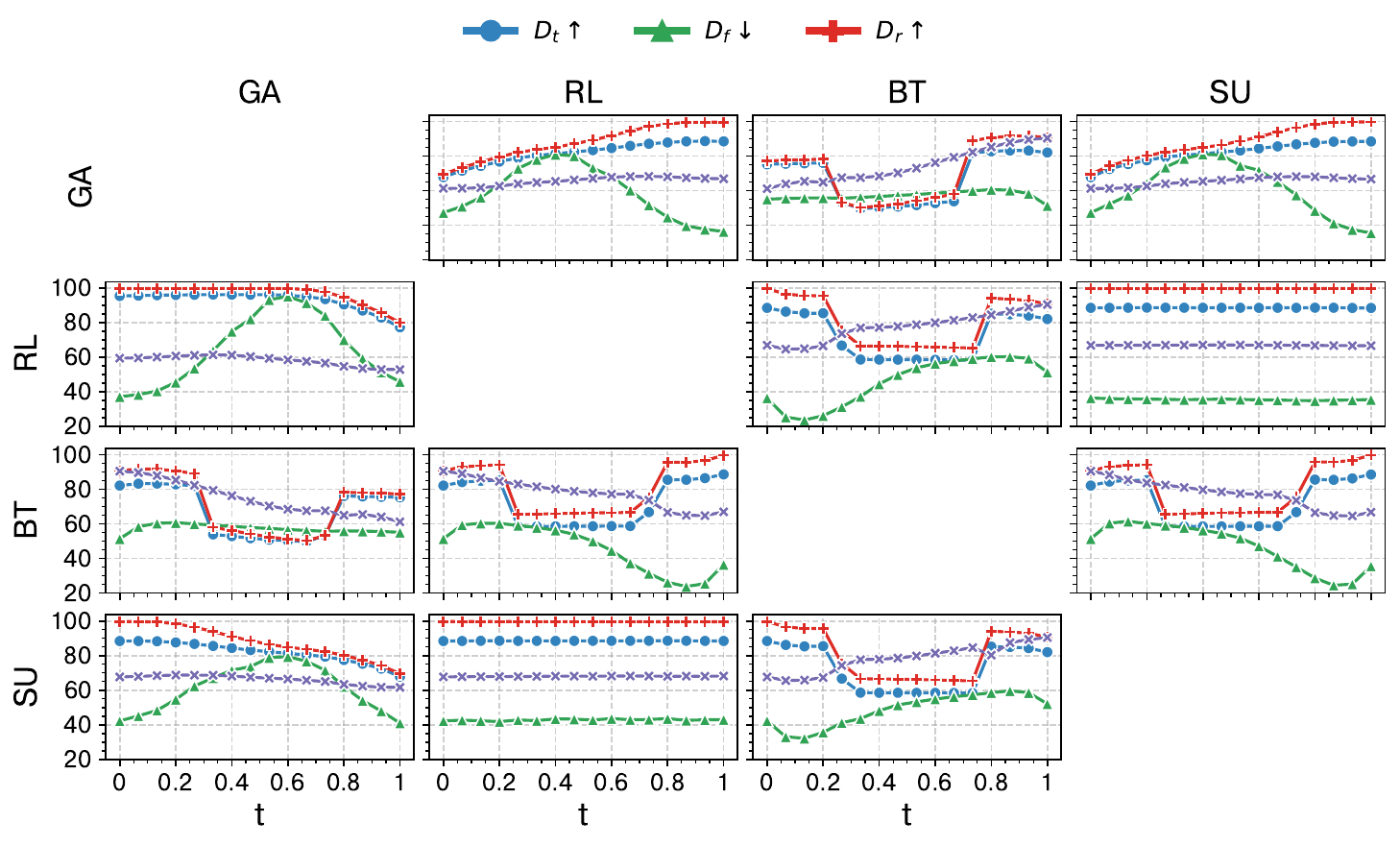}
        \caption{Linear MCU when $|D_f|=2.0\%$}
    \end{subfigure}
    \begin{subfigure}{0.49\textwidth}
        \includegraphics[width=\linewidth]{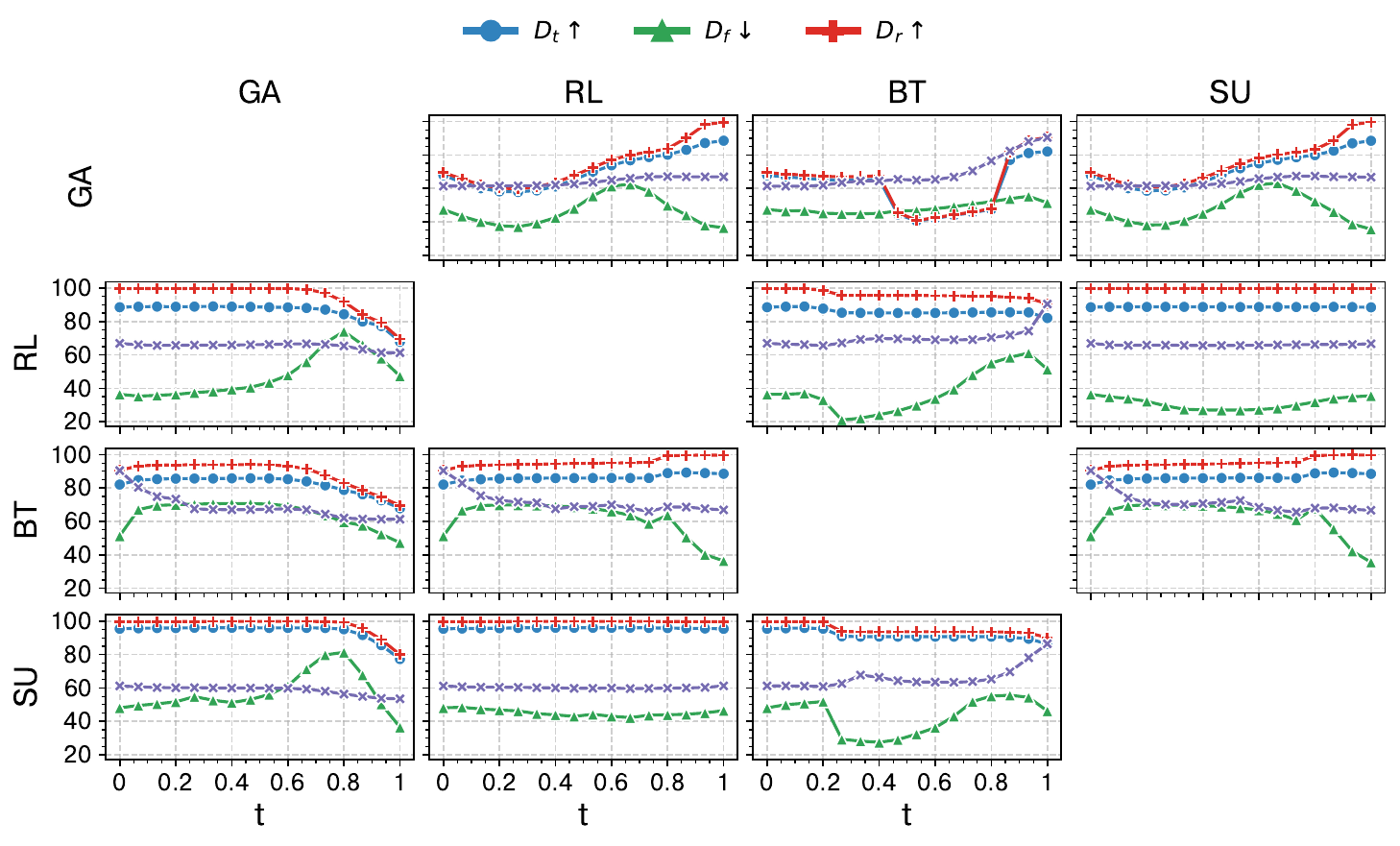}
        \caption{Quadratic MCU when $|D_f|=2.0\%$}
    \end{subfigure}
    \begin{subfigure}{0.49\textwidth}
        \centering
        \includegraphics[width=\linewidth]{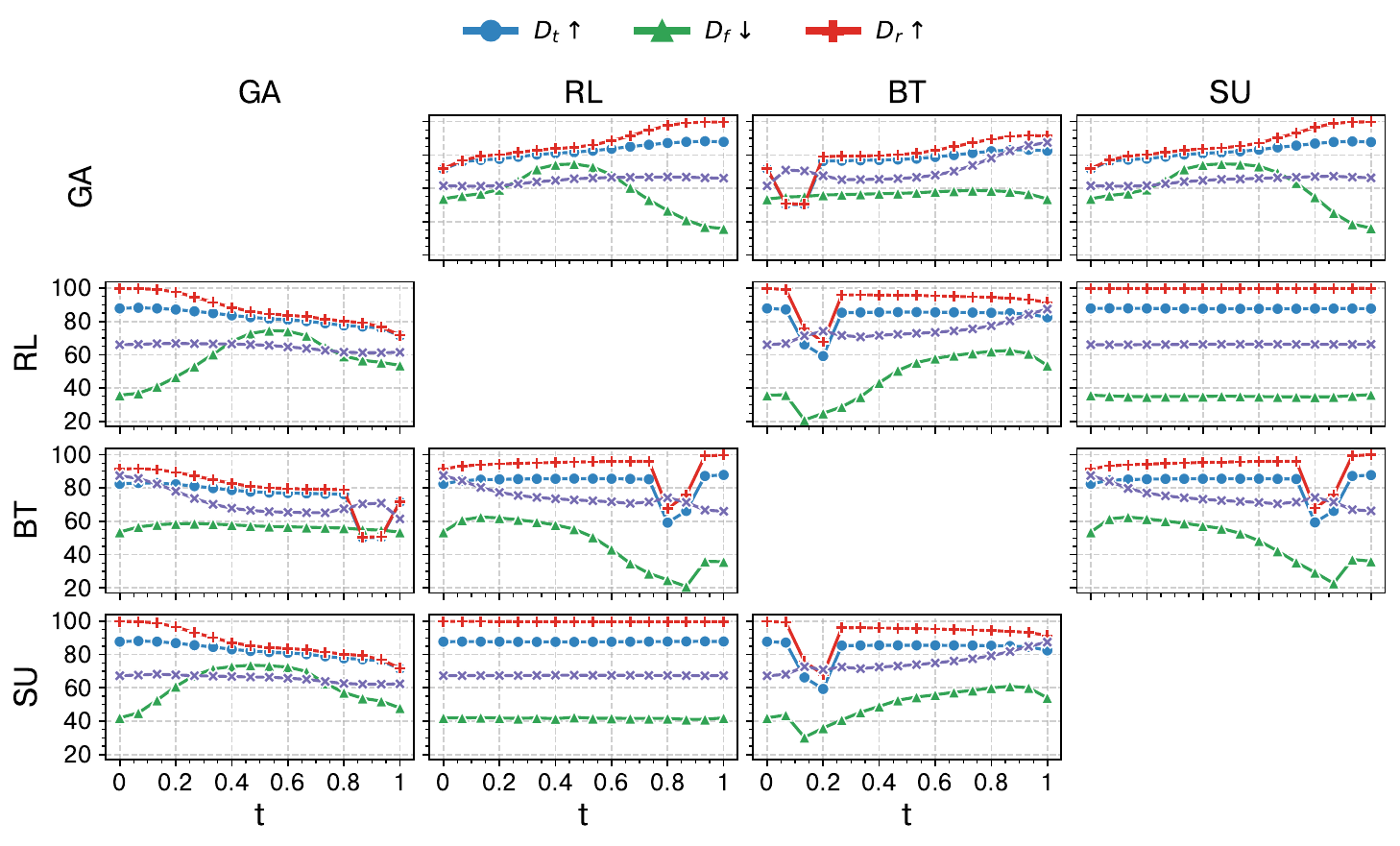}
        \caption{Linear MCU when $|D_f|=4.0\%$}
    \end{subfigure}
    \begin{subfigure}{0.49\textwidth}
        \centering
        \includegraphics[width=\linewidth]{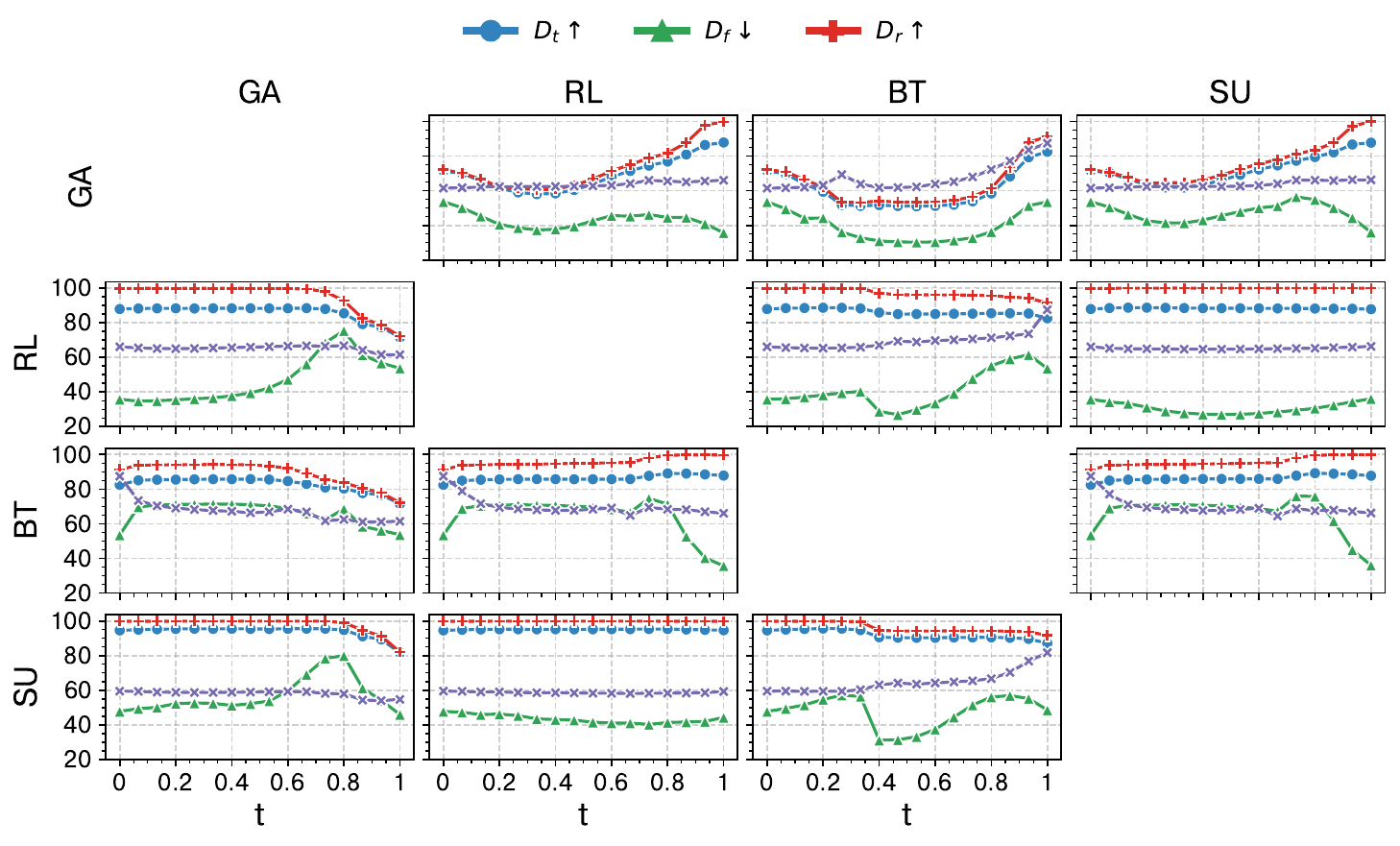}
        \caption{Quadratic MCU when $|D_f|=4.0\%$}
    \end{subfigure}
\end{center}
\end{figure}
\begin{figure}
\begin{center}
    \begin{subfigure}{0.49\textwidth}
        \centering
        \includegraphics[width=\linewidth]{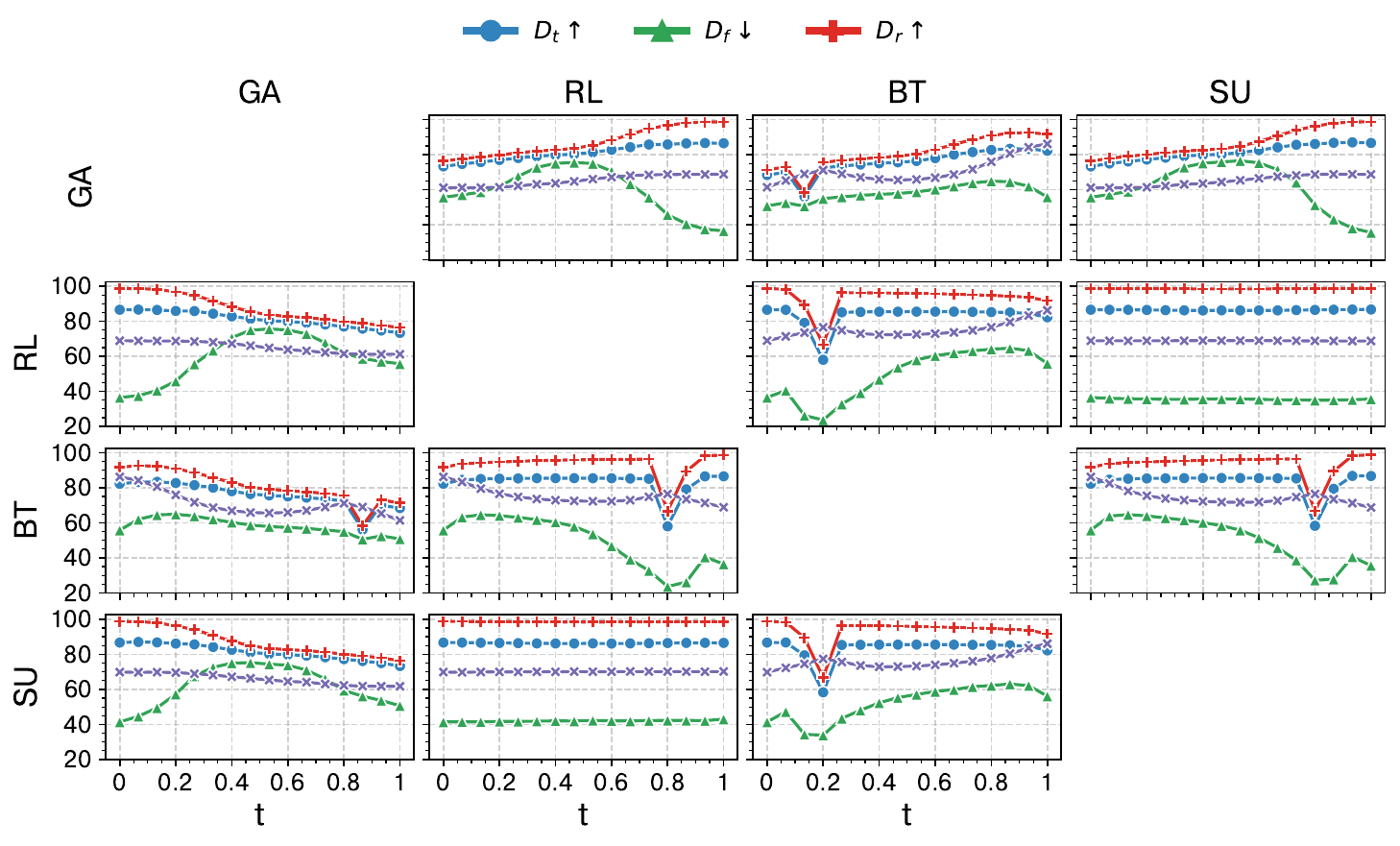}
        \caption{Linear MCU when $|D_f|=6.0\%$}
    \end{subfigure}
    \begin{subfigure}{0.49\textwidth}
        \centering
        \includegraphics[width=\linewidth]{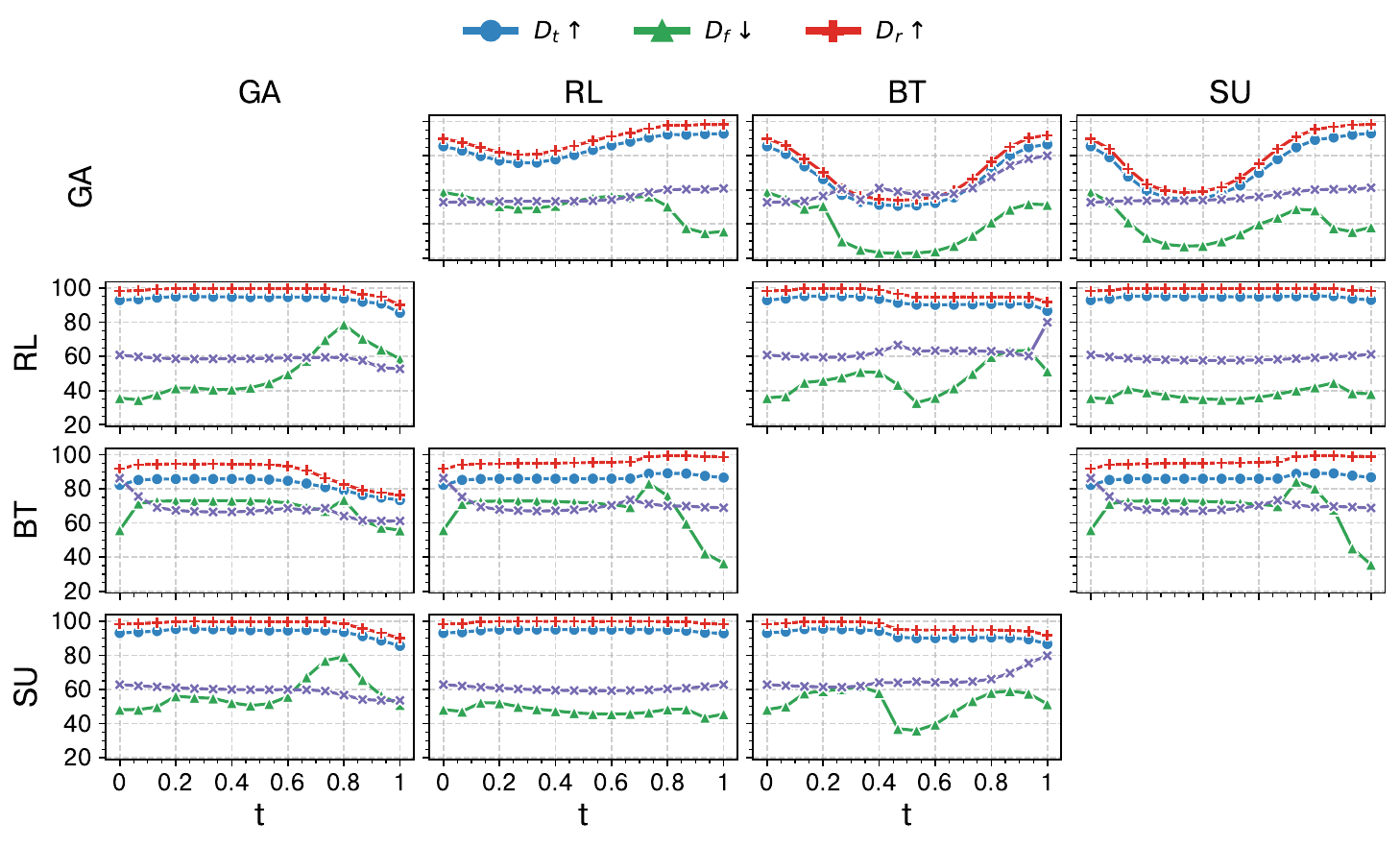}
        \caption{Quadratic MCU when $|D_f|=6.0\%$}
    \end{subfigure}
    \begin{subfigure}{0.49\textwidth}
        \centering
        \includegraphics[width=\linewidth]{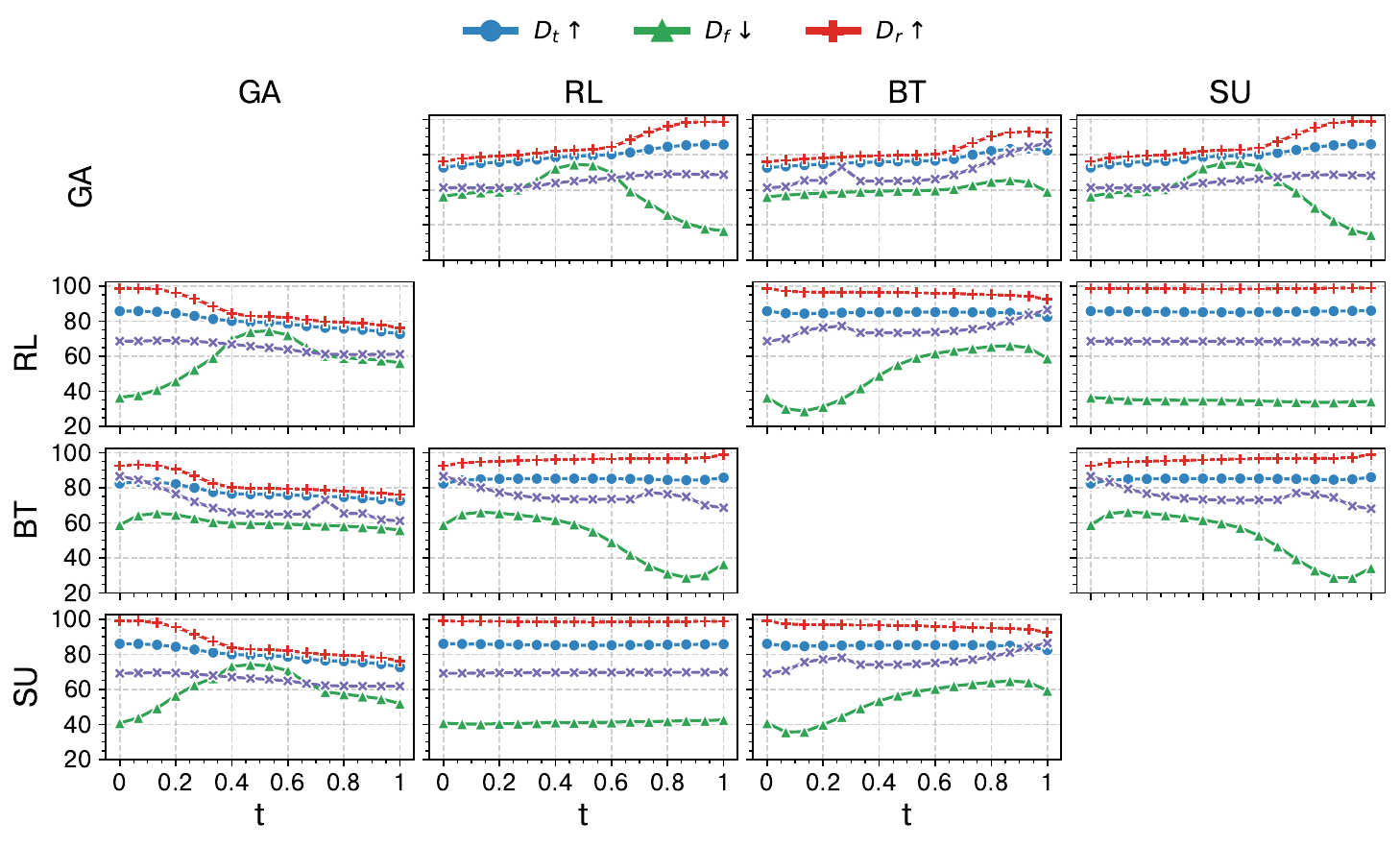}
        \caption{Linear MCU when $|D_f|=8.0\%$}
    \end{subfigure}
    \begin{subfigure}{0.49\textwidth}
        \centering
        \includegraphics[width=\linewidth]{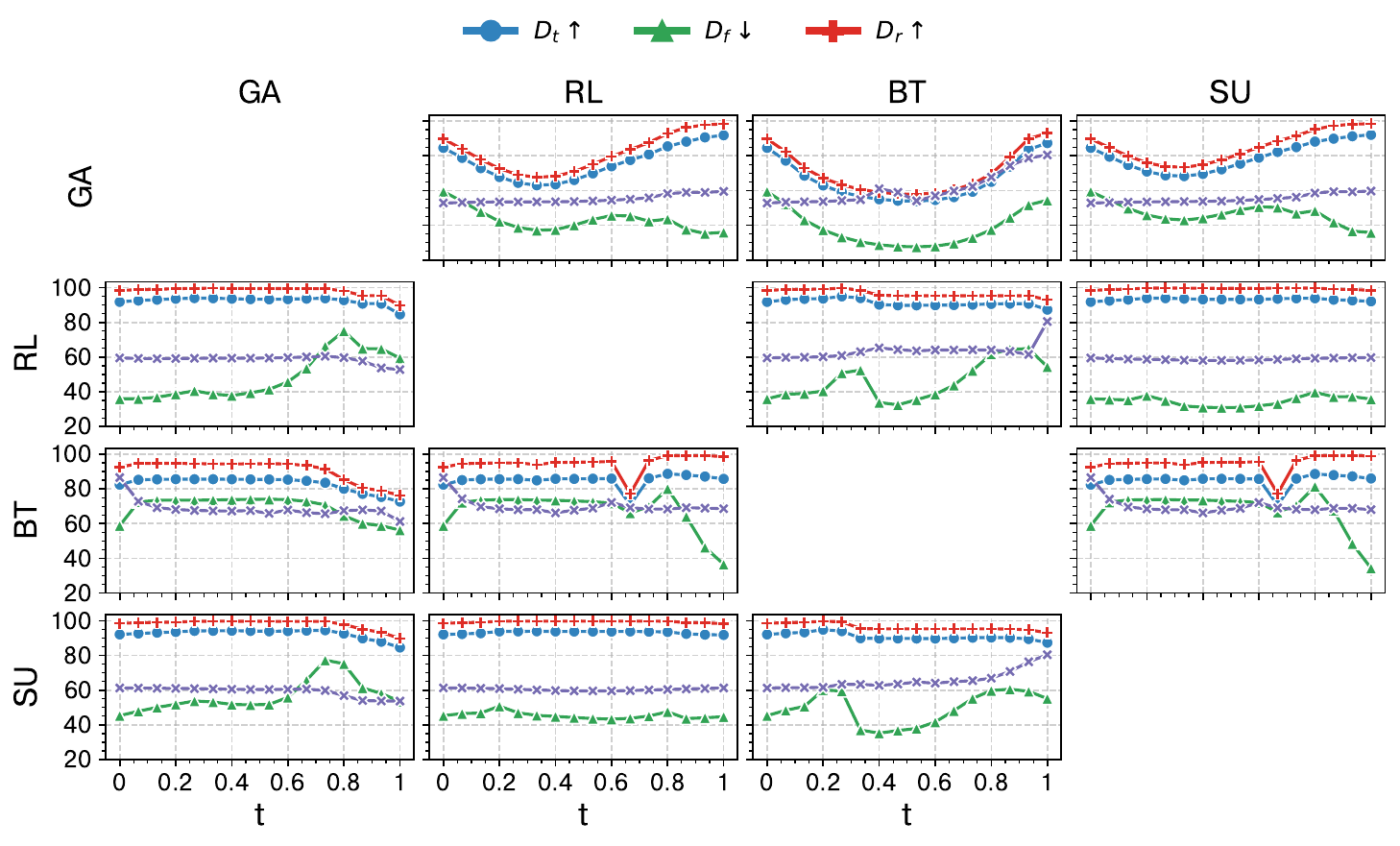}
        \caption{Quadratic MCU when $|D_f|=8.0\%$}
    \end{subfigure}
    \begin{subfigure}{0.49\textwidth}
        \centering
        \includegraphics[width=\linewidth]{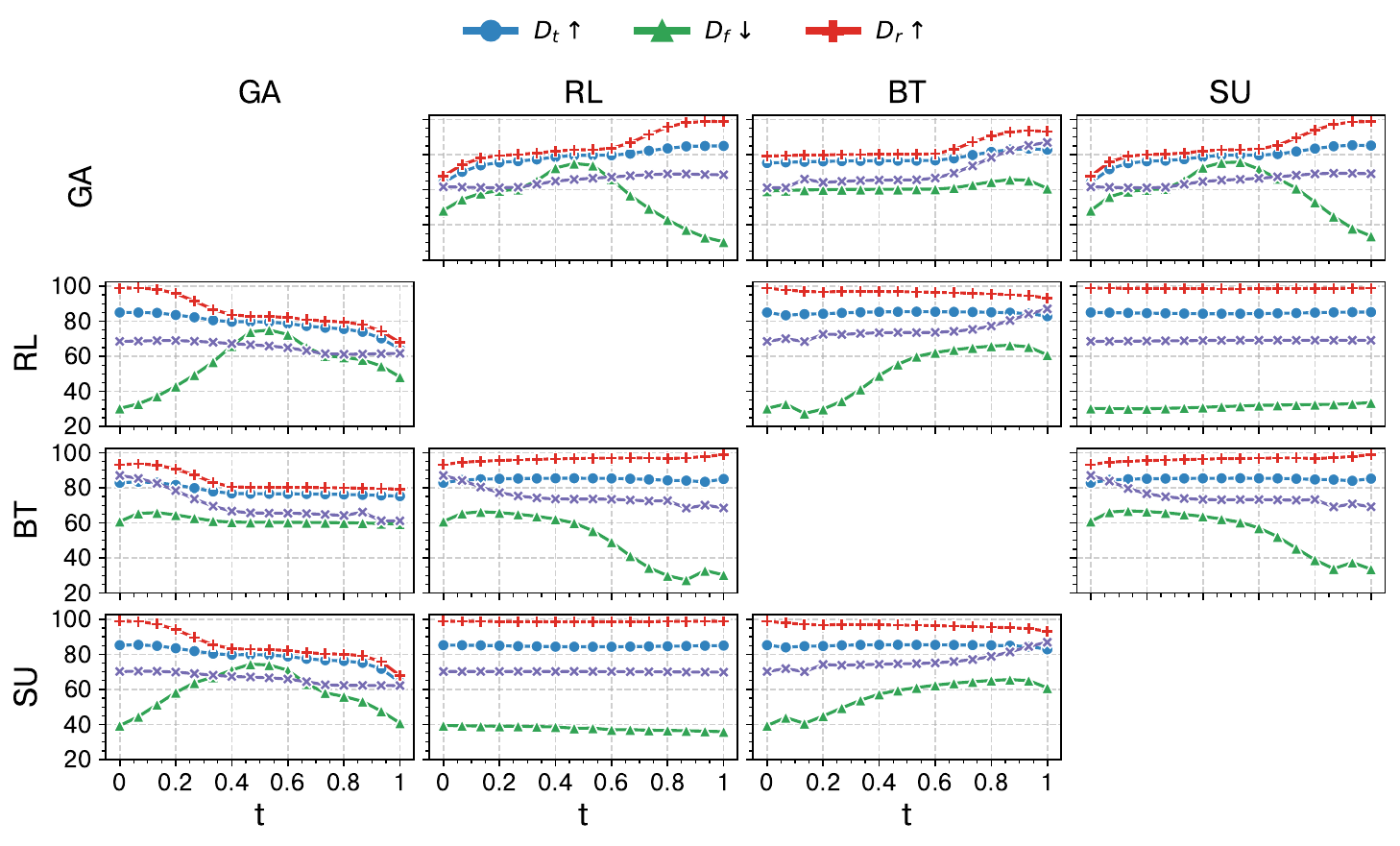}
        \caption{Linear MCU when $|D_f|=10.0\%$}
    \end{subfigure}
    \begin{subfigure}{0.49\textwidth}
        \centering
        \includegraphics[width=\linewidth]{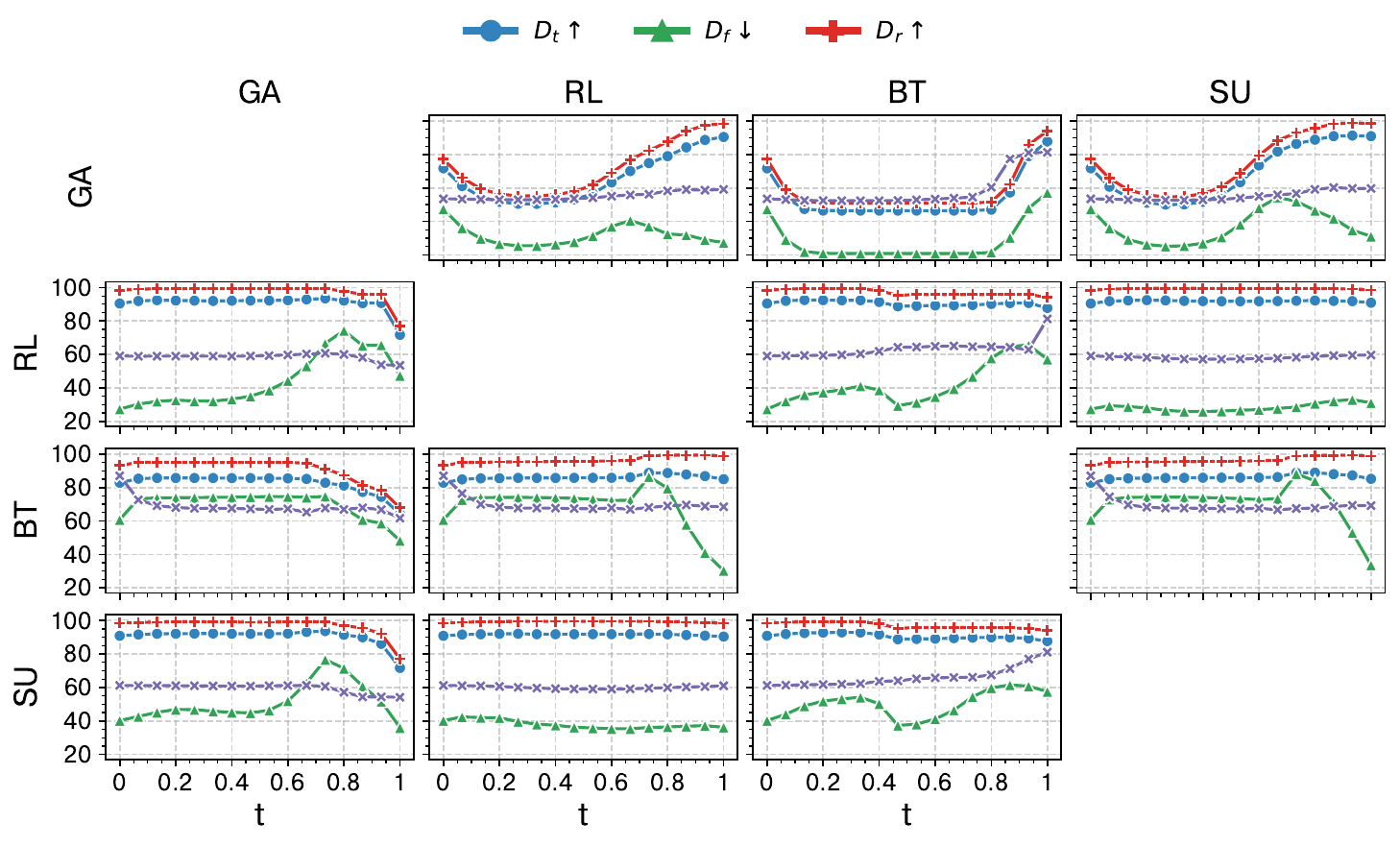}
        \caption{Quadratic MCU when $|D_f|=10.0\%$}
    \end{subfigure}
\end{center}
\caption{MCU under \textbf{Met-CL-Non-CL} setting on \textbf{classification datasets}.}
\label{fig:cls-met-cl-non-cl}
\end{figure}

\begin{figure}
\begin{center}
    \begin{subfigure}{0.49\textwidth}
        \includegraphics[width=\linewidth]{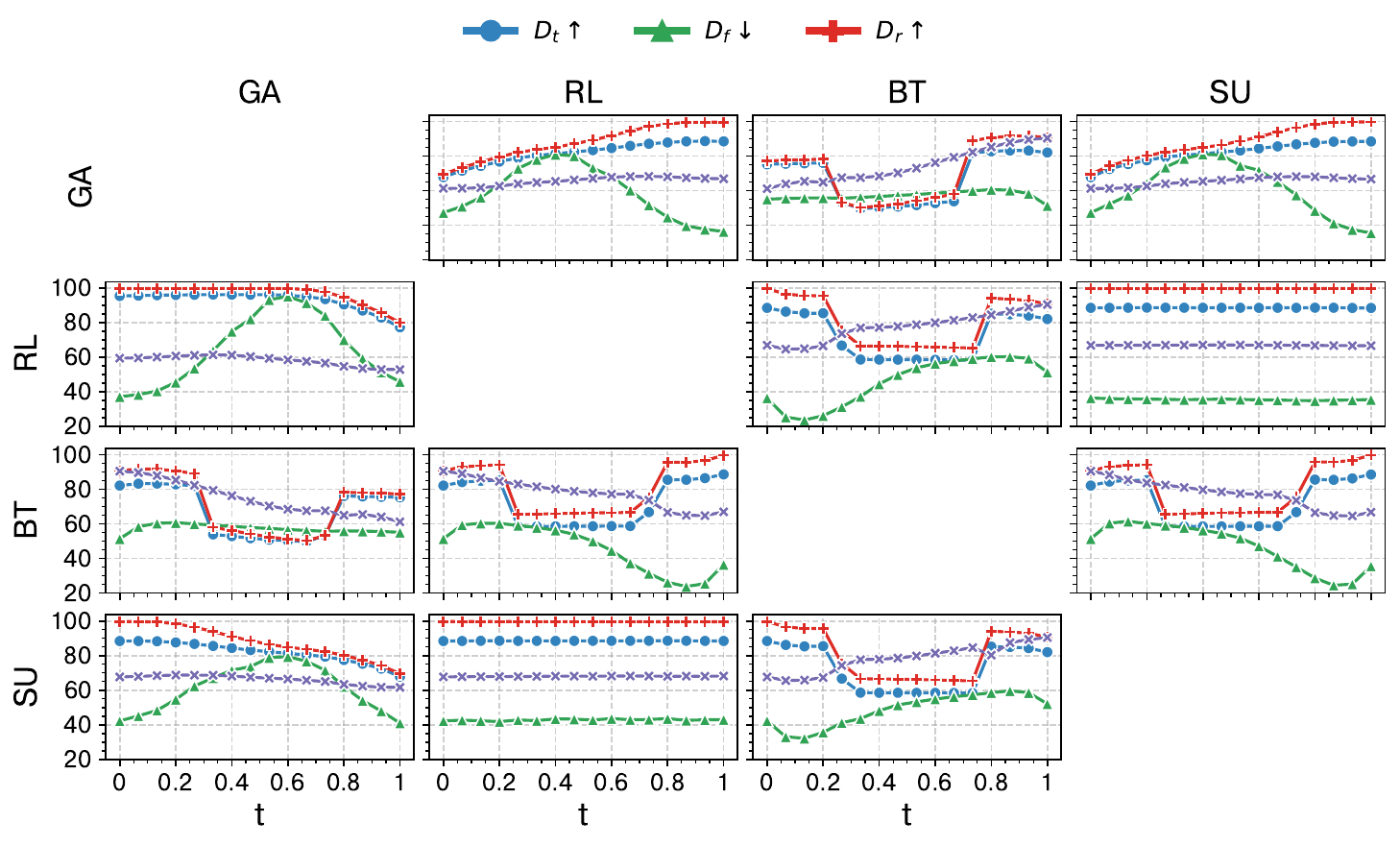}
        \caption{Linear MCU when $|D_f|=2.0\%$}
    \end{subfigure}
    \begin{subfigure}{0.49\textwidth}
        \includegraphics[width=\linewidth]{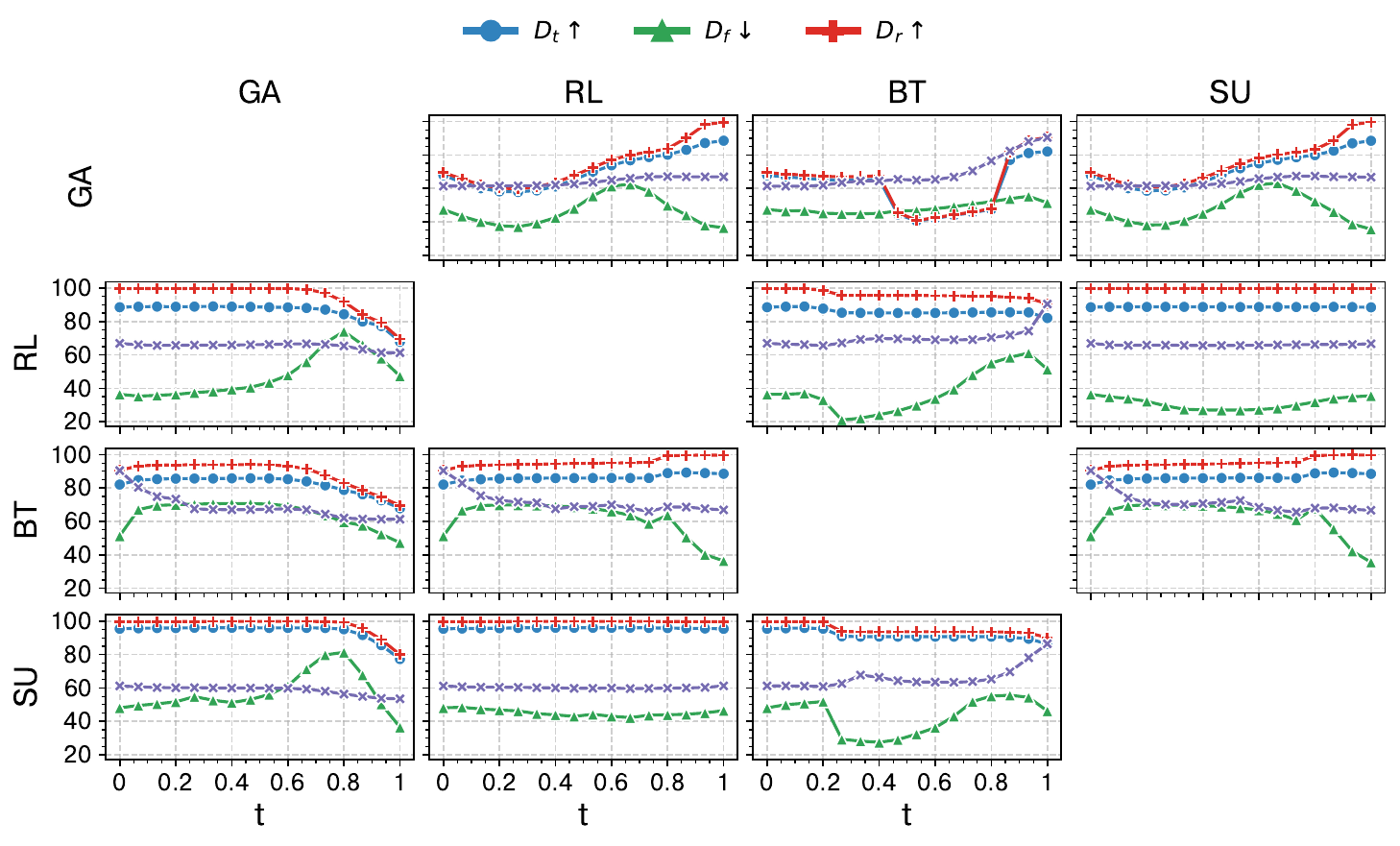}
        \caption{Quadratic MCU when $|D_f|=2.0\%$}
    \end{subfigure}
    \begin{subfigure}{0.49\textwidth}
        \centering
        \includegraphics[width=\linewidth]{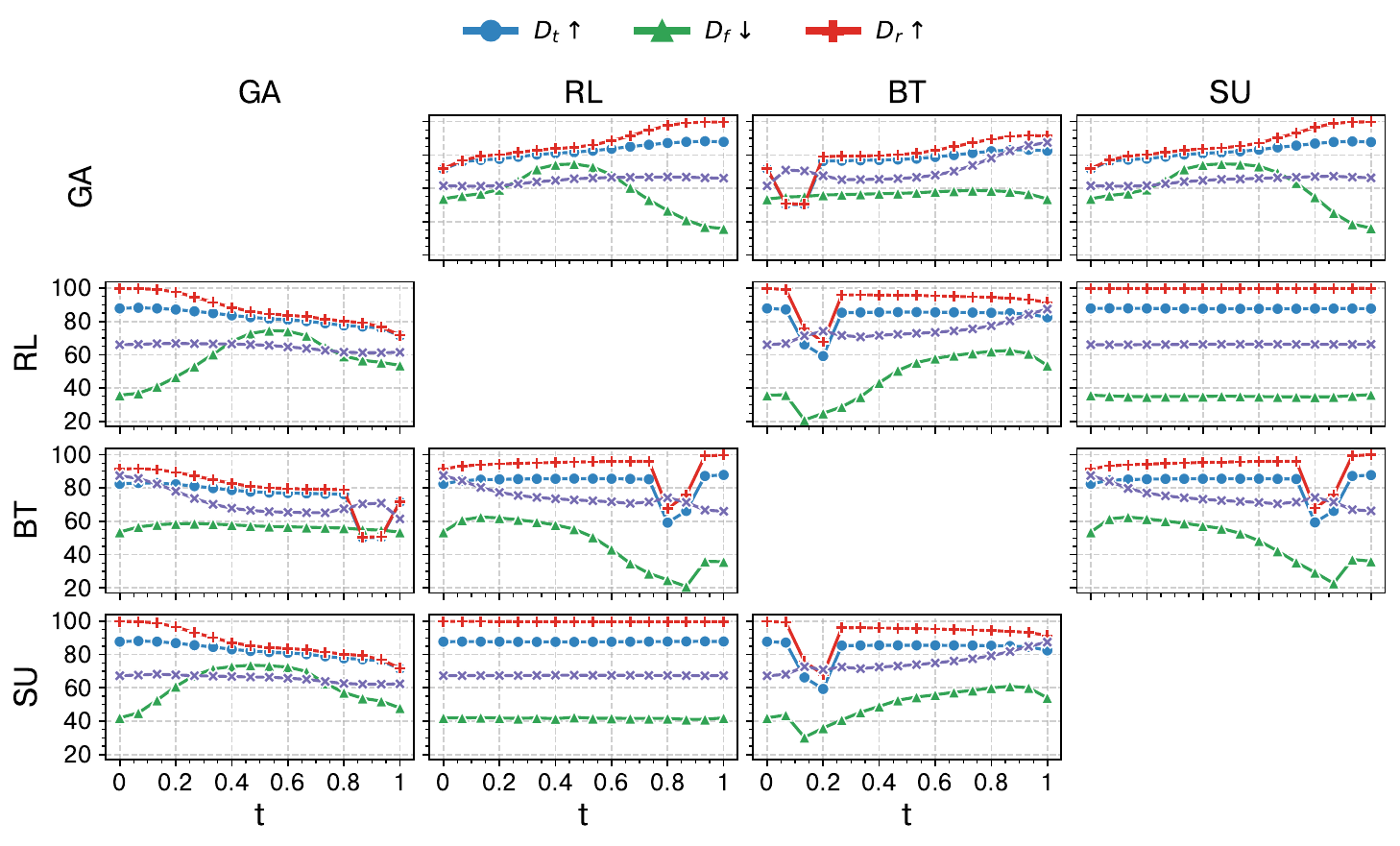}
        \caption{Linear MCU when $|D_f|=4.0\%$}
    \end{subfigure}
    \begin{subfigure}{0.49\textwidth}
        \centering
        \includegraphics[width=\linewidth]{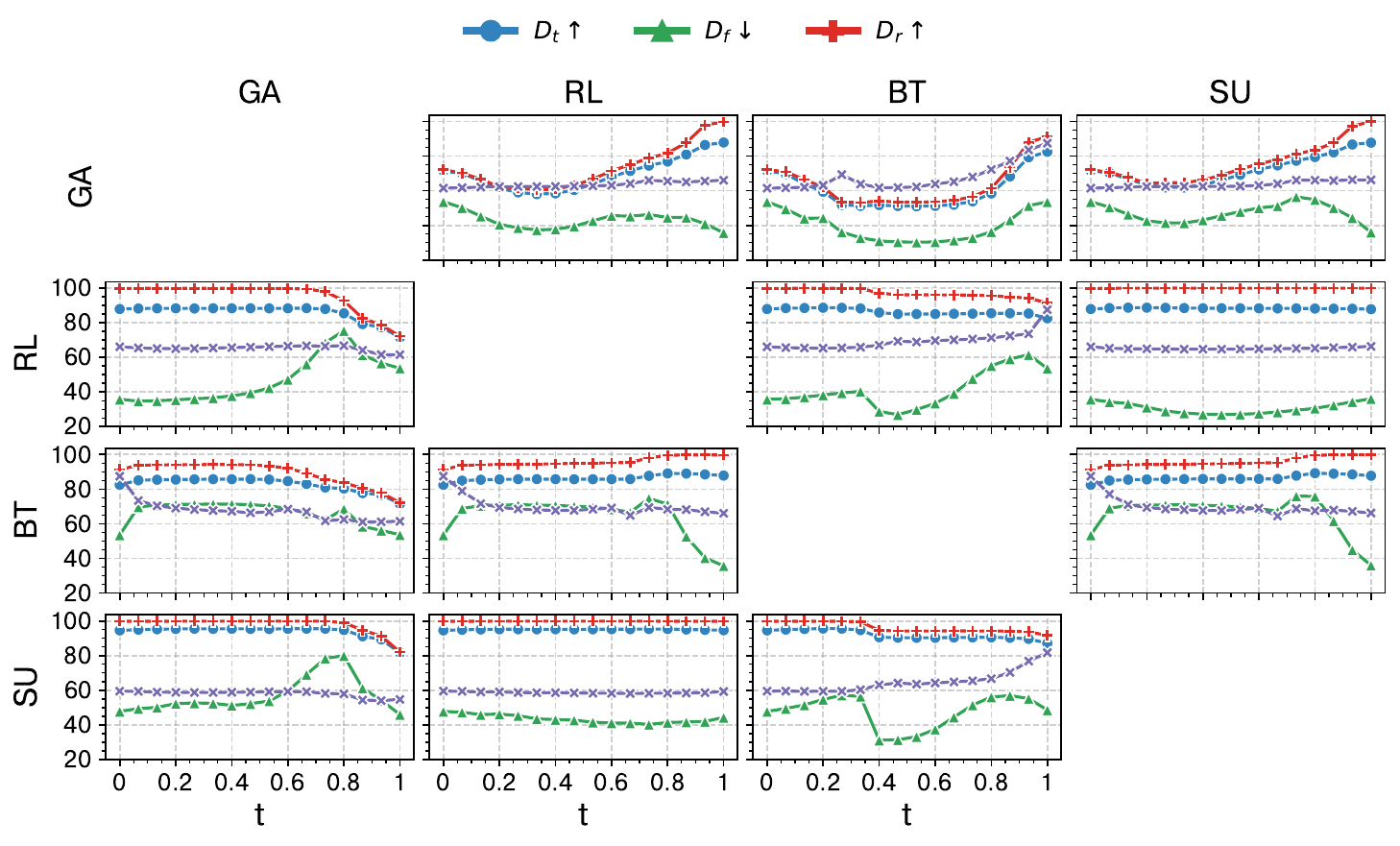}
        \caption{Quadratic MCU when $|D_f|=4.0\%$}
    \end{subfigure}
\end{center}
\end{figure}
\begin{figure}
\begin{center}
    \begin{subfigure}{0.49\textwidth}
        \centering
        \includegraphics[width=\linewidth]{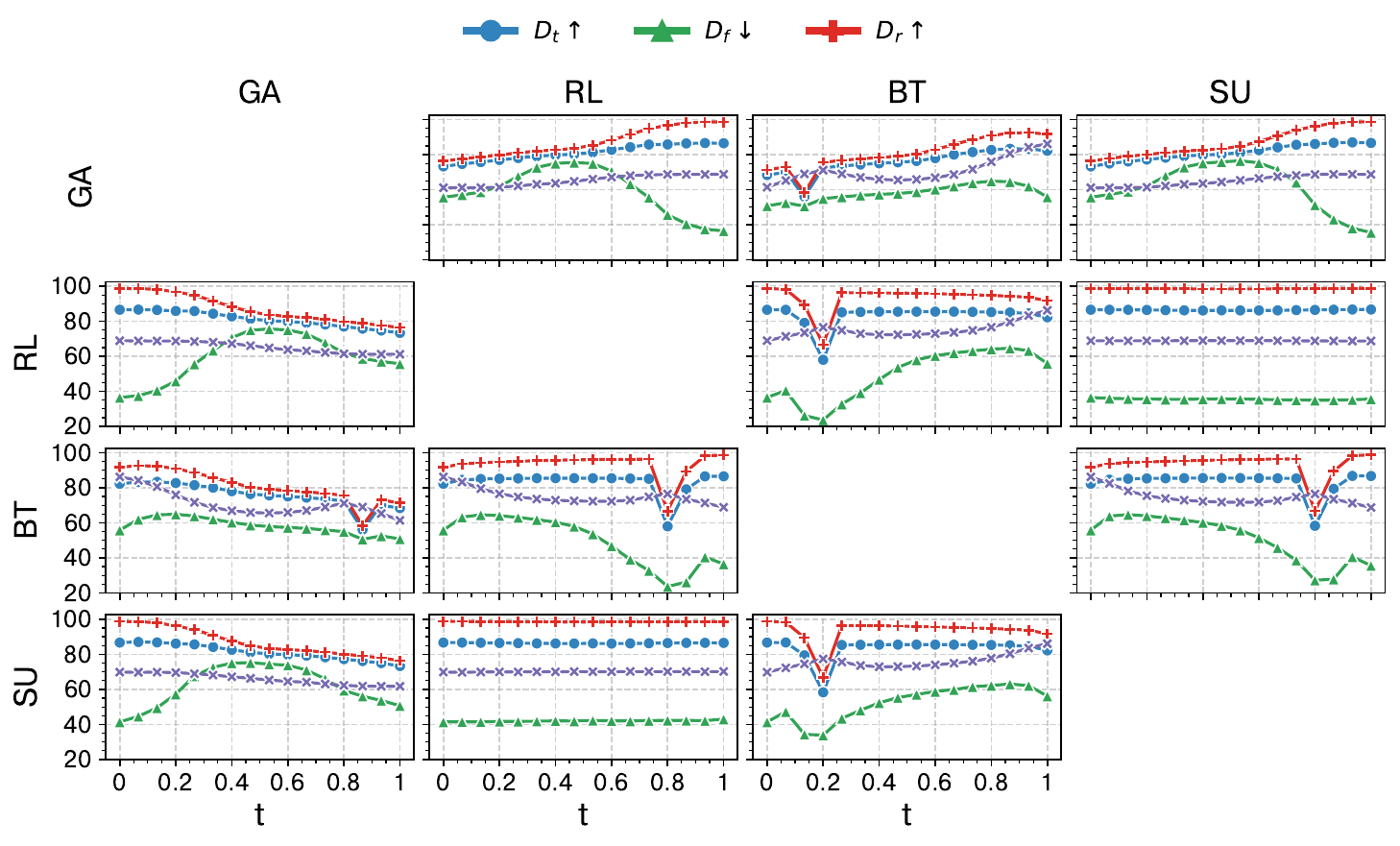}
        \caption{Linear MCU when $|D_f|=6.0\%$}
    \end{subfigure}
    \begin{subfigure}{0.49\textwidth}
        \centering
        \includegraphics[width=\linewidth]{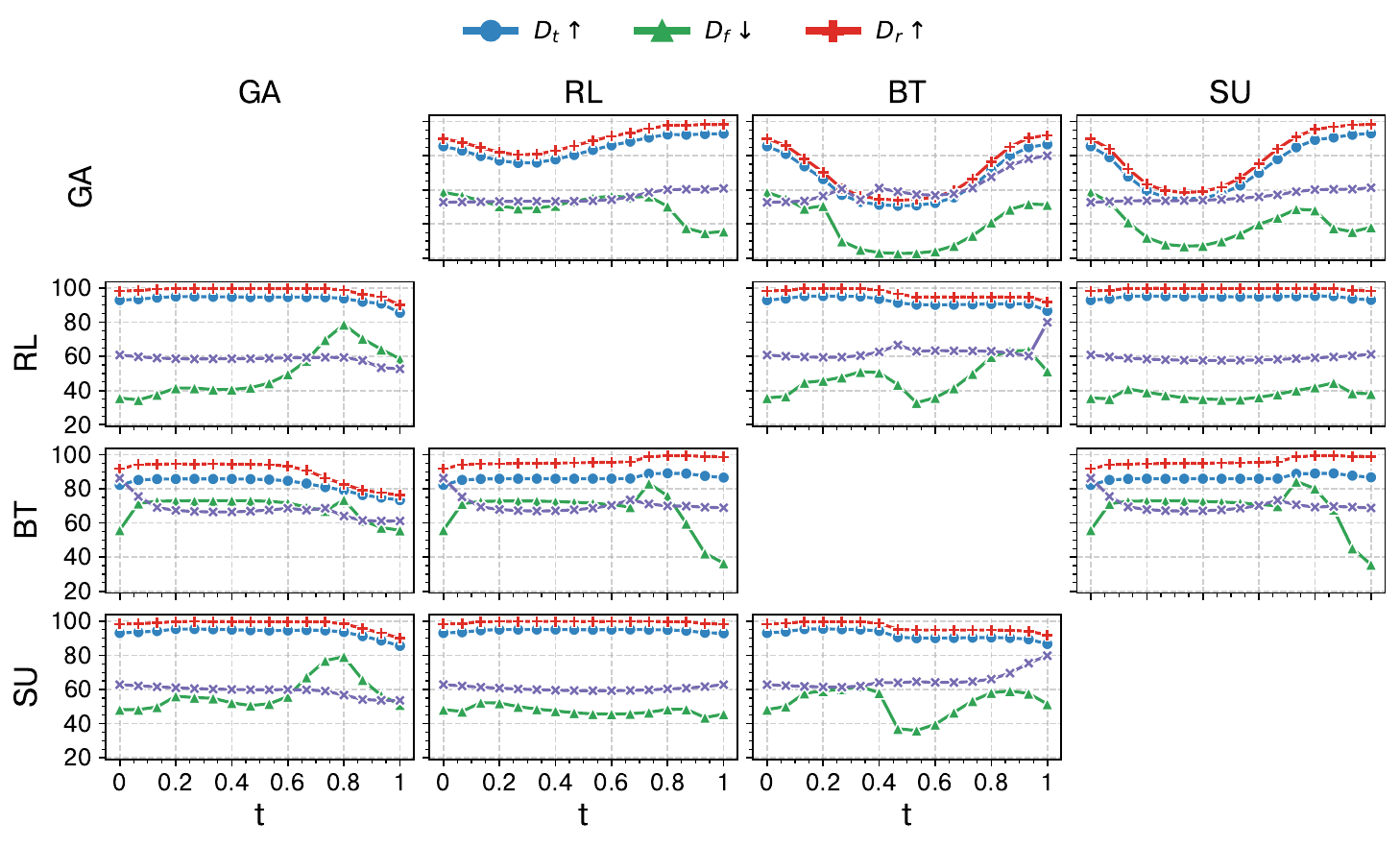}
        \caption{Quadratic MCU when $|D_f|=6.0\%$}
    \end{subfigure}
    \begin{subfigure}{0.49\textwidth}
        \centering
        \includegraphics[width=\linewidth]{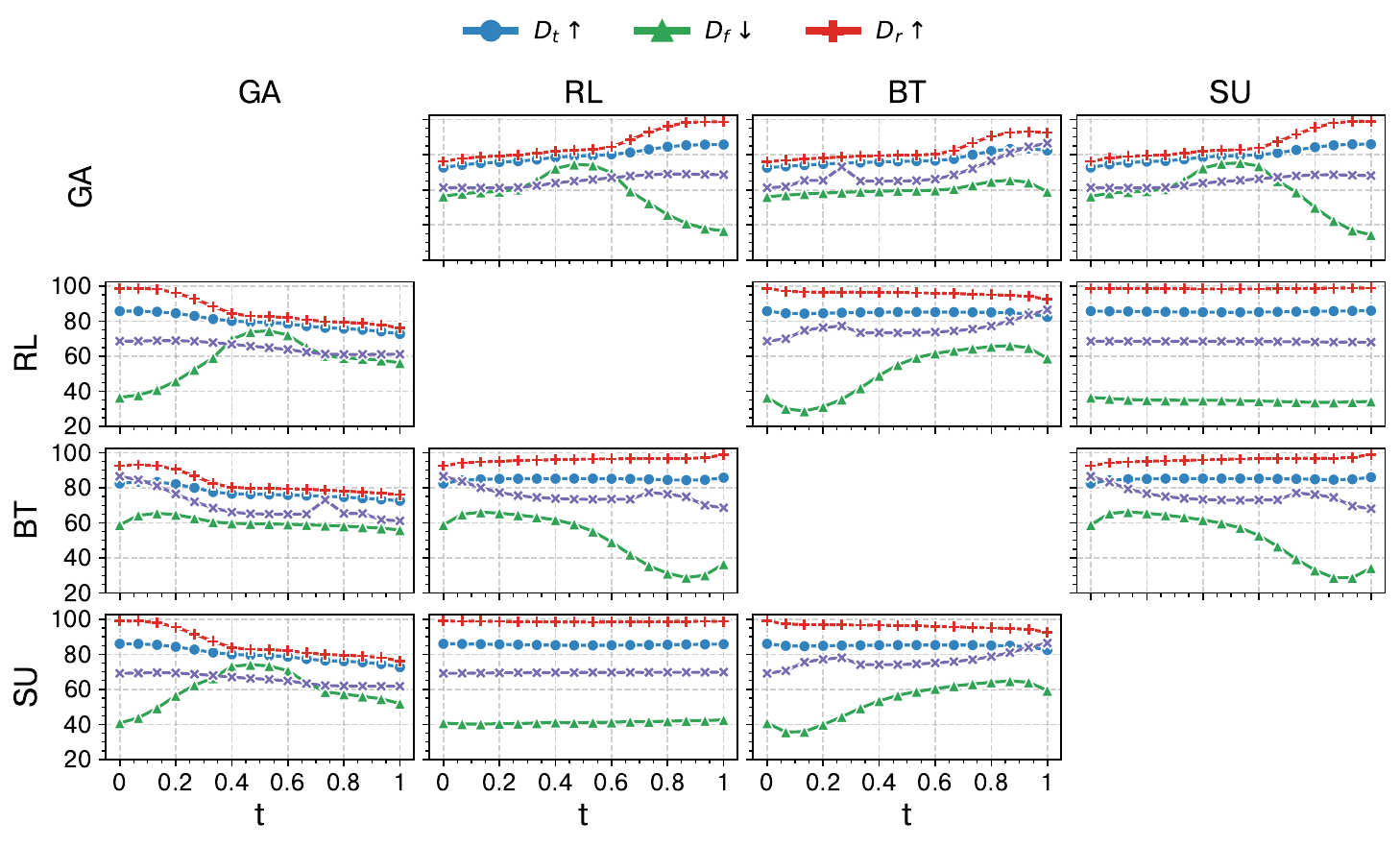}
        \caption{Linear MCU when $|D_f|=8.0\%$}
    \end{subfigure}
    \begin{subfigure}{0.49\textwidth}
        \centering
        \includegraphics[width=\linewidth]{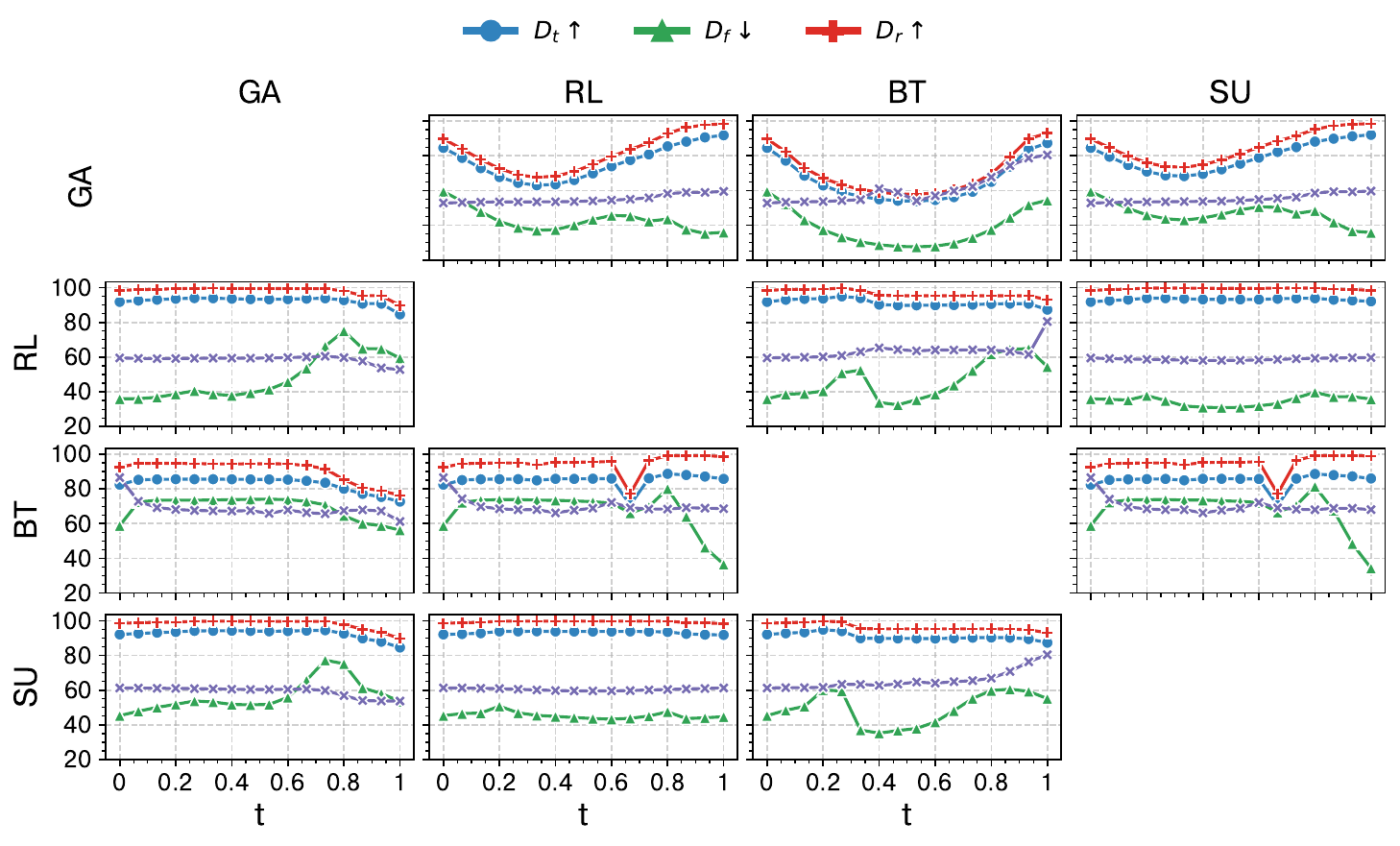}
        \caption{Quadratic MCU when $|D_f|=8.0\%$}
    \end{subfigure}
    \begin{subfigure}{0.49\textwidth}
        \centering
        \includegraphics[width=\linewidth]{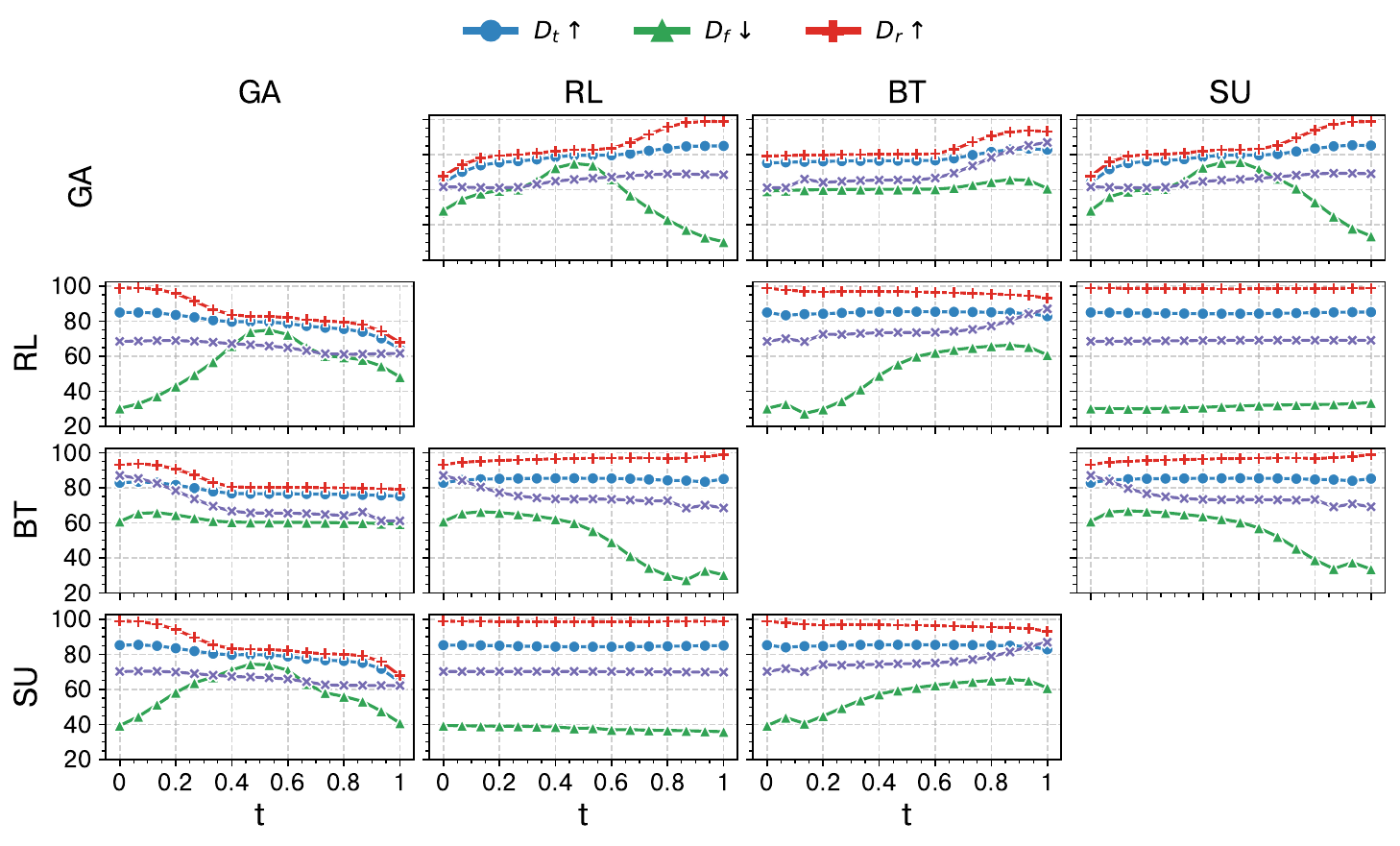}
        \caption{Linear MCU when $|D_f|=10.0\%$}
    \end{subfigure}
    \begin{subfigure}{0.49\textwidth}
        \centering
        \includegraphics[width=\linewidth]{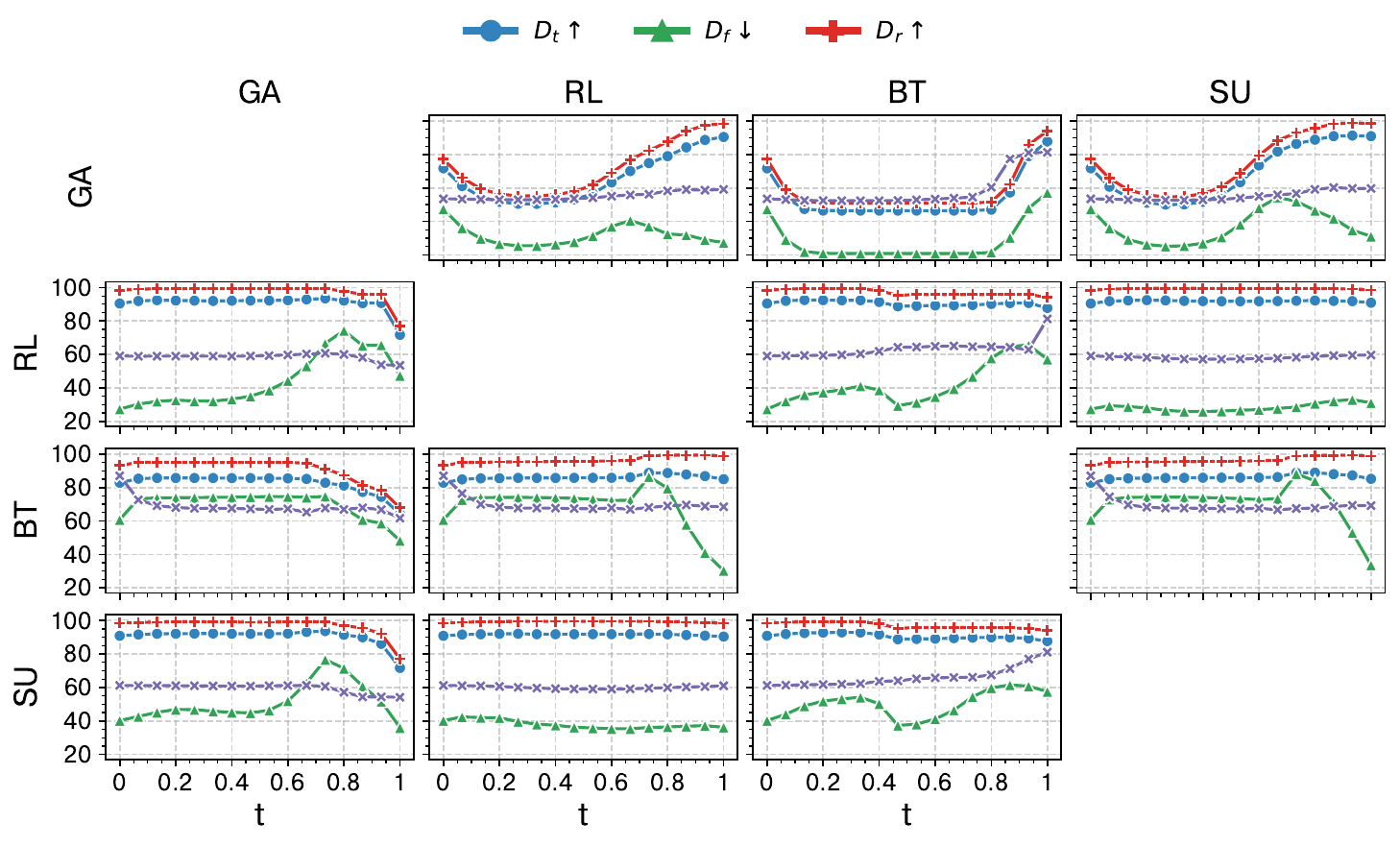}
        \caption{Quadratic MCU when $|D_f|=10.0\%$}
    \end{subfigure}
\end{center}
\caption{MCU under \textbf{Met-FO-SO} setting on \textbf{classification datasets}.}
\label{fig:cls-met-fo-so}
\end{figure}

\end{document}